\documentclass[5p,times,authoryear]{elsarticle}

\usepackage[english]{babel}
\usepackage{microtype}
\usepackage{setspace}
\usepackage{parskip}
\usepackage{graphicx}
\usepackage{subcaption}
\usepackage{booktabs}
\usepackage{multirow}
\usepackage{tabularx}
\usepackage{longtable}
\usepackage{bm}
\usepackage{array}
\usepackage{makecell}
\usepackage{adjustbox}
\usepackage{amsmath}
\usepackage{amsthm}
\usepackage{amssymb}
\usepackage[most]{tcolorbox}
\usepackage{xcolor}
\usepackage{colortbl}
\usepackage{tikz}
\usepackage{placeins}
\usepackage{hyphenat}
\usepackage{dblfloatfix}
\usepackage{flushend}
\usepackage{caption}
\usepackage{ragged2e}
\usepackage{xurl}
\usepackage[colorlinks=true,linkcolor=blue!55!black,citecolor=blue!55!black,urlcolor=blue!55!black]{hyperref}
\usepackage[noabbrev,nameinlink,capitalise]{cleveref}

\journal{Expert Systems with Applications}

\usetikzlibrary{shapes,arrows,positioning,fit,calc,backgrounds,trees,shapes.geometric}

\graphicspath{{./}}

\usetikzlibrary{positioning}
\usetikzlibrary{calc}
\usetikzlibrary{shapes.geometric}

\providecolor{tradingagent_main}{HTML}{37D2A6}
\providecolor{tradingagent_blue}{HTML}{377EB8}
\providecolor{tradingagent_orange}{HTML}{FF7F00}
\providecolor{tradingagent_red}{HTML}{D95F02}
\providecolor{tradingagent_purple}{HTML}{9E58AD}
\providecolor{tradingagent_gray}{HTML}{808080}
\providecolor{tradingagent_light}{HTML}{E8F5E9}
\providecolor{tradingagent_accent}{HTML}{FF6F00}

\providecolor{percept_green}{HTML}{37D2A6}
\providecolor{memory_blue}{HTML}{377EB8}
\providecolor{reason_orange}{HTML}{FF7F00}
\providecolor{alpha_purple}{HTML}{9E58AD}
\providecolor{portfolio_teal}{HTML}{009688}
\providecolor{risk_red}{HTML}{D95F02}
\providecolor{learn_indigo}{HTML}{3F51B5}
\providecolor{multi_brown}{HTML}{8D6E63}
\providecolor{evo_cyan}{HTML}{00BCD4}

\tikzset{
	block/.style={
			rectangle,
			draw,
			fill=tradingagent_main!20,
			text width=5em,
			text centered,
			rounded corners=2pt,
			minimum height=3em,
			font=\sffamily\bfseries\small,
			align=center
		},
}

\tikzset{
	arrow/.style={
			->,
			>=stealth,
			thick,
			color=tradingagent_gray
		},
}

\tikzset{
	arrow bidirectional/.style={
			<->,
			>=stealth,
			thick,
			color=tradingagent_gray
		},
}

\tikzset{
	arrow feedback/.style={
			->,
			>=stealth,
			thick,
			dashed,
			color=tradingagent_gray
		},
}

\tikzset{
	container/.style={
			rectangle,
			draw,
			dashed,
			inner sep=10pt,
			fill=gray!10,
			rounded corners=3pt
		},
}

\tikzset{
	labelstyle/.style={
			font=\sffamily\small,
			color=black
		},
}

\tikzset{
	archcomponent/.style={
			block,
			minimum width=8em,
			fill=tradingagent_blue!20,
			font=\sffamily\bfseries
		},
}

\tikzset{
	process/.style={
			block,
			fill=tradingagent_main!20
		},
}

\tikzset{
	decision/.style={
			diamond,
			draw,
			fill=tradingagent_orange!20,
			text width=4em,
			text centered,
			inner sep=0pt,
			font=\sffamily\small,
			aspect=2
		},
}

\tikzset{
	panel/.style={
			rectangle,
			draw,
			fill=white,
			minimum width=10em,
			minimum height=8em,
			rounded corners=3pt,
			font=\sffamily\small
		},
}

\tikzset{
	timeline/.style={
			circle,
			draw,
			fill=tradingagent_main!30,
			minimum size=1cm,
			font=\sffamily\small\bfseries
		},
}

\tikzset{
	block blue/.style={
			block,
			fill=tradingagent_blue!20
		},
	block orange/.style={
			block,
			fill=tradingagent_orange!20
		},
	block red/.style={
			block,
			fill=tradingagent_red!20
		},
	block purple/.style={
			block,
			fill=tradingagent_purple!20
		},
	block teal/.style={
			block,
			fill=portfolio_teal!20
		},
	block indigo/.style={
			block,
			fill=learn_indigo!20
		},
	block brown/.style={
			block,
			fill=multi_brown!20
		},
	block cyan/.style={
			block,
			fill=evo_cyan!20
		},
}

\tikzset{
	vertical chain/.style={
			node distance=1.5cm,
			every node/.style={block}
		}
}

\tikzset{
	horizontal chain/.style={
			node distance=2cm,
			every node/.style={block}
		}
}

\pgfdeclarelayer{background}
\pgfsetlayers{background,main}

\newcommand{\CandidateRecordsN}{246}
\newcommand{\RecordsAfterDedupN}{246}

\newcommand{\EvidenceIncludedN}{185}
\newcommand{\EvidencePrimaryN}{80}
\newcommand{\EvidenceBackgroundN}{105}

\newcommand{\CandidateExcludedN}{61}
\newcommand{\ExcludedOutOfScopeN}{61}

\newcommand{\ReproRZeroN}{45}
\newcommand{\ReproROneN}{33}
\newcommand{\ReproRTwoN}{2}
\newcommand{\ReproRThreeN}{0}
\newcommand{\ReproRTwoPlusN}{2}

\newcommand{\ReproRZeroPct}{56.2}
\newcommand{\ReproROnePct}{41.2}
\newcommand{\ReproRTwoPct}{2.5}
\newcommand{\ReproRThreePct}{0.0}
\newcommand{\ReproRTwoPlusPct}{2.5}

\newcommand{\ProtocolSplitReportedN}{50}
\newcommand{\ProtocolCostReportedN}{47}
\newcommand{\ProtocolUniverseReportedN}{48}
\newcommand{\ProtocolExecutionReportedN}{47}
\newcommand{\ProtocolArtifactsReportedN}{35}

\newcommand{\ProtocolSplitReportedPct}{62.5}
\newcommand{\ProtocolCostReportedPct}{58.8}
\newcommand{\ProtocolUniverseReportedPct}{60.0}
\newcommand{\ProtocolExecutionReportedPct}{58.8}

\newcommand{\PrimaryRecentN}{79}
\newcommand{\PrimaryPeerReviewedN}{19}
\newcommand{\PrimaryNonPeerReviewedN}{61}
\newcommand{\PrimaryPeerReviewUnknownN}{0}

\newcommand{\WindowExtensionRecordsN}{154}

\def\ExtractionAuditSampleN{15}
\def\ExtractionAuditFieldComparisonsN{150}
\def\ExtractionAuditAgreementPct{100.0}
\def\ExtractionAuditKappaNote{field-dependent; see Supplementary Material S3}

\definecolor{tradingagent_main}{HTML}{37D2A6}
\definecolor{tradingagent_blue}{HTML}{377EB8}
\definecolor{tradingagent_orange}{HTML}{FF7F00}
\definecolor{tradingagent_red}{HTML}{D95F02}
\definecolor{tradingagent_purple}{HTML}{9E58AD}
\definecolor{tradingagent_gray}{HTML}{808080}
\definecolor{tradingagent_light}{HTML}{E8F5E9}

\tcbset{
  breakable,
  before upper={\RaggedRight\setlength{\emergencystretch}{3em}%
    \widowpenalty=10000\clubpenalty=10000},
  fontupper=\small
}

\newtcolorbox{classificationbox}[2][]{
  colback=tradingagent_light!20,
  colframe=tradingagent_blue!65,
  colbacktitle=tradingagent_blue!15,
  coltitle=black,
  title={\bfseries #2},
  boxrule=0.6pt,
  arc=2pt,
  parbox=false,
  before skip=5pt,
  after skip=10pt,
  left=5pt,
  right=5pt,
  #1
}

\newtcolorbox{evidencebox}[2][]{
  breakable,
  title after break={\bfseries #2 (continued)},
  before upper app={\widowpenalty=10000\clubpenalty=10000},
  colback=tradingagent_light!16,
  colframe=tradingagent_blue!55,
  colbacktitle=tradingagent_blue!12,
  coltitle=black,
  title={\bfseries #2},
  boxrule=0.6pt,
  arc=2pt,
  parbox=false,
  before skip=5pt,
  after skip=10pt,
  left=5pt,
  right=5pt,
  floatplacement=tbp,
  #1
}

\newcommand{\evidenceitem}[1]{\item\parbox[t]{\linewidth}{#1}}


\begin{document}
\raggedbottom
\sloppy

\begin{frontmatter}

\title{Agentic Trading: When LLM Agents Meet Financial Markets}

\author[szu]{Yihan Xia}
\ead{xiayihan40@gmail.com}
\author[szu]{Panpan You}
\ead{xiaomu2311@gmail.com}
\author[szu]{Taotao Wang\corref{cor1}}
\ead{ttwang@szu.edu.cn}
\author[szu]{Fang Liu}
\ead{liuf@szu.edu.cn}
\author[safi]{Han Qi}
\ead{steffan@szu.edu.cn}
\author[szu]{Xiaoxiao Wu}
\ead{xxwu.eesissi@szu.edu.cn}
\author[szu]{Shengli Zhang}
\ead{zsl@szu.edu.cn}

\cortext[cor1]{Corresponding author.}
\address[szu]{College of Electronic and Information Engineering, Shenzhen University, Shenzhen, China}
\address[safi]{Shenzhen Audencia Financial Technology Institute, Shenzhen University, Guangdong, People's Republic of China}

\begin{abstract}
The integration of Large Language Models (LLMs) with autonomous agent architectures is transforming quantitative finance, shifting algorithmic trading from isolated predictive modeling toward fully autonomous, closed-loop trading agents. These systems perceive multimodal market signals, maintain active context, associative vector stores, and structured knowledge bases, execute complex strategic reasoning, emit executable trading actions, and adapt continuously to non-stationary market regimes. This survey provides a comprehensive, systematic, and forward-looking synthesis of the rapidly evolving landscape of agentic trading. We structure the literature through a unified \textbf{Architecture--Capability--Adaptation (A-C-A)} conceptual framework: (i)~\textit{Architecture} deconstructs the foundational cognitive loop into perception (textual, numerical, and multimodal), memory systems (working memory, associative vector storage, and structured knowledge bases), cognitive reasoning (reactive, reflective, and planning-driven), and execution mechanisms (order routing and friction management); (ii)~\textit{Capability} examines primary financial functions, including automated alpha discovery, portfolio optimization, and multi-tier risk control; and (iii)~\textit{Adaptation} investigates how agents evolve through in-context learning, supervised fine-tuning, reinforcement learning, multi-agent coordination, and architectural self-evolution. Drawing on \EvidenceIncludedN~systematically selected studies (including a core empirical subset of $n=\EvidencePrimaryN$~systems evaluated under closed-loop market conditions), we analyze design spaces, operational trade-offs, and emerging paradigms. Furthermore, as an essential cross-cutting dimension, we examine empirical evaluation practices across the literature---including backtesting split schemes, transaction-cost realism, survivorship-bias handling, and execution fidelity---and distill methodological best practices to guide future research. By bridging artificial intelligence, market microstructure, and quantitative finance, this survey aims to serve as a foundational roadmap for designing, benchmarking, and deploying reliable agentic trading systems.
\end{abstract}


\end{frontmatter}


\section{Introduction}
\label{sec:intro}


The integration of Large Language Models (LLMs) into financial trading
marks a shift from traditional quantitative prediction
models toward more autonomous, agentic systems.
Conventional quantitative finance
approaches typically employ black-box machine learning models that generate
price or return predictions without explicit reasoning chains or adaptive
capabilities. In contrast, LLM-based trading agents often implement an explicit
loop---perceiving market information, storing experiences, reasoning about
decisions, executing trades, and learning from outcomes \cite{wu2024finagent, xiao2024tradingagents, yu2024fincon}.

\begin{figure*}[!ht]
	\centering
	\includegraphics[width=0.95\linewidth]{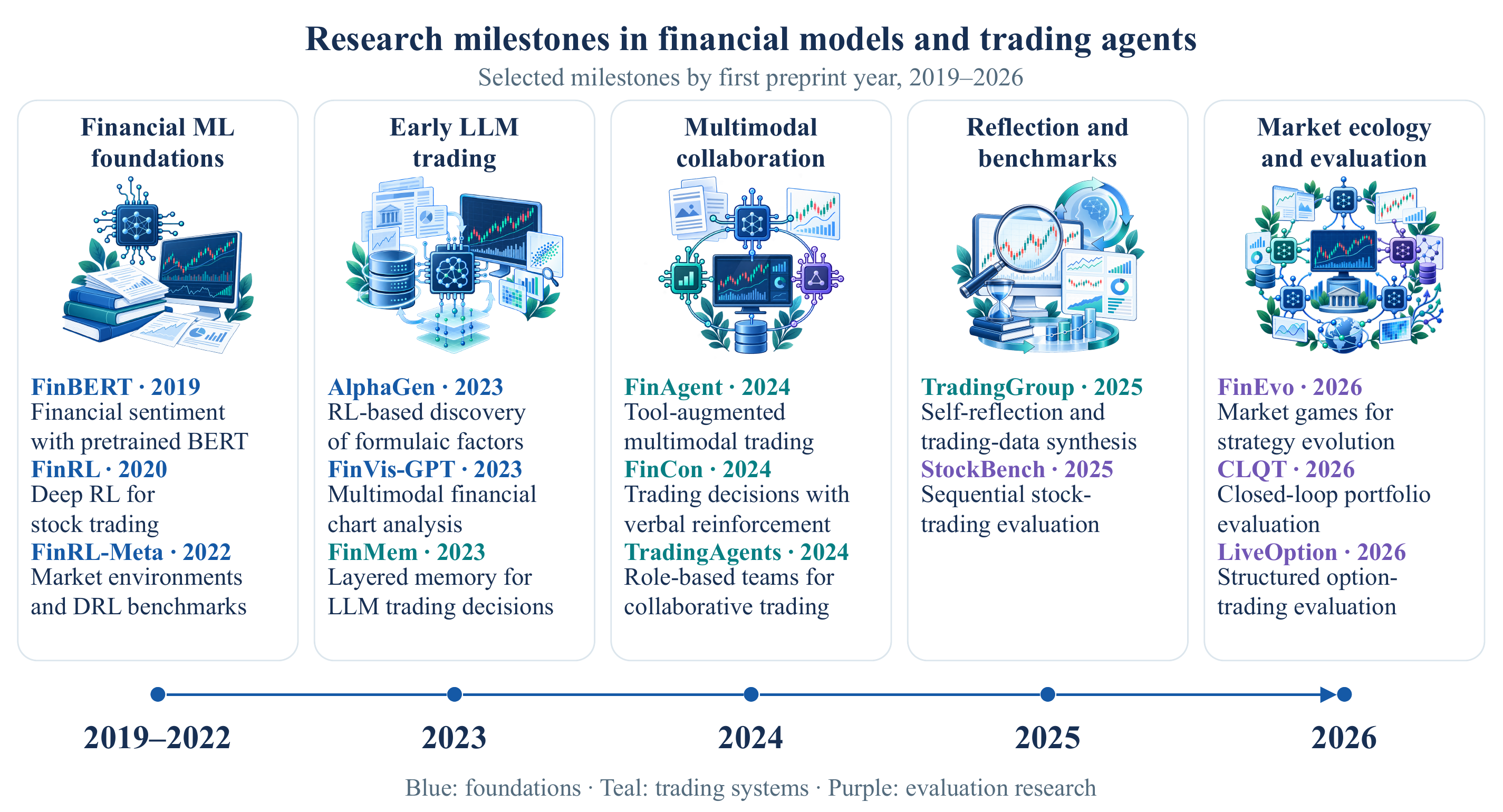}
	\caption{Selected research timeline of financial language models and agentic
        trading, 2019--2026. Five thematic periods connect financial language
        modeling and reinforcement-learning foundations with early LLM trading,
        multimodal collaboration, reflection and evaluation, and market ecology with expanded evaluation.
        Each paper is labeled by its first arXiv submission year. Panel widths are
        for readability and do not represent period duration. Foundation papers
        provide historical context rather than expanding the empirical review corpus
        \cite{araci2019finbert,liu2020finrl,liu2022finrlmeta,yu2023alphagen,wang2023finvis,yu2023finmem,wu2024finagent,yu2024fincon,xiao2024tradingagents,tradinggroup_2025,stockbench_2025,zou2026finevo,qu2026clqt,luo2026liveoption}.}
	\label{fig:agency-spectrum}
\end{figure*}

\Cref{fig:agency-spectrum} situates representative papers in the chronological
development of the field, illustrating the evolution from domain-adapted
financial language models to multimodal and multi-agent trading systems.
To provide a structured and comprehensive navigation of this design space,
this survey synthesizes \textbf{\EvidenceIncludedN} included studies screened
through \textbf{2026-09-30}, combining an initial multi-source ledger
(cutoff 2026-03-09) with a formal extension window covering 2026-03-10 to
2026-09-30. To help readers distinguish full trading systems from component
modules, the literature is organized into a primary empirical cohort
($n=\EvidencePrimaryN$) evaluated under closed-loop market conditions (satisfying
\textit{Action Output + Closed-Loop Evaluation}) and an informative background
tier (\textbf{\EvidenceBackgroundN} studies) covering foundation models,
isolated predictors, and specialized tools. Furthermore, to provide an objective
view of current empirical practices, our synthesis also systematically documents
how key experimental protocols are reported across the literature---including
time-consistent split designs (\textbf{\ProtocolSplitReportedN/\EvidencePrimaryN}),
explicit transaction costs (\textbf{\ProtocolCostReportedN/\EvidencePrimaryN}),
universe survivorship controls (\textbf{\ProtocolUniverseReportedN/\EvidencePrimaryN}),
and execution semantics (\textbf{\ProtocolExecutionReportedN/\EvidencePrimaryN}).

\begin{figure*}[!tb]
	\centering
	\includegraphics[width=\linewidth]{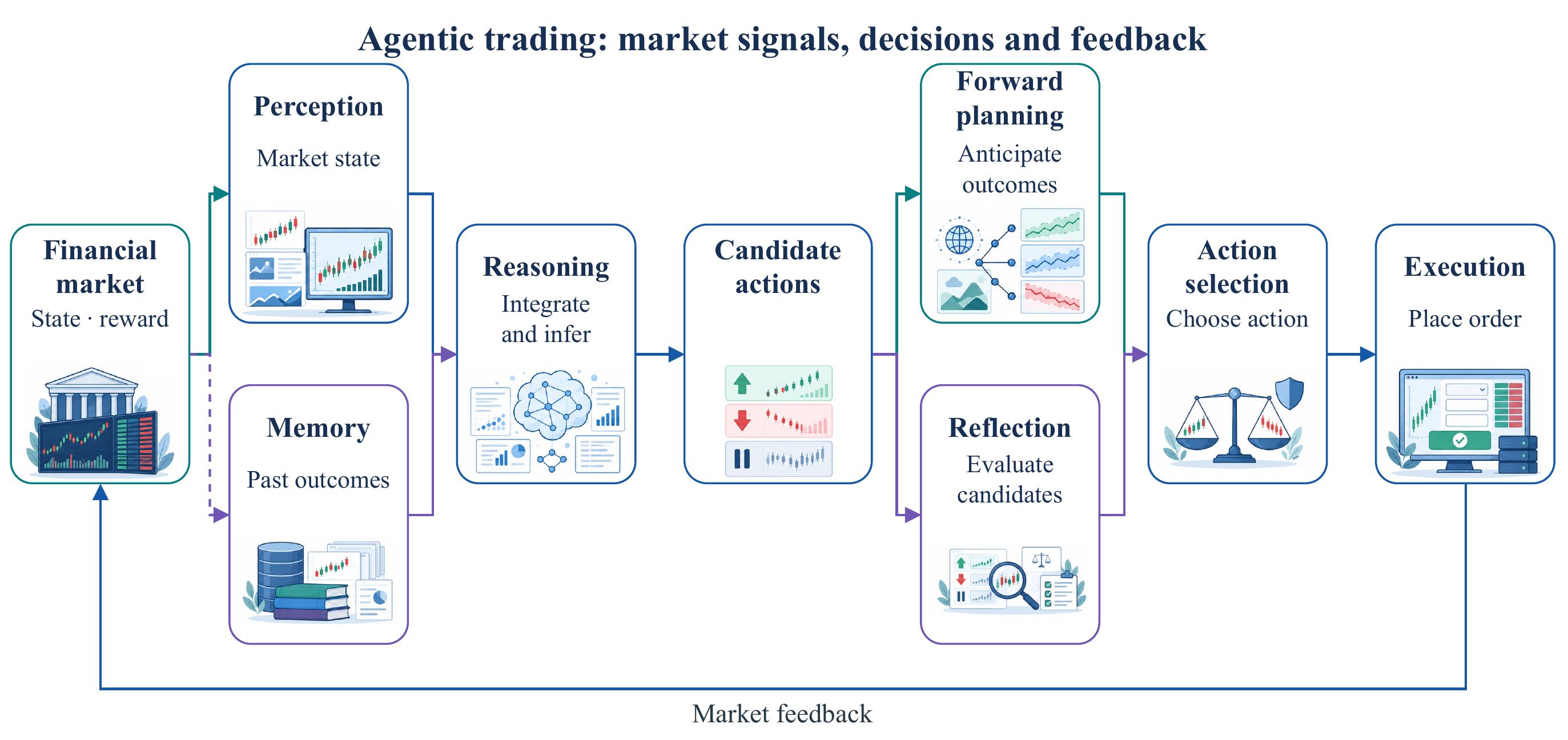}
	\caption{End-to-end decision and execution loop in agentic trading. Market observations and historical experiences are gathered via perception and memory modules; cognitive reasoning produces candidate actions that are validated through forward planning or reflection; the execution module maps decisions to market orders under risk constraints; and market outcomes provide feedback for continuous adaptation.}
	\label{fig:reasoning-flow-intro}
\end{figure*}

Realizing the potential of agentic trading systems involves distinct engineering
and architectural considerations. First, LLM-based agents generate natural-language
rationales and multi-step reasoning traces; ensuring that these explanations
faithfully reflect execution requires grounded observation spaces and structured
decision logs. Second, agents can adapt their strategies under changing market
regimes by drawing on episodic memory and dynamic reflection, demanding robust
risk boundaries against catastrophic drift. Third, modular designs allow
specialized components for perception, memory, reasoning, and execution
(see \Cref{fig:reasoning-flow-intro}), enabling systems to be decoupled,
tested, and enhanced incrementally.

Trading agents are essentially autonomous financial decision-support pipelines:
they transform raw, multimodal market observations into executable actions
subject to real-world friction, liquidity constraints, and supervisory controls.
A comprehensive survey of this domain must therefore examine the complete
decision loop: what information channels are perceived, how memory is indexed,
which reasoning paradigms generate trading hypotheses, how orders are routed,
and how agents adapt to non-stationary market feedback.


\subsection{Related Surveys and Complementary Positioning}
\label{ssec:related-surveys}

As artificial intelligence intersects with financial engineering, several
surveys and benchmark suites have mapped distinct facets of this evolving
domain (summarized in \Cref{tab:intro-related}). We position our work as
complementary to these existing contributions across four main lines:

\begin{table*}[!tb]
	\centering
	\small
	\setlength{\tabcolsep}{4.5pt}
	\renewcommand{\arraystretch}{1.18}
	\caption{Scopes and organizing perspectives of related surveys, benchmarks, and financial AI frameworks.}
	\label{tab:intro-related}
	\begin{tabularx}{\linewidth}{@{}>{\raggedright\arraybackslash}p{2.75cm} >{\raggedright\arraybackslash}p{1.55cm} >{\raggedright\arraybackslash}p{2.3cm} >{\raggedright\arraybackslash}p{2.7cm} >{\raggedright\arraybackslash}p{3.15cm} >{\raggedright\arraybackslash}X@{}}
		\toprule
		\textbf{Work} & \textbf{Type} & \textbf{Target Domain} & \textbf{Primary Review Focus} & \textbf{System Decomposition} & \textbf{Evaluation Perspective} \\
		\midrule
		\textbf{This survey} & Survey & Trading agents & End-to-end trading pipelines & Perception, memory, reasoning, execution & A-C-A framework and empirical protocols \\
		\citet{wang2024llmagentsurvey} & Survey & General AI agents & Cognitive agent architectures & Profile, memory, planning, action & General capabilities and benchmarks \\
		\citet{giglio2022factormodels} & Survey & Asset pricing and ML & Factor models and econometrics & Statistical estimation and pricing kernels & Econometric testing and predictability \\
		FinBen\newline\citep{finben_2024} & Benchmark & Financial NLP and quant tasks & Standardized task evaluation & Model-level task mapping & Financial tasks across difficulty tiers \\
		FinGPT\newline\citep{fingpt_2023} & Model/\allowbreak tool & Financial NLP & Open financial AI models & Models and data frameworks & Text and sentiment benchmarks \\
		InvestorBench\newline\citep{investorbench_2024} & Benchmark & Trading decisions & Investor persona evaluation & Simulated decision agents & Performance across investor styles \\
		\citet{llm_agent_financial_trading_survey_2024} & Survey & LLM trading agents & Trading agent architectures & Agent modules, inputs, challenges & Backtest patterns and open questions \\
		\citet{llm_agents_finance_survey_2025} & Survey & LLM agents in finance & Broad financial applications & Multi-role workflows and deployment & Research, advisory, and compliance \\
		\bottomrule
	\end{tabularx}
	\vspace{4pt}
	\parbox{\linewidth}{\footnotesize \textit{Note:} Categorizations describe each work's primary organizing scope and core objectives. Benchmarks (FinBen, InvestorBench) and foundation frameworks (FinGPT) establish essential task-level foundations, while complementary surveys focus on general agent principles \cite{wang2024llmagentsurvey}, asset-pricing econometrics \cite{giglio2022factormodels}, trading agent fundamentals \cite{llm_agent_financial_trading_survey_2024}, and broad financial application domains \cite{llm_agents_finance_survey_2025}.}
\end{table*}

\textbf{General Autonomous Agent Architectures.}
Foundational surveys such as \citet{wang2024llmagentsurvey} provide overarching
taxonomies for LLM-based autonomous agents across general domains, structuring
the field around agent profiling, memory systems, planning capabilities, and
action spaces. While essential for establishing general cognitive concepts,
these frameworks naturally do not address domain-critical financial realities
such as order execution semantics, market frictions, slippage, and temporal
causality. The present survey builds upon these general agent concepts and
specializes them to the financial market environment.

\textbf{Financial Machine Learning and Econometric Factor Modeling.}
Methodological surveys in quantitative finance, such as \citet{giglio2022factormodels},
synthesize machine learning in asset pricing, focusing on expected return
estimation, factor discovery, stochastic discount factor modeling, and
econometric testing. These works provide statistical and theoretical grounding
for alpha discovery. In contrast, our survey centers on the \textit{agentic}
dimension---how automated models transition from statistical predictors into
active, autonomous decision-makers that manage portfolios and execute orders.

\textbf{Financial Foundation Models and Standardized Task Benchmarks.}
Initiatives such as FinBen \cite{finben_2024}, PIXIU \cite{pixiu_2023}, and
InvestorBench \cite{investorbench_2024}, alongside open models like
FinGPT \cite{fingpt_2023}, provide vital infrastructure and standardized
benchmarks for evaluating LLMs on financial NLP and isolated forecasting tasks.
Our survey complements these task-level benchmarks by examining end-to-end,
closed-loop agent systems where perception, reasoning, and execution operate
together in live or simulated trading environments.

\textbf{Existing Surveys on LLM Agents in Finance and Trading.}
Recently, \citet{llm_agent_financial_trading_survey_2024} presented a pioneering
survey on LLM agents in financial trading, systematically categorizing agent
architectures, data inputs, and broad backtesting performance. Concurrently,
\citet{llm_agents_finance_survey_2025} provided an extensive survey of LLM
agents across diverse financial sectors, highlighting real-world deployment
scenarios across research, advisory, and market operations.
Our survey is designed to be deeply complementary to these works: rather than
focusing solely on broad financial workflows or individual algorithmic setups,
we offer an in-depth, architecture-centric decomposition of the end-to-end
trading loop, systematically analyze execution and risk mechanisms, and provide
a structured synthesis of empirical evaluation practices across the literature.


\subsection{A Conceptual Framework: Disentangling Architecture, Capability, and Adaptation}
\label{ssec:taxonomy-framework}

To analyze the wide diversity of trading agents cohesively, this survey
introduces the \textbf{Architecture--Capability--Adaptation (A-C-A)} framework
as an analytical lens. By separating \textit{how systems are constructed}
from \textit{what financial objectives they pursue} and \textit{how they evolve},
the A-C-A lens clarifies design choices across three orthogonal dimensions:

\textbf{Dimension 1: Architectural Mechanisms (How the System is Built).}
The architectural layer deconstructs an agent into its underlying engineering
blocks: multimodal perception (text, time-series, charts), memory hierarchies
(working memory, associative vector storage, structured knowledge bases),
cognitive reasoning paradigms (reactive, reflective, planning-based), and
action execution modules (order slicing, venue routing, latency and cost modeling).
This decomposition identifies the inputs, retained state, decision procedures,
and execution assumptions that can be compared across systems
(Part~I, \Cref{sec:perception,sec:memory,sec:reasoning,sec:action}).

\textbf{Dimension 2: Functional Capabilities (What the System Accomplishes).}
The capability layer describes the financial tasks executed by the agent,
including automated alpha generation (symbolic factor discovery and mathematical
expression mining), portfolio allocation (multi-asset weighting and dynamic
rebalancing), and proactive risk management (drawdown controls, stop-loss limits,
and supervisory compliance checks). Distinguishing capabilities from architectures
clarifies that diverse architectural setups (e.g., reactive vs. strategic) can
be deployed across identical financial goals with differing operational trade-offs
(Part~II, \Cref{sec:alpha,sec:portfolio,sec:risk}).

\textbf{Dimension 3: Adaptive Pathways (How the System Evolves).}
The adaptation layer characterizes changes to agent behavior, parameters,
and organization over time. It covers in-context learning, supervised
fine-tuning, reinforcement learning, and multi-agent coordination and adaptive
interactions. Self-evolution concerns agent-managed updates retained across
decisions; structural self-evolution is the subset that changes strategy code,
update procedures, or system organization. Memory consolidation and reflection
can support persistent adaptation within an otherwise fixed architecture
(Part~III, \Cref{sec:learning,sec:multiagent,sec:evolution}).

\Cref{tab:taxonomy-comparison} illustrates how representative trading systems
in the literature map across this three-dimensional A-C-A framework, highlighting
how distinct architectural building blocks combine with functional goals and
adaptation mechanisms. \ref{app:aca-reclassification} provides additional
system-level mappings.

\begin{table*}[!tb]
	\centering
	\small
	\setlength{\tabcolsep}{5pt}
	\renewcommand{\arraystretch}{1.18}
	\caption{Illustrative mapping of representative trading systems under the A-C-A conceptual framework, showing how architectural components, functional capabilities, and adaptation pathways intersect.}
	\label{tab:taxonomy-comparison}
	\begin{tabularx}{\linewidth}{@{}p{2.6cm} p{4.2cm} p{4.5cm} X@{}}
		\toprule
		\textbf{System} & \textbf{Architecture Dimension} & \textbf{Capability Dimension} & \textbf{Adaptation Pathway} \\
		\midrule
		FinAgent \cite{wu2024finagent} & Multimodal Perception (Text+Visual) + Execution Layer & Direct Stock Action Selection \& Position Execution & In-Context Learning from market feedback \\
		AlphaAgent \cite{alphaagent_2025} & Reflective/Strategic Reasoning + Feedback Validation & Automated Code-Based Alpha Discovery \& Factor Pruning & Hypothesis-Factor Iterative Refinement \\
		TradingAgents \cite{xiao2024tradingagents} & Hierarchical Role-based Multi-Agent Coordination & Joint Alpha Generation, Portfolio Sizing \& Risk Control & Hierarchical Feedback \& Memory Updating \\
		\bottomrule
	\end{tabularx}
\end{table*}


\subsection{Contributions and Organization}
\label{ssec:contributions}

This survey provides a comprehensive and structured synthesis of the agentic
trading literature. The primary contributions are:

\begin{enumerate}
	\item \textbf{The A-C-A Analytical Framework.} We organize system
	      comparisons by architecture, financial capability, and adaptation.
	      \Cref{tab:taxonomy-comparison} and the illustrative mappings in
	      \ref{app:aca-reclassification} show how the lens applies to
	      representative systems.

	\item \textbf{A Closed-Loop Mechanism Synthesis.} Parts~I--III connect
	      perception, memory, reasoning, and execution with alpha discovery,
	      portfolio management, risk control, and adaptation. The technical
	      chapters describe module interfaces, design alternatives, and
	      operational trade-offs across this decision loop.

	\item \textbf{Transparent Literature Mapping.} Based on
	      \EvidenceIncludedN~systematically mapped studies, with a formal search window of 2022--2026 and earlier foundational references
	      (including a core empirical subset of $n=\EvidencePrimaryN$~closed-loop
	      trading systems), we link representative mechanisms to their
	      literature context. \Cref{sec:protocol,tab:selection-flow-summary}
	      document the selection boundary and corpus composition.

	\item \textbf{Empirical Evaluation and Methodological Landscape.} We
	      characterize reported evaluation practices in the empirical
	      subset across temporal
	      splitting (\ProtocolSplitReportedN/\EvidencePrimaryN), transaction-cost
	      modeling (\ProtocolCostReportedN/\EvidencePrimaryN), survivorship-bias
	      controls (\ProtocolUniverseReportedN/\EvidencePrimaryN), and execution
	      semantics (\ProtocolExecutionReportedN/\EvidencePrimaryN).
	      \Cref{tab:cohort-protocol-coverage} reports the initial and extension
	      cohorts separately to contextualize the pooled counts.

	\item \textbf{Forward-Looking Challenges and Best-Practice Guidelines.} We
	      identify key engineering and financial bottlenecks (including latency,
	      market impact, hallucination cascades, and strategy degradation) and
	      connect them to methodological recommendations (MR-1--MR-7).
	      \Cref{sec:challenges,tab:mr-matrix} organize these directions and
	      the recommendations' applicability by study type.
\end{enumerate}


The remainder of this survey is organized as follows.
\Cref{sec:protocol} details the literature selection protocol and synthesis scope.
\textbf{Part I: Architecture} (\Cref{sec:perception,sec:memory,sec:reasoning,sec:action})
deconstructs the core architectural pillars: perception, memory, reasoning, and action execution.
\textbf{Part II: Capability} (\Cref{sec:alpha,sec:portfolio,sec:risk})
examines the primary financial capabilities: alpha discovery, portfolio optimization, and risk control.
\textbf{Part III: Adaptation} (\Cref{sec:learning,sec:multiagent,sec:evolution})
explores how agents learn, collaborate, and evolve.
\Cref{sec:challenges} discusses open challenges and methodological guidelines,
and \Cref{sec:conclusion} concludes the survey.


We begin by defining what constitutes an agentic trader, drawing on cognitive
science and artificial intelligence literature to identify the essential
architectural components.

\label{part:architecture}

Part I addresses the fundamental question: \textit{What constitutes an agentic
	trader?} We characterize an agentic trader by four core architectural
components---perception, memory, reasoning, and action/execution---that together support
autonomous decision-making in financial markets. Whereas traditional trading systems focus
primarily on prediction, agentic systems typically include an explicit cognitive
loop that interacts with and learns from the market environment through decisions
mapped to tradable actions under market constraints.


\section{Review Protocol and Evidence Scope}
\label{sec:protocol}

\newcommand{\PrimarySubsetN}{80}
\providecommand{\EvidencePrimaryN}{80}

To synthesize this rapidly expanding literature with conceptual clarity and
methodological rigor, this review adopts a structured literature selection and
corpus mapping methodology. We define transparent inclusion boundaries to
differentiate end-to-end trading agent implementations from broader financial
tools, and map the design space along well-defined
architectural, functional, and empirical evaluation dimensions.

The formal search for contemporary trading-agent work covered records published from \textbf{2022-01-01}
through \textbf{2026-09-30}; the 2022 start date marks the period in which
LLM-based systems began to appear as end-to-end trading agents rather than as
isolated NLP or forecasting components. Citation chaining also retained nine
pre-2022 foundational background works, including classical portfolio and
learning methods; these references belong to the included ledger but fall
outside the contemporary search window. The initial-window search covered 2022-01-01
through
\textbf{2026-03-09} (last query date: 2026-03-09), using ACM
Digital Library (including ICAIF), IEEE Xplore, arXiv, SSRN, and Google
Scholar, complemented by backward and forward citation chaining. For this
revision we ran a pre-specified extension window over \textbf{2026-03-10 to
2026-09-30} on arXiv reusing the three initial query families and adding trading-agent,
multi-agent, and alpha-mining patterns
(agentic trading, LLM finance agents, evaluation/validity, trading-agent
phrases, multi-agent markets, and alpha mining); the extension retrieved
1{,}539 unique records, of which \WindowExtensionRecordsN{} entered
eligibility assessment under the same inclusion boundary
(\ref{app:search}). After deduplication and screening, the current ledger
contains \EvidenceIncludedN{}~included studies from \CandidateRecordsN{}~candidate
records. The windows are not matched search samples: the initial window spans
multiple databases and citation chaining, whereas the extension uses arXiv
only. Window-level differences therefore describe the assembled ledger and
should not be interpreted as a controlled comparison of search coverage.

To make the review boundary reproducible, we distinguish three fixed
denominators throughout the manuscript. The \emph{candidate registry}
contains \CandidateRecordsN{}~unique records after deduplication. The
\emph{included evidence-mapping ledger} contains \EvidenceIncludedN{}~records
retained for either primary evidence or background/design context. The
\emph{primary empirical subset} contains \EvidencePrimaryN{}~records that
satisfy all three conditions: Action Output, Closed-Loop Evaluation, and an
eligible empirical evaluation type (\texttt{backtest}, \texttt{simulation},
\texttt{live}, or \texttt{benchmark});
all protocol-field and reproducibility percentages use this
\EvidencePrimaryN-record subset unless explicitly stated otherwise;
the design-theme synthesis draws on representative primary and background work.
\Cref{tab:selection-flow-summary} summarizes the resulting selection flow.

\begin{table*}[t]
		\caption{Study selection summary using the canonical candidate registry, combining the initial-window ledger (cutoff 2026-03-09) with the extension window (2026-03-10 to 2026-09-30). The denominator is the \RecordsAfterDedupN-record deduplicated registry. Final exclusions are outside the survey domain; finance-relevant records that lack Action Output, Closed-Loop Evaluation, or eligible evaluation type remain in the included background tier. Detailed logs and per-record decisions are provided in Supplementary Material S1.}
	\label{tab:selection-flow-summary}
	\small
	\centering
	\setlength{\tabcolsep}{6pt}
	\renewcommand{\arraystretch}{1.1}
	\begin{tabular}{l r}
		\toprule
		\textbf{Stage}                                        & \textbf{Count}             \\
		\midrule
		Registry candidate records\textsuperscript{a}         & \CandidateRecordsN         \\
		Included in evidence mapping set                      & \EvidenceIncludedN         \\
		Excluded at screening/eligibility                     & \CandidateExcludedN        \\
		\midrule
		\textit{Final exclusion reason}     &                            \\
		\hspace{1em}Outside finance/trading/portfolio/risk-management scope & \ExcludedOutOfScopeN       \\
		\midrule
		\textit{Included evidence-map partition}     &                            \\
		\hspace{1em}Primary empirical subset         & \EvidencePrimaryN         \\
		\hspace{1em}Background/context tier          & \EvidenceBackgroundN      \\
		\bottomrule
	\end{tabular}
	\begin{flushleft}
		\textsuperscript{a}\footnotesize{The canonical registry currently contains \RecordsAfterDedupN{}~unique records after deduplication (\EvidenceIncludedN{}~included in the evidence mapping set and \CandidateExcludedN{}~excluded at screening/eligibility). The included evidence mapping set is partitioned into \EvidencePrimaryN{}~primary empirical studies and \EvidenceBackgroundN{}~background/context records.}
	\end{flushleft}
\end{table*}

The inclusion boundary for the \emph{primary empirical subset} is deliberately
restrictive: a study must satisfy all three conditions: \emph{(i) Action
Output}, meaning that the system emits tradable actions such as orders, position
changes, or portfolio allocations; \emph{(ii) Closed-Loop Evaluation}, meaning
that those actions are evaluated in backtesting, simulation, or live trading;
and \emph{(iii) an eligible evaluation type} coded as \texttt{backtest},
\texttt{simulation}, \texttt{live}, or \texttt{benchmark}. The third condition
keeps narrative-only descriptions outside empirical denominators even when
action output and a closed-loop claim are coded present. In the current S0 v2
registry, 102 included records carry the raw \texttt{primary} role tag; 22 of
these are narrative-only and are assigned to the background tier for empirical
denominators. Thus the 80-study primary empirical subset plus 105
background/context records partitions the 185 included studies. Under this
rule, the present snapshot yields
\EvidencePrimaryN{}~primary empirical studies and \EvidenceBackgroundN{}~background
studies.

The screening workflow proceeded in four steps. First, database and
preprint searches were run with agent/trading/finance keyword families and then
expanded through backward and forward citation chaining. Second, duplicate
records were removed by title, arXiv identifier, DOI, and manual inspection.
Third, title/abstract screening removed records outside financial trading,
portfolio construction, execution, or risk-management settings. Fourth,
full-text screening assigned included records to the empirical primary subset
only when all three criteria above were met; finance-relevant records failing
any criterion were retained as background/context, not final exclusions.
Accordingly, the initial window contributes 19 primary empirical and 58
background records, while the extension contributes 61 and 47, respectively.
These cohorts also differ in coding depth: the initial-window coding was
followed by a targeted extraction audit, whereas the extension cohort was coded
in a single pass with focused full-text extraction of protocol paragraphs.
\Cref{tab:cohort-protocol-coverage} reports both cohorts alongside pooled
coverage. Search composition and coding depth can both affect these rates;
the ledger does not identify their separate contributions.

\begin{table*}[!tb]
\centering\small
\caption{Protocol-field coverage by retrieval and coding cohort. Each cell gives the number reporting the field, cohort denominator, and percentage. The initial cohort received a targeted extraction audit; the extension received single-pass coding. Differences are descriptive and combine search composition with coding depth.}
\label{tab:cohort-protocol-coverage}
\begin{tabular}{lccc}
\toprule
Field & Initial ($n=19$) & Extension ($n=61$) & Pooled ($n=80$) \\
\midrule
Time split & 2/19 (10.5\%) & 48/61 (78.7\%) & 50/80 (62.5\%) \\
Transaction costs & 1/19 (5.3\%) & 46/61 (75.4\%) & 47/80 (58.8\%) \\
Universe / survivorship & 1/19 (5.3\%) & 47/61 (77.0\%) & 48/80 (60.0\%) \\
Execution semantics & 11/19 (57.9\%) & 36/61 (59.0\%) & 47/80 (58.8\%) \\
\bottomrule
\end{tabular}
\end{table*}

The technical chapters synthesize architectural scope through representative
systems and the illustrative A-C-A notes in \ref{app:aca-reclassification}.
The current registry does not contain populated A-C-A assignments for all
80 empirical studies, so it supports corpus-level protocol statistics rather
than numerical chapter-level taxonomy prevalence. Protocol-critical fields
include time-consistent split
discipline, execution timing or semantics, transaction-cost modeling,
universe/survivorship handling, and reproducibility artifacts. Missing or
unclear reporting is coded conservatively as \emph{NR/Unknown} rather than
imputed. This structured coding enables a transparent, multifaceted synthesis that
contextualizes reported findings with their respective execution realism,
temporal validity controls, and methodological transparency.

We audited the screening stage on a stratified double-screening sample
of 182 raw screening records in Supplementary Material S1. Because that sample
predates the final initial-window canonical registry, we retain it only as a
screening-boundary quality check; it does not inform extraction-level protocol
fields, reproducibility tiers, or A-C-A
classification robustness. To address the more consequential coding decisions,
we added a targeted extraction-level audit over \ExtractionAuditSampleN{}~fixed-seed
initial-window records sampled from the legacy 41-record primary-role frame
(seed = 20260426). Seven of the 15 sampled records are narrative-only under the
current eligible-evaluation-type rule. This audit therefore checks coding
agreement on its sampled fields; it is not a representative reliability estimate
for the current 80-record primary empirical subset. The audit covers action output,
closed-loop evaluation, evidence role, evaluation type, split protocol, cost
model, execution semantics, universe/survivorship handling, artifacts, and
R0--R3 reproducibility (\Cref{tab:reproducibility-summary}). The per-study coding sheet and adjudication notes are
provided in Supplementary Material S3. The extension-window cohort
(\WindowExtensionRecordsN{} records) was coded in a single pass from
abstract-level screening plus focused full-text extraction of protocol
paragraphs; its per-record decisions are released in the v2 registry, and its
field-level codings should be read as more preliminary than the audited
initial-window subset. \Cref{tab:screening-reliability} summarizes the two
audit layers and their scope.

\begin{table*}[!tb]
	\centering
	\footnotesize
	\setlength{\tabcolsep}{3pt}
	\renewcommand{\arraystretch}{1.12}
	\caption{Reliability and coding-quality audits reported with the submission package.}
	\label{tab:screening-reliability}
	\begin{tabularx}{\linewidth}{@{}>{\RaggedRight\arraybackslash}p{2.7cm}
		>{\RaggedRight\arraybackslash}p{2.1cm}
		>{\RaggedRight\arraybackslash}p{2.6cm}
		>{\RaggedRight\arraybackslash}p{3.0cm}
		>{\RaggedRight\arraybackslash}X@{}}
		\toprule
		\textbf{Audit layer} & \textbf{Sample} & \textbf{Observed agreement} & \textbf{Agreement note} & \textbf{Scope} \\
		\midrule
		S1 screening-boundary audit & 182 raw records & Reported in S1 & Boundary only & Inclusion/exclusion and screening-stage labels \\
		S3 extraction-level audit & \ExtractionAuditSampleN{}~legacy primary-role records & All \ExtractionAuditFieldComparisonsN{} field comparisons agreed pre-adjudication (\ExtractionAuditAgreementPct\%) & \ExtractionAuditKappaNote & Extraction fields used for protocol and reproducibility statistics \\
		\bottomrule
	\end{tabularx}
\end{table*}

A source check on 2026-10-03 corrected Alpha$^2$ from R2 to R1 because its
official repository provides pseudocode rather than a runnable implementation.
The journal version of the explainable portfolio-policy study also links public
code, correcting its old R0 assignment to R1 \citep{escudero2024explainabl}.
Its complete runnable reproduction is not established here.
The current registry and distributions incorporate these amendments; historical
S3 records are retained as coding snapshots. Neither amended work was in the 15-record
reliability sample, so the recorded 150/150 agreement is unchanged.

The A-C-A framework organizes representative trading systems by architecture,
financial capability, and adaptation. Corpus-level descriptive statistics for protocol
reporting, reproducibility, and peer-review status are summarized in
\ref{app:evidence-annotation-summary}; boundary decisions and final exclusions
are summarized in \ref{app:excluded-studies}. Detailed query strings,
initial-window screening logs, the combined registry, current protocol tables,
targeted reliability-audit files, and coding notes are provided in
Supplementary Materials S1--S3. Historical mapping and coding snapshots are
identified separately from the current registry and protocol table.

\subsection{Agentic Trading as an Expert-System Decision Pipeline}
\label{ssec:expert-system-pipeline}

Within ESWA's expert-systems framing, we interpret LLM-based trading agents as
expert-system decision pipelines, which makes the evaluation problem
explicit. Perception modules perform knowledge acquisition from prices, news,
filings, social media, and market microstructure signals. Memory,
retrieval-augmented generation, and tool-mediated context stores form a
knowledge base whose provenance and update rules must be inspectable. Reasoning
modules act as inference engines that convert state representations into
hypotheses, rationales, and candidate decisions. Action and execution modules
form the external action interface, where target weights, orders, fills,
latency assumptions, and market frictions must be made explicit. Risk controls,
compliance filters, and portfolio limits serve as rule constraints on the path
from inference to action. Finally, logs, immutable artifacts, and human
oversight provide the explanation and accountability layer expected of
decision-support systems.

This expert-system framing provides a cohesive conceptual spine for our synthesis.
A trading agent's real-world viability depends not only on its conceptual design
but also on the rigor of its empirical evaluation---including temporal split integrity,
transaction-cost realism, universe survivorship controls, and execution fidelity.
The A-C-A framework organizes this multifaceted design space systematically:
architecture describes how market knowledge is acquired, stored, inferred, and acted
upon; capability delineates which core financial functions are realized; and
adaptation examines how the decision pipeline evolves under non-stationary market feedback.


\section{Perceptual Architecture}
\label{sec:perception}


The preceding section introduced the cognitive architecture of agentic traders.
This section addresses the question: \textquotedblleft How do agents perceive
financial markets?\textquotedblright

In this survey, we define \textbf{Perception} as the mechanism by which an agent acquires a state observation $O_t$ from the environment state $S_t$ at decision time $t$. Under the primary-subset boundary specified in \Cref{sec:protocol}, standalone financial NLP or time-series forecasting models (e.g., isolated BERT or LSTM predictors) are treated as \textbf{Background (BG)} unless they form, or are explicitly integrated into, an evaluated action-producing loop: component models lack the closed-loop integration required for agentic classification, although they may still supply perceptual capabilities that end-to-end agents incorporate.

A perception module must therefore do more than predict; it must provide actionable state representations---such as sentiment scores, market regime indicators, or visual summaries---that the agent's reasoning component can use to formulate valid actions.

\paragraph{Architectural Role and Design Dimensions}
In an autonomous trading agent, perception functions as the critical observation interface between external market feeds and internal cognitive reasoning. Rather than functioning as an isolated prediction engine, the perception layer in an end-to-end agentic architecture must construct a structured, temporally synchronized observation $O_t$ that is immediately actionable for memory retrieval, factor computation, and order formulation. Current literature explores this design space across three core modalities:
\begin{itemize}
	\item \textbf{Textual Perception} (\Cref{ssec:text}): Parsing unstructured financial news, earnings announcements, filings, and social sentiments.
	\item \textbf{Time-Series Perception} (\Cref{ssec:timeseries}): Encoding numerical sequences, including order book microstructure, price trends, and macroeconomic fundamentals.
	\item \textbf{Multimodal Perception} (\Cref{ssec:multimodal}): Aligning and fusing graphical representations (e.g., candlestick charts, technical patterns) with textual and numerical signals.
\end{itemize}
Across these modalities, the central design trade-offs revolve around feature representation density, cross-modal synchronization, and latency constraints, balanced against the critical requirement of preventing temporal information leakage.

\begin{figure*}[tbp]
	\centering
	\includegraphics[width=0.95\linewidth]{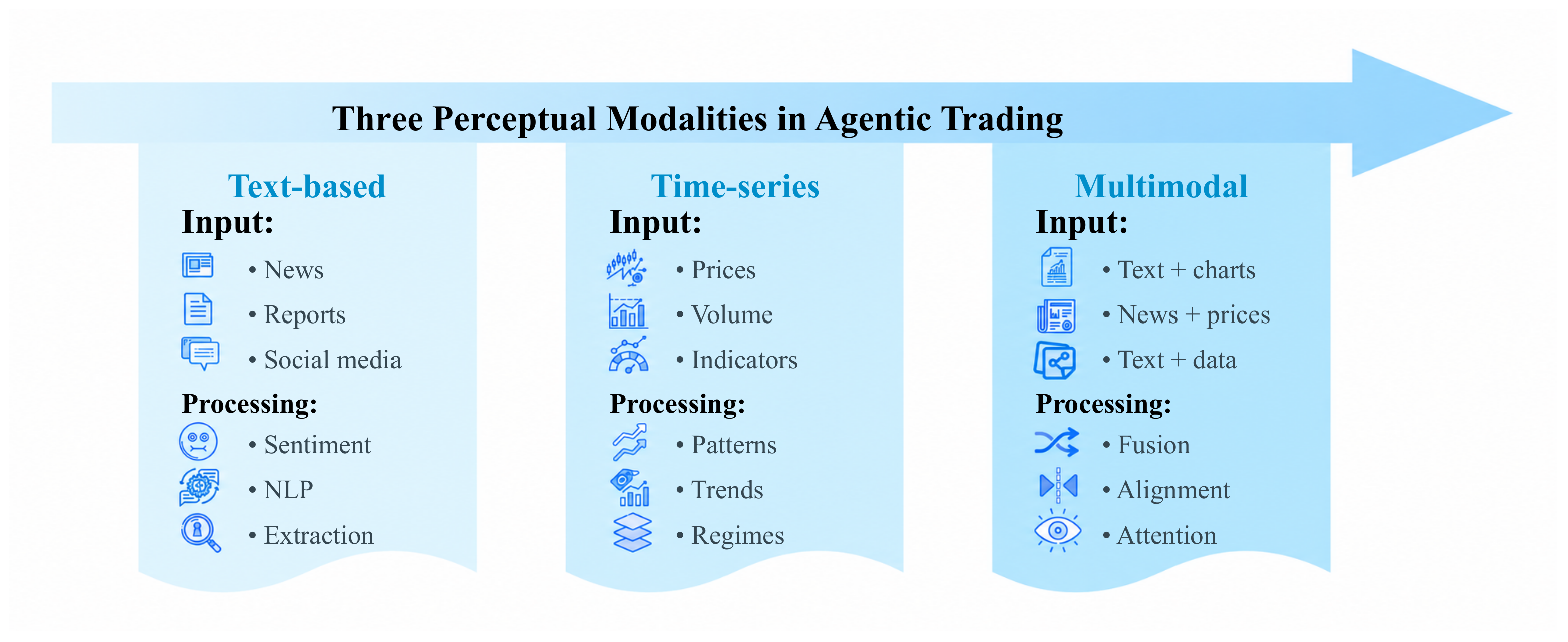}
	\caption{Three perceptual modalities in agentic trading. Textual perception processes linguistic information from news and reports. Time-series perception analyzes numerical patterns in price and volume data. Multimodal perception integrates heterogeneous data sources through cross-modal fusion mechanisms.}
	\label{fig:perceptual-modalities}
\end{figure*}

\Cref{fig:perceptual-modalities} compares three complementary perceptual modalities by information source and processing mechanism.

\subsection{Textual Perception}
\label{ssec:text}

Textual perception refers to the acquisition of state observations $O_t$
from linguistic information sources (news articles, earnings reports, analyst
commentary, social media posts, regulatory filings) at decision time $t$. The
module encodes textual inputs into representations---such as sentiment scores,
event embeddings, or contextualized vectors---that the reasoning component can
use to formulate actions.

\textbf{Representative Approaches}. Several lines of work have developed
specialized architectures for financial text understanding.
\citet{du2024financial} provide a comprehensive survey of financial sentiment
analysis techniques, reviewing prior evidence that sentiment signals are often
associated with subsequent market movements.
\citet{iacovides2024finllama} introduce FinLlama, an LLM-based approach
fine-tuned for financial sentiment analysis in algorithmic trading scenarios.
\citet{instruct_fingpt_2023} propose Instruct-FinGPT, which uses
instruction tuning to adapt general-purpose LLMs to financial sentiment tasks,
while \citet{zhang2023enhancing} explore retrieval-augmented
generation to enhance sentiment analysis with external knowledge.

\textbf{Technical Mechanisms}. Textual perception typically follows a
pipeline of tokenization, encoding, and extraction. Domain-specific language
models such as FinBERT \citep{araci2019finbert} provide contextualized
representations of financial text, capturing terminology and semantic nuances
that are often underrepresented in general-purpose models.
\citet{rodriguez2024large} show that adapting large language models through
task-specific fine-tuning can improve performance on financial sentiment tasks.
Sentiment classification models then map these representations to polarity labels
(e.g., positive, negative, and neutral) or to more fine-grained affective states.
\citet{xing2024designing} further studies heterogeneous LLM agents
for financial sentiment analysis and reports improved benchmark performance
relative to selected baselines.

\textbf{Critical Challenge: Temporal Alignment}. A pervasive risk in text
perception is look-ahead bias. News timestamps often record publication time
rather than the moment an item actually becomes available to a trading
pipeline, creating a mismatch between historical records and executable decision
contexts; agents trained on historical news corpora without explicit
ingestion-lag modeling or embargo enforcement are therefore likely to exhibit
inflated backtest performance. A related risk arises from post-publication
revisions: some financial disclosures and structured reports are updated after
initial release, so using retrospectively cleaned or final versions in
backtesting can introduce serious data leakage \cite{lopezprado2018dangers}.

\Cref{tab:text-perception} summarizes representative textual perception
methods, highlighting the diversity of data sources and technical approaches.
Because these methods are effective on textual information but cannot directly
process numerical price data or visual chart patterns, agents also require the
complementary perceptual modalities discussed next.

\begin{table*}[!ht]
	\centering
	\caption{{Textual perception methods. \textbf{Audit Status}: \texttt{BG}=Background Model (not integrated into agent loop); \texttt{R0--R3}=Reproducibility Level (see \ref{app:evidence-annotation-summary}). \textbf{Time Disc.}=Time Discipline (\texttt{Static}=offline batch processing; \texttt{RT}=real-time streaming with specified latency).}}
	\label{tab:text-perception}
	\scriptsize
	\setlength{\tabcolsep}{2pt}
	\begin{tabularx}{\linewidth}{@{}l l X l l@{}}
		\toprule
		Reference                                     & Input Source     & Representation            & Time Disc. & Audit       \\
		\midrule
		FinBERT \cite{araci2019finbert}               & News Headlines   & Sentiment Score (Scalar)  & Static     & \texttt{BG} \\
		FinGPT \cite{fingpt_2023}                     & News/Social      & Summary/Embedding         & Static     & \texttt{BG} \\
		BloombergGPT \cite{bloomberggpt_2023}         & Financial Corpus & General LM                & Static     & \texttt{BG} \\
		FinLlama \cite{iacovides2024finllama}         & Reports          & Instruction Tuned         & Static     & \texttt{BG} \\
		Instruct-FinGPT \cite{instruct_fingpt_2023}   & Financial Text   & Instruction Tuned         & Static     & \texttt{BG} \\
		RAG-FinGPT \cite{zhang2023enhancing}          & Mixed Sources    & Retrieval-Augmented       & Static     & \texttt{R0} \\
		Heterogeneous Agents \cite{xing2024designing} & Multi-source     & Specialized Architectures & Static     & \texttt{R0} \\
		\bottomrule
	\end{tabularx}
\end{table*}

\subsection{Time-Series Perception}
\label{ssec:timeseries}

Time-series perception refers to the acquisition of state
observations $O_t$ from numerical sequences (price, volume, Limit Order Books---LOB) and structured fundamental data (e.g., balance sheet tables) at decision time $t$. This modality encodes temporal patterns---such as trend indicators, volatility estimates, or regime classifications---into representations the reasoning component can use. Its inputs range from familiar price and volume sequences (Open-High-Low-Close-Volume---OHLCV) to deeper market structures such as LOB and options-implied state variables. Because these inputs evolve over time, agents must model temporal dependence, seasonality, structural breaks, and non-stationarity when constructing actionable state representations.

\textbf{Representative Approaches}. Representative work in this area studies how
sequence models encode information from market data streams into trading-relevant
state representations. Recent systems use transformer-style or attention-based
encoders to summarize price dynamics, while classical econometric approaches
remain useful for regime identification and structural-change analysis. In the
current agentic trading literature, however, most implemented systems still rely
on readily accessible OHLCV-style inputs, while explicit use of high-frequency
LOB data and richly structured fundamentals remains limited.

\textbf{Technical Mechanisms}. Key challenges include non-stationarity, noise,
and sensitivity to temporal misalignment. Successful approaches often employ
attention mechanisms to focus on informative time points, regime-switching models
to capture structural breaks \cite{hamilton1989new,bai2003multiple}, and
representation-learning objectives that improve robustness under changing market
conditions. Standard positional encodings and related sequence-ordering schemes
can also help models capture periodic structure and temporal context in
sequential financial data \cite{vaswani2017attention}.

\begin{table*}[!tb]
	\centering
	\caption{Time-series and structured perception methods. Audit and Time Disc. tags follow the convention in \Cref{tab:text-perception}.}
	\label{tab:timeseries-perception}
	\scriptsize
	\setlength{\tabcolsep}{2pt}
	\begin{tabularx}{\linewidth}{@{}l l X l l@{}}
		\toprule
		Reference                               & Input Data                   & Mechanism                   & Time Disc. & Audit       \\
		\midrule
		VISTA \cite{khezresmaeilzadeh2025vista} & Chart Images + Text          & Zero-Shot VLM Inference     & Static     & \texttt{R0} \\
		Regime-Switching \cite{hamilton1989new} & Returns                      & Hidden Markov Model         & Static     & \texttt{BG} \\
		\bottomrule
	\end{tabularx}
\end{table*}

\Cref{tab:timeseries-perception} compares representative time-series perception
methods. Methods that operate on numerical sequences alone cannot directly
incorporate the rich contextual information available in text; vision-language
variants such as VISTA that render series as charts already illustrate the
cross-modal direction developed next.

\subsection{Multimodal Perception}
\label{ssec:multimodal}

Multimodal perception refers to the acquisition of state observations
$O_t$ from heterogeneous sources (chart images, text, numerical series, audio) at
decision time $t$, integrating these into unified representations through
cross-modal alignment mechanisms. This mirrors trading practice, in which traders synthesize visual technical patterns with news flow and price action.

\textbf{Representative Approaches}. For visual perception,
\citet{wang2023finvis} introduce FinVis-GPT for financial chart analysis, which
treats chart-like market representations as visual inputs amenable to
language-based interpretation. For multimodal fusion,
\citet{wu2024finagent} introduce FinAgent, which combines textual news and
numerical market information through a cross-modal pipeline designed for trading
decisions. More broadly, multimodal systems differ in where fusion occurs along
the pipeline, a distinction examined under Technical Mechanisms below.

\textbf{Critical Challenge: Cross-Modal Asynchrony}. Fusing modalities with
different update frequencies creates synchronization risks. News arrives
sporadically, whereas market and LOB data may update at much higher frequencies.
Naive framing strategies such as pairing the latest available news item with the
current price can therefore misrepresent the true decision context when the text
signal is stale. Moreover, visual interpretation and multimodal inference
introduce nontrivial latency, so the perceived market state may already be
outdated by the time the agent acts in live settings. Protocol reporting should
therefore state perception-latency assumptions and alignment procedures
explicitly.

\textbf{Technical Mechanisms}. Multimodal fusion poses significant challenges
because modalities differ in dimensionality, temporal resolution, and noise
characteristics. Early fusion concatenates raw features from all modalities
before processing, while late fusion maintains separate processing pipelines and
combines high-level representations. Cross-attention mechanisms provide a
flexible alternative by enabling one modality to attend selectively to another
during representation learning \cite{vaswani2017attention,tsai2019mult}. The
temporal-alignment problem noted above is especially acute in financial
settings, since textual events arrive discretely while prices update
continuously; preventing look-ahead bias therefore requires strict timestamp
discipline, explicit embargo handling, and clearly reported latency budgets
\cite{lopezprado2018dangers}.

\begin{figure*}[!tb]
	\centering
	\includegraphics[trim=50bp 185bp 30bp 155bp,clip,width=0.72\linewidth]{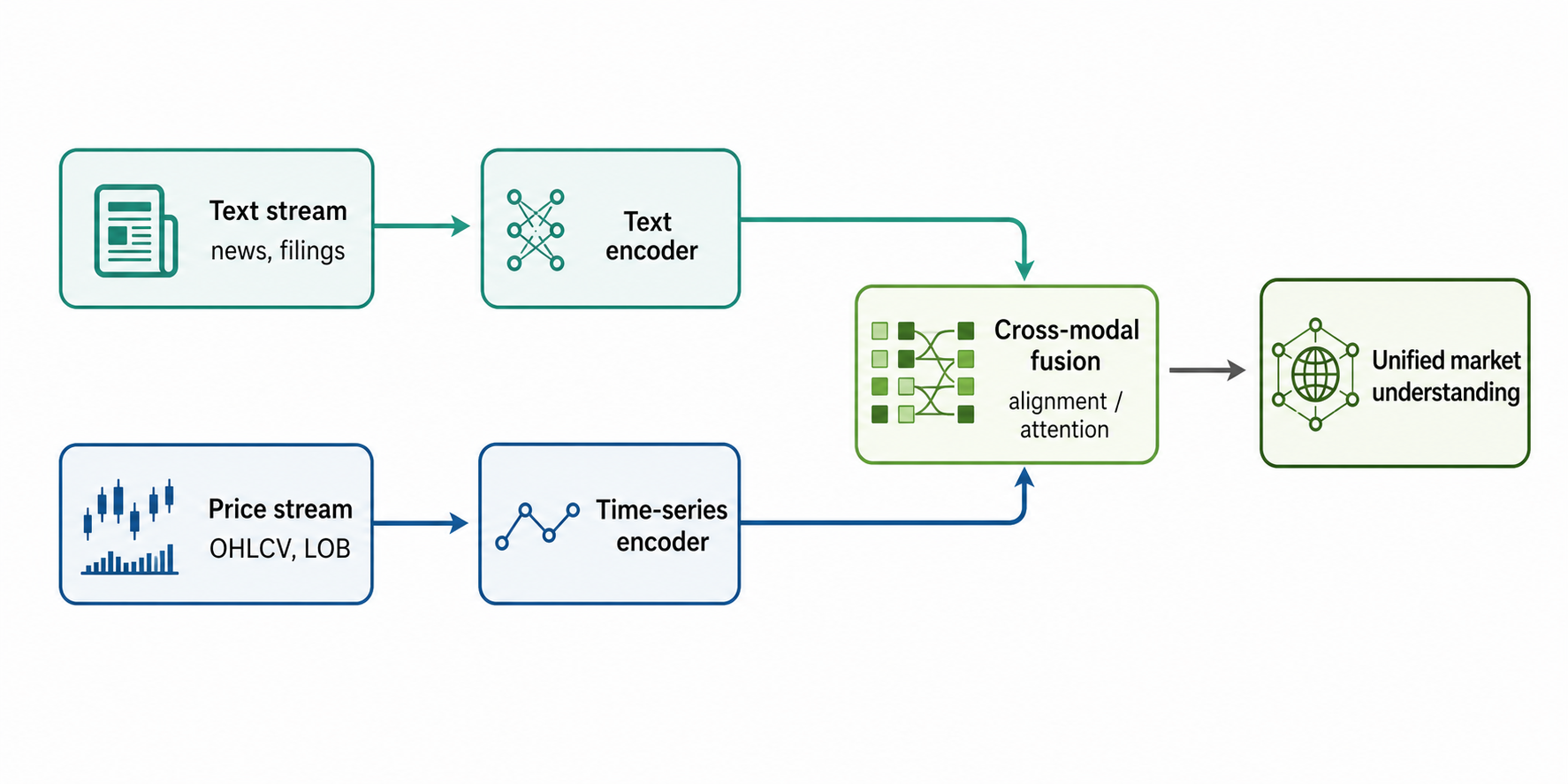}
	\caption{Multimodal fusion flow in agentic trading. The diagram illustrates
		how text encoders and time-series encoders process their respective inputs
		before a cross-modal fusion layer integrates the representations into a unified
		market understanding.}
	\label{fig:multimodal-fusion}
\end{figure*}

\Cref{fig:multimodal-fusion} illustrates the multimodal fusion pipeline, showing
how text encoders process news and reports while time-series encoders analyze price
movements, with a cross-modal fusion layer integrating these representations.

\begin{table*}[!tb]
	\centering
	\caption{Visual and multimodal perception methods. \textbf{Latency}: Estimated inference latency (critical for alignment). \textbf{Note}: Latency estimates are author-reported; see Supplementary Material S2 for detailed benchmarking protocols. Audit tags follow the convention in \Cref{tab:text-perception}.}
	\label{tab:multimodal-perception}
	\scriptsize
	\setlength{\tabcolsep}{2pt}
	\begin{tabularx}{\linewidth}{@{}l l X l l@{}}
		\toprule
		Reference                                  & Modalities    & Fusion Strategy          & Latency & Audit       \\
		\midrule
		FinVis-GPT \cite{wang2023finvis}           & Chart Images  & Vision-Language Analysis & Low     & \texttt{BG} \\
		FinAgent \cite{wu2024finagent}             & News + Prices + Kline Charts & Cross-Attention          & High    & \texttt{R0} \\
		Multimodal Transformer \cite{tsai2019mult} & Text + Audio + Vision  & Cross-Modal Attention    & Med     & \texttt{BG} \\
		\bottomrule
	\end{tabularx}
\end{table*}

\Cref{tab:multimodal-perception} summarizes representative visual and multimodal
perception approaches. Taken together, these studies suggest that richer market
state estimation often benefits from combining heterogeneous signals rather than
relying on a single modality in isolation.

\begin{table*}[!tb]
	\centering
	\scriptsize
	\setlength{\tabcolsep}{3pt}
	\caption{Summary of perceptual architectures in trading agents. The \textbf{Protocol Risk} column highlights specific alignment challenges for each modality.}
	\label{tab:perception-comparison}
	\begin{tabularx}{\linewidth}{@{}l p{3cm} p{3.5cm} X l@{}}
		\toprule
		\textbf{Modality} & \textbf{Primary Inputs}     & \textbf{Mechanism}                              & \textbf{Protocol Risk}                       & \textbf{Examples}    \\
		\midrule
		Textual           & News, Reports, Social Media & LLM Fine-tuning, Prompting, Sentiment Analysis  & Publication Lag, Embargo Violations          & FinBERT, FinGPT      \\
		Time-Series       & OHLCV, LOB, Fundamentals    & Temporal Attention, Regime Switching            & Look-ahead Bias, Data Revisions              & VISTA                \\
		Multimodal        & Charts, Text+Price+Audio    & Vision-Language Analysis, Cross-Modal Attention & Synchronization Latency, Causal Misalignment & FinVis-GPT, FinAgent \\
		\bottomrule
	\end{tabularx}
\end{table*}


\textbf{Extension-window additions.} The extension cohort adds perception-side
infrastructure rather than new modalities: mechanism-based decomposition of
market dynamics for downstream agents \cite{chen2026decomposing} and
synthesized financial tool-use data that couples perception to callable tools
\cite{huang2026fintoolsyn}. Both remain background-level designs in our ledger,
but they illustrate how the perception layer is being engineered as a service
for agent pipelines rather than as a standalone predictor.

\paragraph{Synthesis and Design Considerations}
Perceptual architecture turns heterogeneous market feeds into state observations $O_t$. Textual, time-series, and multimodal inputs differ in source and processing, while all require temporal alignment; multimodal fusion adds cross-modal synchronization, and each design must control look-ahead bias and latency. These observations feed the memory and retrieval systems examined next.



\section{Memory Architecture}
\label{sec:memory}

The preceding section introduced perceptual mechanisms for gathering market
information. This section asks how agents store, retain, and retrieve that
information. It focuses on memory as an organized state store; Learning
Paradigms examines how agents use feedback to update behavior, while
Self-Evolution examines how retained state and update policies change over time.

Memory serves three operational roles: \textbf{working memory} maintains active
trading context, \textbf{associative vector storage} retrieves historical
episodes by similarity, and \textbf{structured knowledge bases} preserve
explicit financial information. The core design challenges are balancing
retrieval relevance against context-window saturation, mitigating retrieval
drift, and maintaining temporal causality across long trading horizons.
These categories are examined in \Cref{ssec:working}, \Cref{ssec:episodic},
and \Cref{ssec:semantic}, respectively.
Earlier agent literature also uses
cognitive terminology: working memory follows \citet{baddeley1974working},
episodic and semantic memory follow \citet{tulving1972episodic}, and multi-store
models draw on \citet{atkinson1968human}. These cognitive descriptions provide
context for memory-augmented designs \cite{zong2024macrohft}; the categories
below identify the storage and retrieval mechanisms rather than biological
memory processes.

\begin{figure*}[!tb]
	\centering
	\includegraphics[width=\linewidth]{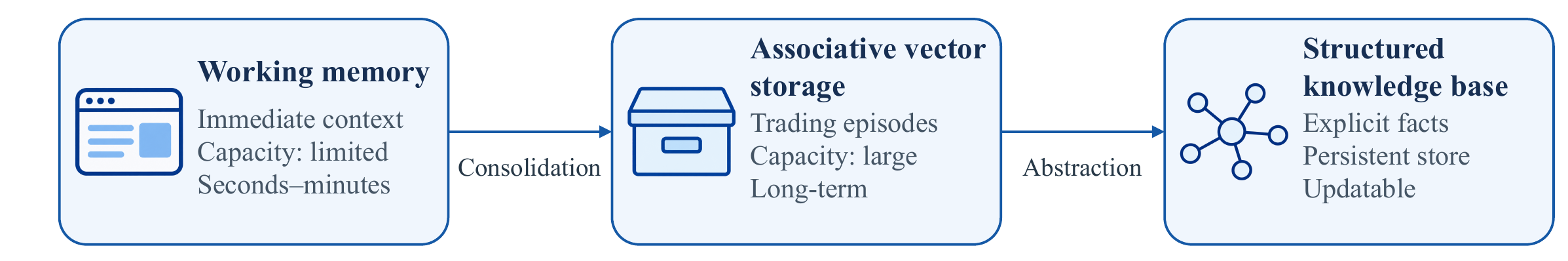}
\caption{Three memory categories in agentic trading. Working memory maintains
        active trading context, associative vector storage supports retrieval of
        historical trading episodes, and structured knowledge bases preserve
        explicit financial information. Arrows denote consolidation from working
        context into stored experience and abstraction from experience into reusable
        structured knowledge.}
	\label{fig:memory-systems}
\end{figure*}

\subsection{Working Memory}
\label{ssec:working}

In agentic trading, working memory is often conflated with a generic
\textquotedblleft context window.\textquotedblright For robust agent
design, we distinguish between the \textbf{Deterministic State Store} (Layer A)
and the \textbf{Generative Context} (Layer B).

\textbf{Layer A: Deterministic State Store (Audit-Truth)}. This layer holds the
ground-truth state of the trading system, including current positions, open
orders, account balances, and risk limits. In a reliable trading architecture, we
argue that it should be \textbf{read-only} to the LLM and updated solely by the
environment (e.g., EMS/OMS feedback). Treating this state as a
\textquotedblleft memory\textquotedblright subject to forgetting or hallucination
is a critical vulnerability; in practice, it is better implemented as a
structured store exposed through tools or state queries, as seen in tool-using
trading agent designs \cite{xiao2024tradingagents}.

\textbf{Layer B: Generative Context (Inference Scope)}. This layer corresponds to the LLM's active context window, containing the sliding history of observations, chain-of-thought reasoning traces, and recent reflections. Unlike Layer A, information here is transient and subject to eviction.

\textbf{Technical Mechanisms}. To manage Layer B's limited capacity, agents
employ eviction policies such as \textbf{Summary-Compression} (condensing old
steps into a narrative) or \textbf{hierarchical memory management} (using
layered storage and retrieval priorities to retain higher-value fragments)
\cite{packer2023memgpt,sun2025hmem}. FinMem \cite{yu2023finmem} adopts the
same layered pattern in a trading setting, buffering recent context in a
working layer atop its deeper memory layers; it is therefore listed under
working memory in \Cref{tab:memory-retrieval-comparison}. A key design
trade-off is the \textit{context-to-noise ratio}: maintaining too much history
can dilute attention and bury decision-relevant content in long contexts, a risk
consistent with the broader \textquotedblleft Lost in the Middle\textquotedblright
finding \cite{liu2023lostmiddle}, while aggressive pruning may drop reasoning
that is still relevant to a live position or pending order.

\begin{table*}[!tb]
	\centering
	\caption{Memory system comparison.}
	\label{tab:memory-comparison}
	\small
	\setlength{\tabcolsep}{3pt}
	\begin{tabularx}{\linewidth}{@{}l l l X@{}}
		\toprule
		\textbf{Memory Category} & \textbf{Capacity} & \textbf{Volatility} & \textbf{Key Characteristics}                            \\
		\midrule
		Working Memory       & Limited (Context) & Volatile            & Fast cache, eviction policies, sliding window           \\
		Associative Vector Storage & Large (Disk) & Persistent & Vector embeddings, similarity search, temporal indexing \\
		Structured Knowledge Base  & Source-dependent & Persistent & Knowledge graphs, sourced records, explicit relations \\
		\bottomrule
	\end{tabularx}
\end{table*}

\Cref{tab:memory-comparison} summarizes the three memory categories, highlighting their differing capacity constraints and temporal scales. Working memory enables rapid decision-making, while associative vector storage and structured KBs support longer-term retrieval and knowledge accumulation.

\subsection{Associative Vector Storage}
\label{ssec:episodic}

Associative vector storage provides a way to index and retrieve stored
experience through learned or engineered similarity. Trading episodes may be
represented this way, but not every episodic store uses vector embeddings. To
keep retrieved episodes auditable, we define a valid episode as a tuple $(S_t, A_t, O_{t+k}, \tau)$, where $S_t$ is the state,
$A_t$ the action, $O_{t+k}$ the outcome realized at step $t+k$, and $\tau$ the
timestamp at which this outcome becomes observable.

\textbf{Information Leakage via Memory}. We refer to a key vulnerability in
trading agents as the \textquotedblleft Oracle Fallacy,\textquotedblright in which
an agent retrieves a similar past episode that contains a \textit{post-hoc}
narrative (e.g., \textquotedblleft this trade failed due to news X released
tomorrow\textquotedblright).
\begin{tcolorbox}[title=Protocol Implication]
	To prevent such leakage, we suggest that reproducible trading-agent protocols enforce an
	\textbf{Outcome Embargo}: an episode recorded at $t$ cannot expose its outcome
	field to retrieval until current time $t_{now} \ge \tau$.
\end{tcolorbox}

Complementary work addresses deterministic replay in financial agents,
providing assurance frameworks for regulatory audit~\cite{khatchadourian2026replayable}.

\textbf{Representative Approaches}. FinAgent \citep{wu2024finagent} uses layered
memory and diversified retrieval, with temporal control primarily at the prompt
level rather than through explicit database constraints. FinMem
\citep{yu2023finmem} organizes memory in layers, while Reflexion
\citep{shinn2023reflexion} stores unsuccessful trajectories as lessons. These
examples illustrate storage and retrieval choices; their cross-time update
governance is considered in \Cref{sec:evolution}.

\textbf{Technical Mechanisms}. Vector databases
\cite{johnson2017billionscale,malkov2020hnsw} enable similarity search, but
naive cosine similarity is often insufficient for financial time series, where
relevance also depends on recency and regime. \textquotedblleft Time-Aware
Retrieval\textquotedblright therefore becomes a useful design principle.
\begin{tcolorbox}[title=Protocol Implication]
	We suggest a relevance score $f(q, k)$ that includes a decay term
	$e^{-\lambda(t_{now} - t_k)}$ to prioritize recent regimes.
\end{tcolorbox}
Furthermore,
\textbf{Drift Handling} is required: as market regimes shift (e.g., from low to
high volatility), old episodes with high feature similarity may still yield
misleading expectations, a risk consistent with recent retrieval-augmented
time-series forecasting evidence \cite{rag_financial_time_series_2025}.

\subsection{Structured Knowledge Bases}
\label{ssec:semantic}

Structured knowledge bases store explicit domain knowledge (e.g.,
\textquotedblleft high interest rates hurt growth stocks\textquotedblright).
Their auditability depends on provenance and update controls, not only on
format (graph vs. text). Model weights can also encode financial knowledge,
but that parametric knowledge is a related, implicit mechanism rather than a
structured KB: it is not an external source-addressable store and should not be
classified as one.

\textbf{Related mechanism: parametric knowledge (implicit)}. This knowledge is
encoded in model weights via pre-training or fine-tuning (e.g., FinGPT
\citep{fingpt_2023}). While efficient, it is \textbf{opaque}---it is difficult
to trace \textit{why} the model knows a fact---and \textbf{static}, since
updates usually require additional training or alignment. For
trading, this raises the risk of \textquotedblleft outdated priors,\textquotedblright
in which older market regularities persist in model behavior after the regime
that produced them has changed.

\textbf{Structured KBs (explicit, non-parametric)} retrieve information from
an external source. We distinguish:
\begin{itemize}
	\item \textbf{Curated KBs}: Verified documents (Rulebooks, SEC Filings) where source integrity is high \cite{li2025finsage}.
	\item \textbf{Uncurated KBs}: Open-web data (Social Media, Forums) susceptible to \textquotedblleft Noise Injection\textquotedblright or adversarial poisoning.
\end{itemize}

\textbf{Technical Mechanisms}. To preserve source integrity, \textquotedblleft
Source Tracking\textquotedblright is required: every retrieved
retrieved fact should carry a provenance log (Document ID + Version Timestamp).
Knowledge Graphs encode explicit entity-relation structure (e.g.,
$Entity_A \xrightarrow{supplier\_of} Entity_B$) and can therefore support
provenance-oriented reasoning and constraint checking more directly than
unstructured text alone.

\begin{figure*}[!tb]
	\centering
	\includegraphics[width=\linewidth]{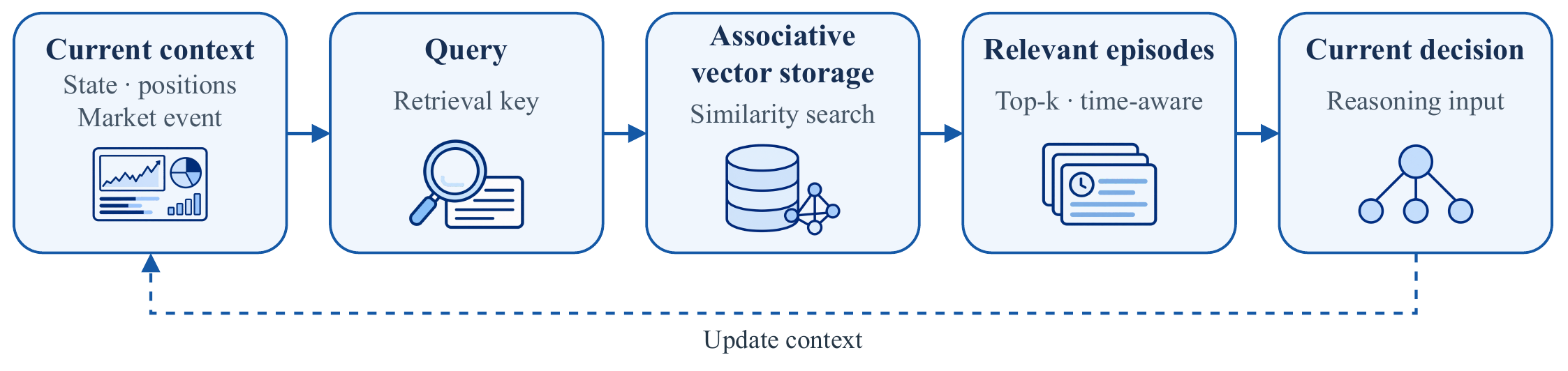}
	\caption{Memory retrieval mechanism. The agent formulates a query based on
		current context, searches an associative vector store using similarity-based
		retrieval, and selects relevant episodes to inform the current decision. }
	\label{fig:memory-retrieval}
\end{figure*}

\begin{table*}[!tb]
	\centering
	\small
	\setlength{\tabcolsep}{3pt}
	\caption{Comparison of memory retrieval mechanisms in trading agents. Tags follow the convention in \Cref{tab:text-perception} for the representative works in each row. NR = not reported; N/A = not applicable.}
	\label{tab:memory-retrieval-comparison}
	\begin{tabularx}{\linewidth}{@{}l X c c c c@{}}
		\toprule
		\textbf{Memory Category}   & \textbf{Representative Ref}    & \textbf{Timestamp Rule} & \textbf{Embargo} & \textbf{Drift Check} & \textbf{Audit Status} \\
		\midrule
		Working Memory                 & FinMem \cite{yu2023finmem}     & N/A       & N/A            & N/A & \texttt{R0} \\
		Episodic memory (paper term)   & FinAgent \cite{wu2024finagent} & NR        & NR             & No  & \texttt{R0} \\
		Parametric knowledge (related)& FinGPT \cite{fingpt_2023}      & Implicit  & No             & No  & \texttt{BG} \\
		Structured Knowledge Base      & FinSage \cite{li2025finsage}   & Versioned & Yes (Doc Date) & Yes & \texttt{BG} \\
		\bottomrule
	\end{tabularx}
\end{table*}


\subsection{Design Trade-offs}
\label{ssec:memory-tradeoffs}

A systematic architectural analysis of these systems reveals key design trade-offs that performance-only benchmarks tend to obscure.

\textbf{Latency vs. Recall Depth}. Deeper retrieval in high-fidelity RAG systems
can introduce latency that is incompatible with high-frequency or rapid
mid-frequency strategies. Agents therefore need to balance
\textit{retrieval depth} against \textit{decision latency} and
\textit{execution speed}.

\textbf{Context Length vs. Distraction}. Although newer LLMs support 1M+ token windows, loading all retrieved memories into context degrades reasoning through the \textquotedblleft Lost in the Middle\textquotedblright{} phenomenon \cite{liu2023lostmiddle}. A disciplined memory design therefore favors \textit{less} but \textit{higher-relevance} context over an unfiltered dump of history.

\textbf{Noise Injection Risk}. Allowing uncurated knowledge sources (e.g.,
scraping Reddit) increases the risk of \textquotedblleft Noise
Injection,\textquotedblright in which irrelevant or manipulated public sentiment
overrides the agent's underlying strategy. We therefore adopt a
\textquotedblleft Hierarchy of Truth\textquotedblright as a design principle:
internal Layer A state should override external noisy signals when the two
conflict.

\textbf{Extension-window additions.} The extension cohort contributes several
memory- and retrieval-facing designs. On the knowledge side, LLM-enhanced
dynamic financial knowledge graphs support cross-entity retrieval for trading
decisions \cite{zhang2026llm}, and a benchmark for event-grounded
personalized memory evaluates retention of investor-specific context
across sessions \cite{wang2026finperma}. On the retrieval side, document-routed hybrid
retrieval \cite{cheng2026sustainable} and time-series augmented generation
\cite{kolonin2026time} target the financial RAG layer that can feed vector
stores and structured KBs. A cautionary complement: information-fidelity analyses show
that aggressive summarization can distort the very inputs a memory layer is
meant to preserve \cite{lee2026when}---reinforcing the
\textquotedblleft Hierarchy of Truth\textquotedblright{} principle above.

\paragraph{Synthesis and Design Considerations}
Memory architecture provides temporal depth and experiential continuity: working memory tracks active trading state, associative retrieval selects historical episodes, and structured knowledge bases preserve explicit information. Retrieval depth must be balanced against latency and context distraction, while temporal rules and provenance checks protect against hindsight leakage and preserve source traceability. The retrieved context feeds the reasoning engine examined in \Cref{sec:reasoning}.



\section{Reasoning Architecture}
\label{sec:reasoning}

Having examined how agents perceive markets and store information, we now
address the question: \textquotedblleft How do agents make trading
decisions?\textquotedblright{} Building on the notation from
\Cref{sec:perception} and \Cref{sec:memory}, we write $O_t$ for the state
observation at time $t$ and $M_{assoc}$ for retrieved associative memory.

We treat reasoning as the decision function that transforms perceptions and
memories into candidate actions. Because decision horizon, computational
budget, and search depth are more directly auditable than broad cognitive
analogies, we organize reasoning architectures along a
\textbf{Time-Scale Reasoning Paradigm}. The three paradigms below loosely echo
the System~1/System~2 distinction of dual-process theory, although that
framework offers no consensus definitions for long-horizon planning; we
therefore use the analogy strictly as a shorthand for latency and compute
budget, not as a cognitive claim.

\paragraph{Architectural Role and Reasoning Spectrum}
Reasoning is the cognitive engine of an agentic trading system, converting state representations $O_t$ and retrieved memories into structured, actionable trading hypotheses. Financial markets present a unique cognitive challenge characterized by non-stationary dynamics, low signal-to-noise ratios, and severe latency constraints. To address these demands, reasoning architectures operate across a spectrum of decision timescales:
\begin{itemize}
	\item \textbf{Reactive Control} (\Cref{ssec:reactive}): Sub-second rule-based or specialized policy execution, optimized for immediate order adjustments and safety stops.
	\item \textbf{Reflective Reasoning} (\Cref{ssec:reflective}): Second-scale deliberative inference utilizing chain-of-thought, self-critique, and tool execution to formulate structured trade rationales.
	\item \textbf{Strategic Planning} (\Cref{ssec:strategic}): Multi-minute or daily planning loops employing tree search (e.g., Monte Carlo Tree Search) and multi-step scenario simulation over extended horizons.
\end{itemize}
The central engineering trade-off lies in balancing cognitive depth and reasoning faithfulness against computational latency and the risk of search-based overfitting.

\Cref{tab:reasoning-comparison-detailed} contrasts the three paradigms in
terms of time-scale, mechanism, reproducibility level, and evaluation considerations.

\begin{tcolorbox}[
		colback=tradingagent_light!10,
		colframe=tradingagent_light!60,
		colbacktitle=tradingagent_light!80,
		coltitle=black,
		title={\bfseries Terminological Clarification: Reactive Reasoning}]
	We use ``reactive'' strictly for \textit{non-LLM} latency-constrained control.
	LLM-based methods (including ReAct adaptations) operate at second-scale or slower,
	placing them in the \textbf{reflective} category. This distinction matters because
	it prevents the misconception that LLMs can support high-frequency trading decisions.
\end{tcolorbox}

\begin{figure*}[!tb]
	\centering
	\includegraphics[width=\linewidth]{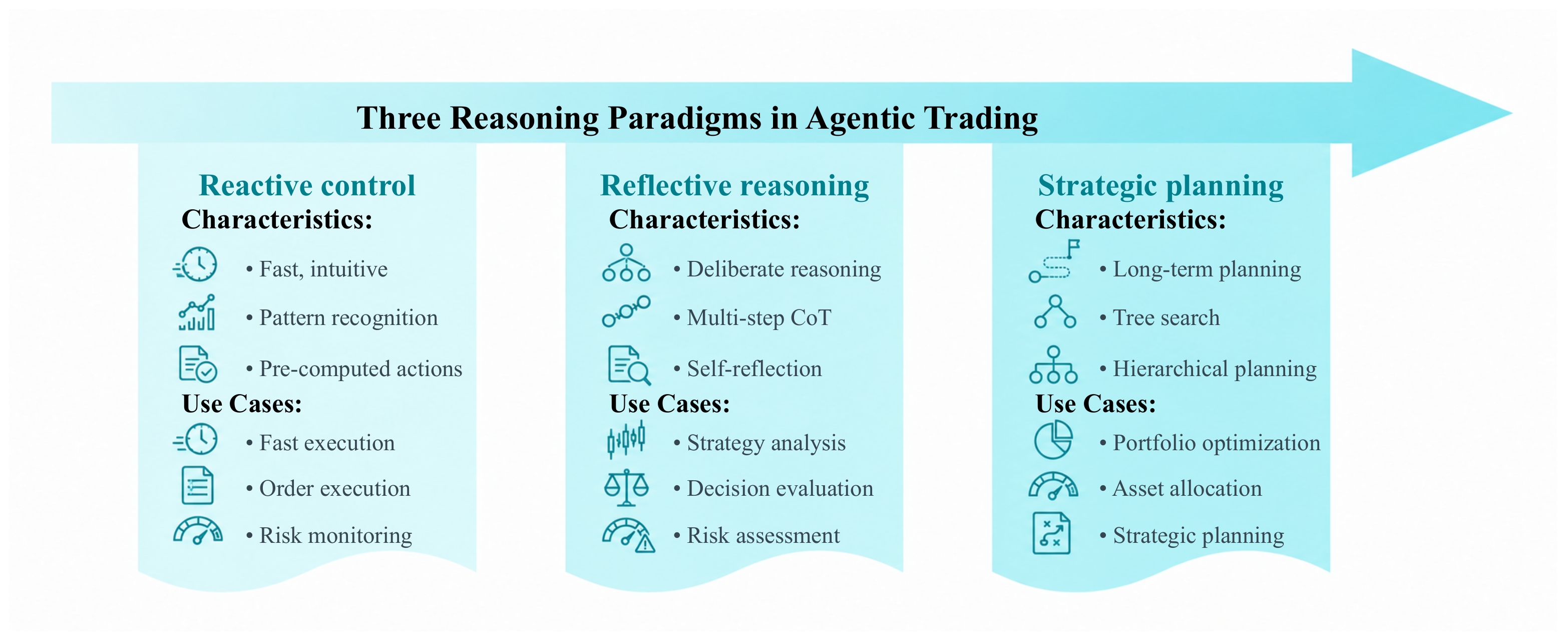}
	\caption{Three complementary reasoning roles: reactive control supports fast
        operational responses, reflective reasoning evaluates decisions through
        deliberation and self-reflection, and strategic planning explores longer
        decision horizons. Their latency depends on the model, hardware, tools,
        and search budget.}
	\label{fig:reasoning-paradigms}
\end{figure*}

\noindent Figure~\ref{fig:reasoning-paradigms} illustrates the time-scale comparison.

\textbf{Reasoning I/O Contract}. Formally, we model the agentic reasoning module as a structured decision function $f(O_t, M_{assoc}, P_{portfolio}) \to (A_{cand}, \alpha, V_{check})$, where:
\begin{itemize}
	\item \textbf{Inputs}: timestamped observation $O_t$, retrieved associative memory $M_{assoc}$, and read-only portfolio state $P_{portfolio}$ (positions, risk limits).
	\item \textbf{Outputs}: a set of candidate actions $A_{cand}$, a confidence score $\alpha \in [0,1]$, and a validity check $V_{check}$ (e.g., ``holds $<$ max\_leverage'').
\end{itemize}
The contract is designed to reduce the risk of ``hallucinated actions'' by enforcing risk constraints \emph{during} or \emph{immediately after} the reasoning step, rather than leaving them entirely to the final execution layer.

\begin{table*}[!tb]
	\centering
	\small
	\setlength{\tabcolsep}{3pt}
	\caption{{Comparison of reasoning paradigms. \textbf{R-Level}: Reproducibility level (R0--R3, see \ref{app:evidence-annotation-summary}). \textbf{Evaluation considerations} summarize protocol dimensions relevant to interpreting reported results.}}
	\label{tab:reasoning-comparison-detailed}
	\begin{tabularx}{\linewidth}{@{}l l p{3cm} c X p{2.8cm}@{}}
		\toprule
		\textbf{Paradigm} & \textbf{Time-Scale}              & \textbf{Mechanism}                             & \textbf{R-Level} & \textbf{Representative Works}                                                                               & \textbf{Evaluation considerations}       \\
		\midrule
		Reactive          & Microsecond--millisecond budgets & Rule-based / Lightweight NN (often $<$100K params) / pre-compiled decision tables & R0--R1           & ReAct \cite{yao2023react} (reflective, not reactive), reactive components in FinAgent \cite{wu2024finagent} & Execution Latency often ignored   \\
		Reflective        & Seconds                          & CoT / Self-Critique                            & R0--R1           & FinCoT \cite{nitarach2025fincot}, FinCon \cite{yu2024fincon}                                                & Inference Cost not reported       \\
		Strategic         & Minutes/Hours                    & MCTS / Planning                                & R0--R1           & Alpha Jungle \cite{navigating_alpha_jungle_2025}, RAPTOR \cite{raptor_2025}                                 & Search Budget \& Overfitting Risk \\
		\bottomrule
	\end{tabularx}
\end{table*}

\subsection{Reactive Reasoning}
\label{ssec:reactive}

Reactive reasoning, the System~1 end of the analogy above, denotes a
\textbf{constrained, simplified decision} optimized for latency budgets measured in
microseconds to milliseconds. \textbf{Importantly}, this paradigm \textit{does not} involve
LLM-mediated deliberation---the latency of even lightweight LLMs (hundreds of
milliseconds) exceeds the budgets of high-frequency trading. Instead, reactive
trading systems employ rule-based controllers, lightweight neural networks
(often $<$100K parameters), or pre-compiled decision tables that execute in
constant time.

\textbf{Representative Approaches}. ReAct \citep{yao2023react} exemplifies \textbf{reflective} (not reactive) reasoning
in general agent architectures---its ``think'' step involves explicit LLM inference
that operates at second-scale latency. In trading contexts requiring millisecond
response times, ReAct-style deliberation must be \textit{decoupled from the
critical path} and confined to offline analysis or non-latency-sensitive
decision support.
Rule-based systems encode expert knowledge as if-then rules (e.g., ``if price
crosses moving average, buy'') and execute immediately once the conditions are
met, sacrificing flexibility for speed because they rely on pre-defined patterns
rather than reasoning from first principles. Consequently, the reactive paradigm
is most effective in relatively stable market regimes where previously learned
patterns remain actionable.
In agentic trading settings, this shallow-loop structure typically appears as a
reactive component inside larger systems---for example, through tool-augmented
signal retrieval coupled with constrained actions \citep{wu2024finagent}, or
through role-specialized teams that decompose perception while preserving a
structured final decision pipeline \citep{xiao2024tradingagents}. Gradient-based
policy optimization can further couple reactive decisions to reward-driven
updates in interactive markets \citep{flag_trader_2025}.

\textbf{Technical Mechanisms}. Reactive systems employ feature extractors that
map raw market data to action logits and then select and execute the
highest-logit action. Shallow neural architectures with limited reasoning depth
are common because they balance prediction speed against accuracy; typical
choices include temporal convolutional networks for processing time-series data
\cite{bai2018tcn} and attention mechanisms for weighting relevant features
\cite{vaswani2017attention}. A further advantage is \emph{relative}
interpretability: because the mapping runs from a small feature set to actions,
it can be inspected and constrained \cite{lundberg2017shap}, although inspection
alone does not guarantee faithful explanations of the decision logic.
The same simplicity, however, limits adaptability: these systems struggle with
novel situations outside their training distribution, and when market structure
shifts abruptly they may continue applying outdated patterns, potentially
leading to significant losses.

Beyond latency, the integration of tool use into financial agents requires
explicit evaluation along at least two dimensions: AgentGuard studies
safety-oriented evaluation for tool orchestration \cite{chen2025agentguard},
whereas FinToolBench measures the capability of LLM agents to use financial
tools in realistic tasks \cite{lu2026fintoolbench}. Autonomous agent frameworks
can likewise strengthen complex reasoning over structured knowledge
\cite{jiang2024kgagent}.

\subsection{Reflective Reasoning}
\label{ssec:reflective}

Reflective reasoning, the System~2 counterpart in the same analogy, produces
slow, deliberate decisions through multi-step chains of thought. This paradigm
enables agents to reason through complex scenarios, consider multiple hypotheses,
and reflect on potential mistakes before acting. Unlike reactive systems that
respond immediately to stimuli, reflective reasoning allocates computational
resources to explore the decision space more thoroughly, which makes it suitable
for medium-frequency trading where execution latency on the order of seconds is
acceptable.

\textbf{Representative Approaches}. Chain-of-Thought (CoT) prompting
\citep{wei2022chain} guides LLMs to explicitly reason through trading decisions,
producing intermediate steps such as ``the stock price dropped because of worse-
than-expected earnings, but the long-term fundamentals remain strong, so this
is a buying opportunity.'' FinCoT \citep{nitarach2025fincot} adapts CoT to
finance through domain-grounded prompting that encourages financial reasoning in
trading tasks. FinCon \citep{yu2024fincon} operationalizes reflective
workflows through multi-agent or role-structured coordination that updates
investment beliefs over time. CryptoTrade \citep{cryptotrade_zero_shot_2024} and TradingGroup
\citep{tradinggroup_2025} further instantiate this reflective pattern by
revising trading decisions based on prior outcomes and synthesizing
multi-source signals.
Reflexion \citep{shinn2023reflexion} extends this paradigm with self-reflection,
in which agents critique their own reasoning and refine their conclusions before
acting. A Reflexion-based trader, for example, might generate an initial trading
rationale, identify potential flaws in its logic (such as overlooking
sector-wide risks), and revise its decision accordingly.

\begin{tcolorbox}[title={Technical Mechanisms: Protocol Implication},
  title after break={Protocol Implication (continued)}]
	We synthesize five stages for reflective trading decisions: (1) analyze the
	current situation; (2) retrieve relevant memories; (3) consider multiple
	hypotheses; (4) assess each candidate action's benefits and risks; and
	(5) select an action and justify the choice.
\end{tcolorbox}
Temperature sampling controls the trade-off between creativity and consistency,
with lower temperatures producing more deterministic reasoning. A recurring
challenge is maintaining coherence across long reasoning chains, since LLMs may
lose context or contradict themselves over dozens of steps; checkpointing
(saving intermediate reasoning states) and consistency checks (verifying that
conclusions follow from premises) help mitigate these issues
\cite{wei2022chain,wang2022selfconsistency,madaan2023selfrefine}.
The computational cost of reflective reasoning scales linearly with the number
of reasoning steps, so it is more expensive than reactive approaches but also
significantly more capable of handling novel situations that require adaptation.

A further caveat concerns feedback timing: Reflexion-style systems assume
relatively prompt feedback, whereas trading outcomes often materialize only
after meaningful market delay \cite{shinn2023reflexion}. Beyond self-critique
loops, modern risk-sensitive frameworks provide structured decision support for
financial markets, although their exact internal decomposition varies by
implementation \cite{liu2025risksensit}.

\subsection{Strategic Reasoning}
\label{ssec:strategic}

Strategic reasoning extends beyond immediate decisions to plan across extended
time horizons, weighing sequences of actions and their long-term consequences.
This paradigm addresses questions such as ``what portfolio allocation will
maximize risk-adjusted returns over the next quarter?'' rather than ``should I
buy this stock right now?'' It operates at time scales of minutes to hours,
which suits portfolio optimization, risk management, and longer-term investment
strategies where thorough planning matters more than immediate execution.

\textbf{Representative Approaches}. Tree-of-Thought (ToT) \citep{yao2023tot}
frames reasoning as search through a tree of possible thoughts, exploring multiple
reasoning paths in parallel and selecting the most promising through explicit
evaluation. Each node in the tree represents a partial reasoning state, and the
algorithm systematically explores alternatives rather than committing to the first
plausible solution. Monte Carlo Tree Search (MCTS) methods such as Navigating
Alpha Jungle \citep{navigating_alpha_jungle_2025} apply the same idea to the
space of possible trading strategies, using rollouts to estimate the long-term
value of each candidate strategy. These methods balance exploration (trying new strategies)
against exploitation (refining known good strategies) through upper confidence
bounds applied to financial domains. Planning ideas from classical AI and
reinforcement learning continue to inform strategic trading design: state
spaces represent market conditions and actions represent trading decisions.
For portfolio-scale planning, RAPTOR \citep{raptor_2025} integrates
deliberative portfolio reasoning with an orchestrated rebalancing pipeline to
produce interpretable, risk-aware allocations.

\textbf{Technical Mechanisms}. Strategic reasoning requires models of the
environment---predicting how market prices will evolve in response to trading
actions. Model-based methods learn explicit transition models that capture
causal relationships between actions and market states, which then allow agents
to simulate future scenarios and evaluate action sequences without executing
them. Model-free methods instead learn value functions directly from
experience, estimating the long-term return of each state-action pair without
an explicit environment model \cite{sutton2018reinforcement}. To keep planning tractable, hierarchical schemes decompose long horizons into shorter sub-problems:
a strategic planner sets high-level targets (e.g., ``reduce tech exposure by
20\%''), and tactical planners execute specific trades to achieve these targets.
Abstraction is critical here---planning over individual stocks is intractable
because the space of possible portfolios explodes combinatorially, whereas
planning over sectors or factors reduces the state space to manageable
dimensions. A strategic planner might therefore operate at the level of sector
allocations (technology, healthcare, energy), while tactical planners handle
individual stock selection within each sector, so that agents can reason about
long-term objectives while remaining computationally feasible
\cite{kocsis2006uct,sutton1999options}.

\textbf{Critical Challenge: Search-Based Overfitting}. Unlike game environments (e.g., Go, Chess) where simulated rollouts are exact, financial planning relies on noisy historical data. Search algorithms (MCTS, ToT) that evaluate thousands of candidate strategies therefore face the \textit{multiple testing} problem---a profitable-looking trajectory may emerge purely by chance. To mitigate this \textit{data snooping} risk, reproducible evaluation requires reporting the \textbf{Search Budget} (number of visited nodes) and enforcing strict \textbf{Walk-Forward Validation}, in which the plan is frozen before being applied to out-of-sample data \cite{bailey2014deflating}.


\textbf{Hybrid Architecture: Cascaded Controller}. To balance these trade-offs, advanced systems can employ what we call a \textbf{Cascaded Controller} architecture (cf.\ \Cref{fig:reasoning-paradigms} for the paradigms it routes among): a master intent router classifies the market regime (e.g., ``Flash Crash'' vs. ``Stable Trend'') and routes execution to the appropriate paradigm, triggering lower-level reactive stops in volatile conditions while scheduling strategic re-allocations during calm periods. This architectural recommendation synthesizes principles observed in hierarchical trading systems \cite{yu2023finmem,xiao2024tradingagents}.

\textbf{Extension-window addition.} Complementing the paradigm taxonomy, work
in the extension cohort examines how market state is \emph{represented} for
reasoning: representation signatures that align internal features with
risk-feedback signals \cite{xue2026representation} probe whether an agent's
reasoning is grounded in decision-relevant structure rather than surface
patterns.

\paragraph{Synthesis and Design Considerations}
Reactive control, reflective reasoning, and strategic planning trade decision speed against deliberation and search depth; reflective methods depend on alignment with market timescales, while strategic search requires explicit exploration budgets and walk-forward verification to guard against data snooping. Candidate decisions then pass to the action execution layer examined next.


\section{Action and Execution Architecture}
\label{sec:action}

The preceding sections examined perception, memory, and reasoning. This section
addresses the complementary question: \textquotedblleft How do agents turn
decisions into market actions?\textquotedblright

\paragraph{Architectural Role and Execution Scope}
The action and execution architecture bridges cognitive reasoning with the physical reality of financial markets, marking the definitive boundary between an abstract forecasting model and a genuine trading agent. While forecasting models evaluate directional accuracy on paper, an autonomous trading agent must transform high-level intentions into executable market orders subject to microstructure frictions, liquidity limits, and latency budgets. The execution design space encompasses three key dimensions:
\begin{itemize}
	\item \textbf{Decision-to-Order Mapping}: Translating target weights or directional convictions into discrete order specifications (order side, quantity, limit price, time-in-force).
	\item \textbf{Execution Policies and Cost Modeling}: Implementing algorithmic order slicing (e.g., TWAP, VWAP, POV) and explicitly modeling fees, spreads, and market impact.
	\item \textbf{Microstructure and Latency Management}: Incorporating order book dynamics, fill probabilities, and exchange round-trip latency into trade scheduling.
\end{itemize}

\begin{figure*}[!tb]
	\centering
	\includegraphics[trim=20bp 160bp 15bp 165bp,clip,width=\linewidth]{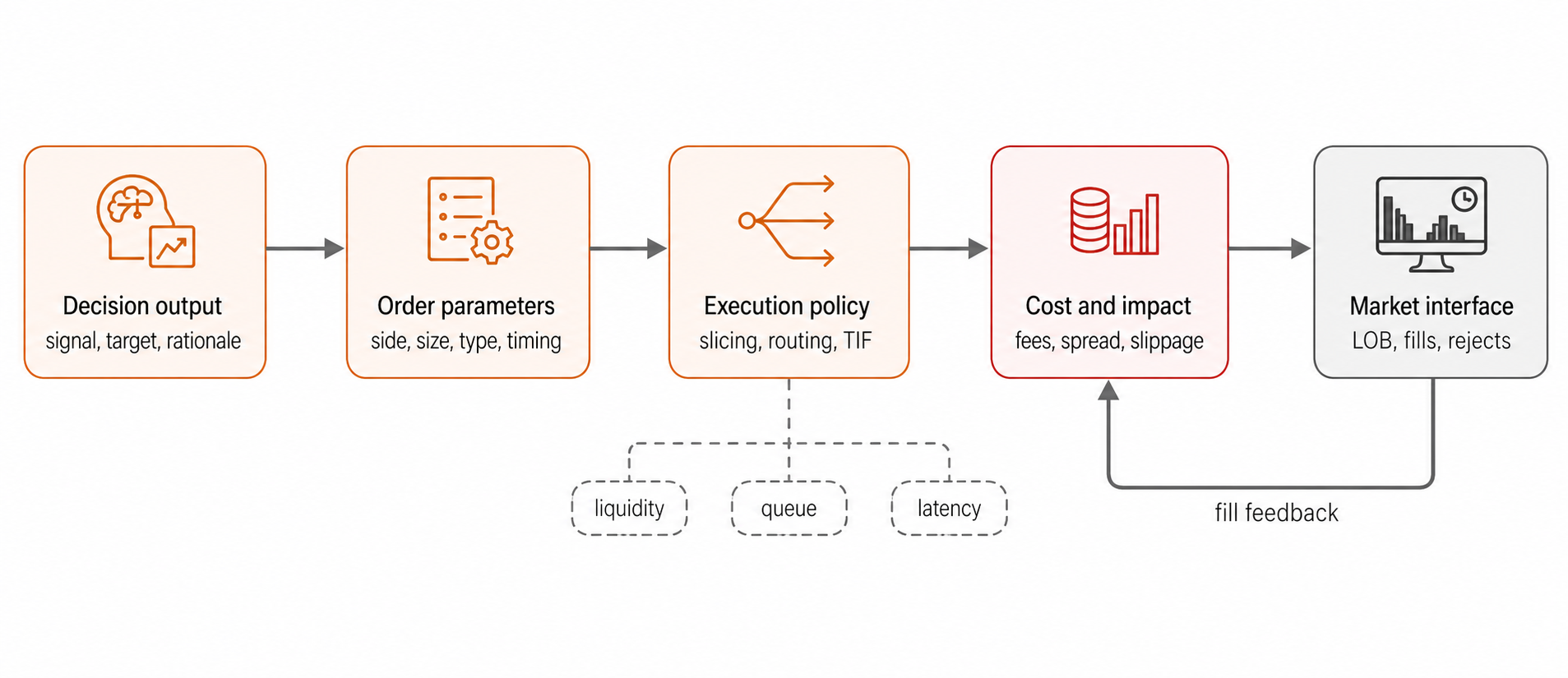}
	\caption{From decision to execution. Agent outputs map into order parameters and
		are executed with explicit cost and impact modeling under market micro\-structure constraints. }
	\label{fig:action-execution-flow}
\end{figure*}

\textbf{Action I/O Contract}. Formally, we model the action execution module as a structured function $f(D_t) \to (A_{order}, A_{algo})$ (see \Cref{fig:action-execution-flow}), where:
\begin{itemize}
	\item \textbf{Input}: Decision output $D_t$ (e.g., target portfolio weights $w^*_t$ or signal $s_t$).
	\item \textbf{Output}: A set of executable orders $A_{order}$ (with strictly typed fields: side, size, limit price, time-in-force (TIF)) and an execution algorithm selection $A_{algo}$ (e.g., ``submit immediately'' or ``TWAP over 5 min'').
	\item \textbf{Failure Modes}: The interface typically handles \textit{Rejection} (risk check failure), \textit{Partial Fill} (liquidity shortage), and \textit{Timeout} (latency violation), feeding these states back to memory.
\end{itemize}
The contract is designed to prevent ``implicit execution,'' in which an agent is assumed to trade at the close price without order-generation delays or constraints.

\paragraph{Comparability note} Protocol-comparable action/execution claims typically report MR-3 (action semantics), MR-4 (execution \& costs), and MR-6 (artifacts \& logs); see \Cref{sec:challenges}.

\begin{table*}[!tb]
	\centering
	\small
	\setlength{\tabcolsep}{3pt}
	\caption{Representative action/execution approaches and evaluation dimensions. \textbf{Type}: E2E = end-to-end agent literature; Algo = algorithmic-execution literature. Note that \textit{Agent} literature often adopts simplified assumptions (Sim-based), whereas \textit{Algo} literature focuses on execution quality but lacks agentic reasoning. \textbf{Evaluation considerations} summarize execution details relevant to interpreting reported results. \textbf{Tags} follow the R0--R3 reproducibility-level convention (see \ref{app:evidence-annotation-summary}).}
	\label{tab:execution-comparison}
	\begin{tabularx}{\linewidth}{@{}l l p{2.1cm} X p{2.8cm} p{2.8cm} p{1.6cm}@{}}
		\toprule
		\textbf{Category} & \textbf{Type} & \textbf{Action Space} & \textbf{Core Mechanism}          & \textbf{Cost Assumption} & \textbf{Evaluation considerations} & \textbf{Tags} \\
		\midrule
		Schedule-based    & Algo          & Order schedule        & TWAP/VWAP-style scheduling       & Explicit fees            & Slippage Calib.             & \texttt{R0}   \\
		RL Execution      & Algo          & Market Orders         & Policy optimization in simulator & Impact-aware reward      & Latency Budget              & \texttt{R0}   \\
		Sim-based RL      & E2E           & Order-level           & Market simulator + RL            & Simulator-dependent      & Decision Timing             & \texttt{R0}   \\
		Microstructure    & Algo          & Limit Orders          & Fill/LOB modeling                & Implicit via LOB         & Realized Spread             & \texttt{R0}   \\
		Hierarchical      & E2E           & Meta-Actions          & Hierarchical RL for order types  & Explicit/implicit        & Failure Modes               & \texttt{R0}   \\
		\bottomrule
	\end{tabularx}
\end{table*}

\subsection{Decision-to-Order Mapping}
\label{ssec:order-mapping}

\textbf{First Principle: Action Definition Risk.} A critical failure mode in agent design is the ambiguity between \textit{target weights} (portfolio intent) and \textit{executable orders} (market instructions). If an agent emits only target weights without explicitly modeling order timing, fill constraints, and costs, evaluation quietly collapses intent into execution and thereby underestimates the \textit{execution gap} and liquidity constraints \cite{wu2024finagent, chen2022cost, cheridito2025reinforcement, finpos_2025, stocksim_2025, olby2025right}.

In the surveyed systems, action modules translate high-level decisions (e.g., \textit{increase exposure}) into concrete order instructions. This mapping depends on explicit choices about order type (limit versus market) and queue position, because microstructure theory models execution probability as a function of order aggressiveness and market state \cite{foucault1999orderflow, parlour1998price}. Constraints such as pre-trade validation and kill switches then act as safeguards that keep theoretical alpha from becoming realized losses once market impact is accounted for \cite{kyle1985continuous, glosten1985bid}.

\subsection{Execution and Cost Modeling}
\label{ssec:execution}

Execution policies determine how orders are placed over time and across venues.
Typical approaches include schedule-based execution (e.g., time-weighted
average price (TWAP) and volume-weighted average price (VWAP)), cost-aware
optimization that trades urgency against market impact
\cite{almgren2001optimal, bertsimas1998optimal, perold1988shortfall}, and
learning-based execution that adapts to market conditions. Recent studies apply
deep learning and reinforcement learning to execution and impact-aware trading
\cite{genet2025deep, byun2023practical, cheridito2025reinforcement, hafsi2024optimal, chen2022cost, olby2025right}.
Within this line, FinPos \cite{finpos_2025} introduces a position-aware trading
agent designed for real financial markets, emphasizing continuous portfolio
adjustment, while StockSim \cite{stocksim_2025} provides a dual-mode
order-level simulator for evaluating multi-agent LLMs in financial markets and
thereby enables realistic execution modeling.
\Cref{tab:execution-comparison} contrasts representative action and execution
approaches across these categories, summarizing their action spaces, cost
assumptions, and common audit gaps.

\subsection{Microstructure and Latency Awareness}
\label{ssec:microstructure}

\sloppy
Microstructure signals (order book imbalance, spread, queue position) act not merely as inputs for optimization but as \textbf{hard feasibility constraints} that filter the action space. Throughout, \textbf{LOB} denotes the real-time queue of outstanding buy and sell limit orders; its features include imbalance (bid versus ask volume), spread, and depth.
\begin{enumerate}
	\item \textbf{Data Availability}: Agents can incur ``look-ahead bias'' when they act on LOB features (e.g., $Imbalance_t$) that are not yet available at decision time $t$ because of processing latency; feasible actions must therefore be conditioned on $O_{t-\delta}$.
	\item \textbf{Latency Budget}: In high-frequency regimes, the \textit{latency budget}---the time from perception to exchange acknowledgment---determines whether an arbitrage opportunity still exists \cite{budish2015arms}: if $T_{compute} > T_{decay}$, the action $a_t$ is invalid regardless of its theoretical alpha.
\end{enumerate}
\fussy
Models of order book dynamics \cite{gould2013lob, briola2024deep} then allow these constraints to be validated before execution.

\begin{tcolorbox}[title=Protocol Implication]
	Beyond execution optimization, LLMs can support \textbf{post-trade narratives} by summarizing execution logs. To reduce ``ex post rationalization'' (in which the LLM invents a strategic reason for a random failure), we recommend grounding these narratives in \textbf{immutable logs} and replayable evidence \cite{tan2025fred,khatchadourian2026replayable}.

	\begin{itemize}
		\item \textbf{Recommendation}: Narrative generation can cite the specific $Order\_ID$ and cryptographic $Snapshot\_Hash$ produced by the execution layer.
		\item \textbf{Verification}: Auditors can replay the $Snapshot\_Hash$ to check whether the market condition claimed by the narrative (e.g., ``high spread'') actually held.
	\end{itemize}
\end{tcolorbox}
Without this \textit{Log-to-Narrative Verification} link, LLM explanations remain qualitative summaries rather than audit artifacts.

\textbf{Extension-window additions.} The extension cohort sharpens the
execution side along two directions. First, evaluation infrastructure:
systematic benchmarking of learning-based equity trading \cite{wang2026deep}
and inference-time optimization that plans before committing orders
\cite{go2026plan} both treat the decision-to-order mapping as a measurable
component rather than an implementation detail, while realistic market-impact
modeling for RL traders \cite{abbade2026realistic} supplies the cost layer that
MR-4 asks papers to report; auditability-oriented execution frameworks such as
AutoQuant \cite{deng2025autoquant} reinforce the same principle from the
engineering side by treating execution constraints as an auditable system
boundary. Second, a crypto-native execution niche: multi-agent MEV generation
and cross-chain arbitrage systems \cite{wen2026evage,ye2026when} operate
directly on order-flow and settlement mechanics, a setting in which execution
semantics constitute the strategy rather than a downstream assumption.

\paragraph{Synthesis and Practical Considerations}
The action execution layer grounds cognitive decisions in market dynamics by accounting for transaction costs, slippage, fill timing, and microstructure constraints. In multi-agent or institutional deployments, structured order logging supports audits that link execution traces to strategic reasoning. With these architectural foundations established, Part~II turns to enabled capabilities, beginning with automated alpha discovery.


\section{Alpha Discovery Capability}
\label{sec:alpha}
\label{part:capability}

The preceding sections established the architectural foundations of agentic
traders. This section addresses the question: \textquotedblleft How do agents
discover trading signals?\textquotedblright

Here, alpha discovery refers to generating and testing candidate trading
signals; consistent with the evidence-map framing, we count autonomous alpha
search as supported only when the reported system connects discovery to
out-of-sample or closed-loop evaluation, rather than merely proposing factors.

\paragraph{Functional Scope and Discovery Paradigms}
Automated alpha discovery represents one of the most prominent applications of LLM agents in quantitative finance, transitioning formulaic factor generation from manual trial-and-error to autonomous, hypothesis-driven exploration. By combining financial reasoning with programmatic code generation, agents search through vast functional hypothesis spaces to uncover predictive trading signals. Across the literature, alpha discovery is realized through three core paradigms:
\begin{itemize}
	\item \textbf{Code-Based Generation} (\Cref{ssec:code-alpha}): Expressing economic hypotheses in executable code and refining them through automated syntax and runtime feedback.
	\item \textbf{Retrieval-Augmented Discovery} (\Cref{ssec:retrieval-alpha}): Grounding factor exploration in academic literature, financial corpora, and historical factor libraries to seed domain-consistent hypotheses.
	\item \textbf{Evolutionary and Genetic Search} (\Cref{ssec:evo-alpha}): Guiding symbolic expression trees through LLM-directed mutation, crossover, and fitness selection.
\end{itemize}
The central methodological considerations include managing search budgets to prevent multiple-testing data snooping, enforcing factor diversity to mitigate crowding, and verifying that statistical predictive power survives net of turnover and execution frictions.

\begin{tcolorbox}[
		colback=gray!10,
		colframe=gray!50,
		colbacktitle=gray!30,
		coltitle=black,
		fonttitle=\bfseries\sffamily\small,
		title={Classical Baseline: Factor Models and Information Coefficient},
		boxrule=0.5pt,
		arc=2pt,
		left=5pt,
		right=5pt,
		top=3pt,
		bottom=3pt
	]
	\small
	Before LLM-based approaches, alpha generation relied on manually engineered
	factor models (e.g., the Fama--French three-factor model, momentum, value), in
	which factor validity is measured by the information coefficient (IC), the
	correlation between factor scores and subsequent returns, and statistical rigor
	required controlling for multiple testing (White's Reality Check, Hansen's SPA)
	to distinguish genuine signals from data mining
	\cite{fama1993common,grinold2000active}. LLM-based agents automate this
	discovery process but inherit the same fundamental challenge: ensuring that
	discovered patterns generalize out-of-sample rather than exploit historical
	noise.
\end{tcolorbox}

\paragraph{Comparability note} Comparable alpha-search evidence should report MR-2 (time split + sealed test), MR-4 (execution \& costs), factor definitions, and MR-6 (artifacts \& logs); see \Cref{sec:challenges} for detailed protocol requirements and reporting checklists.

\begin{tcolorbox}[title=Author Framework,
  title after break={Author Framework (continued)}]
	\textbf{Alpha-to-Trade Contract}. We introduce the Alpha-to-Trade Contract as a conceptual framework for separating statistical significance from tradable signals; the conversion contract specifies three checks:
	\begin{enumerate}
		\evidenceitem{\textbf{Frequency Compatibility}: Does the signal update frequency match the execution latency budget?}
		\evidenceitem{\textbf{Turnover Constraints}: Does the theoretical gain survive realistic transaction costs and market impact?}
		\evidenceitem{\textbf{Liquidity Reality}: Is the factor universe liquid enough for the assumed capital capacity?}
	\end{enumerate}
	Papers omitting this contract risk conflating theoretical IC with realized alpha return, a common pitfall in agentic trading literature \cite{lopezprado2018dangers}.
\end{tcolorbox}

\begin{figure*}[!tb]
	\centering
	\includegraphics[width=\linewidth]{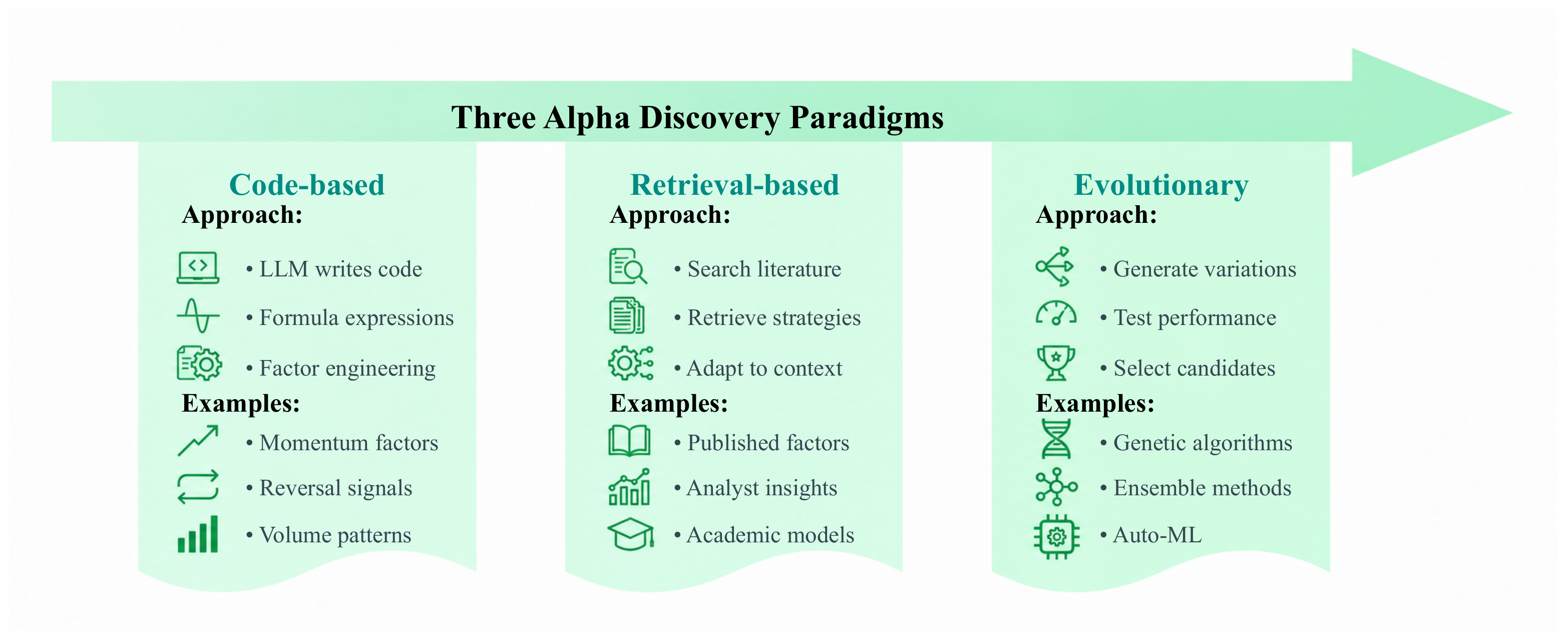}
	\caption{Three alpha discovery paradigms. Code-based discovery translates
		natural language hypotheses into executable code. Retrieval-based discovery
		finds and adapts existing factors from a library. Evolutionary discovery
		searches the factor space through mutation and selection.}
	\label{fig:alpha-paradigms}
\end{figure*}

Figure~\ref{fig:alpha-paradigms} illustrates these discovery paradigms.

\subsection{Code-based Alpha Discovery}
\label{ssec:code-alpha}

Code-based alpha generation draws on the code-synthesis capabilities of LLMs
to translate natural-language hypotheses about market behavior into executable
factor definitions. Because it lowers the programming barrier, this paradigm
lets domain experts express and test their trading ideas directly.

\textbf{Representative Approaches}. CogAlpha
\citep{cogalpha_2025} extends code-based generation with structured prompting templates that guide the LLM
toward syntactically correct and semantically meaningful factor definitions.
AlphaAgent \citep{alphaagent_2025} further introduces regularization mechanisms (originality enforcement, hypothesis alignment, and complexity control) to counteract alpha decay. FactorMAD \citep{factormad_2025} uses a
multi-agent debate framework in which LLM agents critique and refine
factor proposals to enhance interpretability. Automate Strategy Finding
\citep{automate_strategy_2025} implements a three-stage pipeline with risk-aware multi-agent
validation and ensemble strategy selection, and reports 53.17\% cumulative returns on SSE50 under its backtest
setup (\textit{Split}: \texttt{NR}; \textit{Execution}: \texttt{NR};
\textit{Cost}: \texttt{NR}; \textit{Universe}: SSE50; \textit{Tier}:
\texttt{BG}); because protocol-critical fields are \texttt{NR}/\texttt{Unknown}, we
read such results as illustrative rather than protocol-comparable evidence.

Beyond factor synthesis, recent work combines LLM reasoning with reinforcement
learning for structured trading decision-making~
\cite{xiao2025tradingr1},
and ensemble strategies built on deep reinforcement learning report strong
performance on stock trading tasks~
\cite{yang2025deep}.

\begin{figure*}[!tb]
	\centering
	\includegraphics[width=0.82\linewidth]{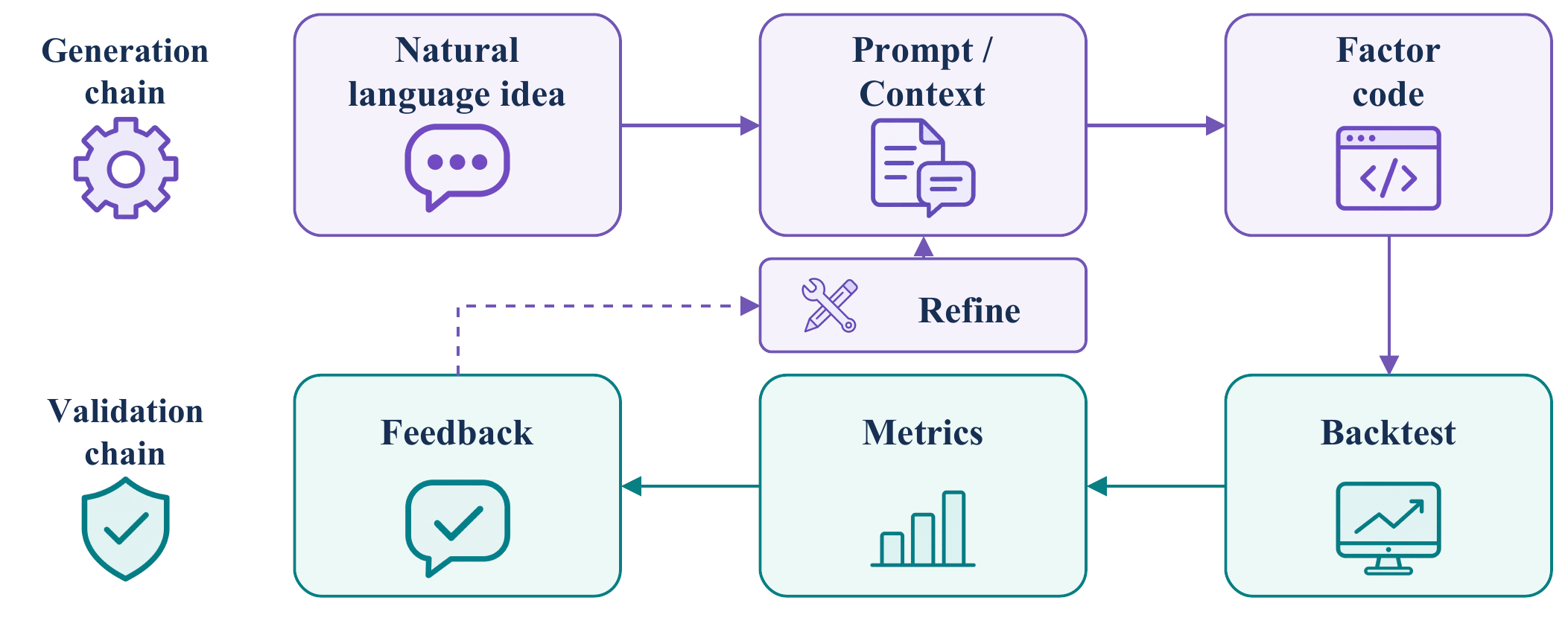}
	\caption{Chain-of-Alpha workflow. The generation chain (top) translates
		natural language into factor code. The validation chain (bottom) backtests
		the factor, computes performance metrics, and provides feedback for refinement.
		The two chains interact iteratively until a satisfactory factor is produced.}
	\label{fig:coa-workflow}
\end{figure*}

\textbf{Technical Mechanisms}. The generation pipeline operates as an \textit{adaptive search process} susceptible to overfitting: it (1) parses the natural-language hypothesis to identify key components, (2) maps them to code templates and APIs, (3) generates executable code, (4) validates it through sandboxed execution, and (5) backtests it on historical data; \Cref{fig:coa-workflow} illustrates this generation--validation loop. Crucially, the ``validation chain'' feeds performance feedback into the next generation step, so the loop risks \textbf{adaptive overfitting}: as the agent iterates to maximize the validation Sharpe ratio, it effectively ``mines'' the validation set \cite{bailey2014deflating, lopezprado2018dangers}. A rigorous protocol therefore requires reporting the \textbf{Search Budget} (the number of generated hypotheses) and ensuring that final performance is measured on a sealed \textit{test set} that is never exposed to the agent during the refinement loop; without this separation, reported ``validation'' metrics are statistically inflated. Sandboxed execution, standard practice for mitigating security risks, addresses a different concern and does not by itself enforce split integrity. Whatever form the output takes---synthesized factors include momentum and reversal signals---its validity depends entirely on the integrity of the data split.

\subsection{Retrieval-based Alpha Discovery}
\label{ssec:retrieval-alpha}

Retrieval-based alpha generation adapts existing factor knowledge rather than
generating factors from scratch: the agent retrieves related signals or
financial context that can inform the current situation.

\textbf{Representative Approaches}. RAG-based financial retrieval systems \citep{rag_fintech_2025}
index domain knowledge as embeddings and retrieve contextually similar items
using semantic search. PRISM \citep{prism_financial_retrieval_2025} refines this retrieval through
prompt-optimized in-context modeling for financial domains, and retrieval-augmented
frameworks for financial time-series forecasting \citep{rag_financial_time_series_2025} illustrate how
LLM generation can be combined with historical pattern retrieval.

\textbf{Technical Mechanisms}. Retrieval relies on vector embeddings that encode factor semantics (formula structure) and market context. The validity of this paradigm, however, hinges on \textbf{Factor Library Governance}, a data-integrity requirement that is easy to overlook.
\begin{itemize}
	\item \textbf{Survivorship Bias}: If the library collects only ``historically successful'' factors, the retrieval process implicitly conditions on future outcomes, inflating backtest performance.
	\item \textbf{Look-ahead Leakage in Embeddings}: If the embedding model is trained on textual descriptions containing future knowledge (e.g., ``a successful high-volatility factor''), the retrieval query $q_t$ effectively leaks information from $t+k$.
\end{itemize}
\begin{tcolorbox}[title=Protocol Implication]
	We recommend \textbf{Point-in-Time (PIT) Governance} for verifiable evidence: the retrieval corpus $L_t$ available at time $t$ must contain only documents and performance metadata known at $t$ \cite{rag_fintech_2025, prism_financial_retrieval_2025}.
\end{tcolorbox}

\subsection{Evolutionary Alpha Discovery}
\label{ssec:evo-alpha}

Evolutionary alpha generation searches the factor space through iterative
mutation, crossover, and selection, treating factor discovery as an optimization
problem. This paradigm can discover novel factor combinations that human experts
might not conceive.

\textbf{Representative Approaches}. Navigating Alpha Jungle \citep{navigating_alpha_jungle_2025}
employs Monte Carlo Tree Search (MCTS) to efficiently explore the combinatorial
space of possible factors: each node in the search tree represents a partial
factor definition (e.g., a specific transformation applied to a price series),
and MCTS balances exploration (trying new factor components) against
exploitation (refining promising factors). Alpha\textsuperscript{2} \citep{alpha2_2024}
uses deep reinforcement learning, in which an agent learns to construct factors
by sequencing primitive operations and receives rewards based on backtest
performance. EFS (Evolutionary Factor Searching) \citep{efs_2025} reformulates asset selection
as a top-m ranking task guided by LLM-generated factors, incorporating an
evolutionary feedback loop for iterative refinement.

\textbf{Technical Mechanisms}. Evolutionary methods define a search space of primitive operations (price transformations, technical indicators) and combination rules, which MCTS and Genetic Programming (GP) navigate by balancing exploration (new operations) against exploitation (refining promising factors). The paradigm, however, faces a severe \textbf{Multiple Testing Scalability} risk: as the search space grows exponentially, the probability of finding a false positive approaches 1.0 unless rigorous statistical corrections are applied \cite{white2000reality}. A valid protocol must therefore report (1) a \textbf{Primitive Ops Audit} (the exact set of allowed functions), (2) the \textbf{Total Search Depth/Budget}, and (3) statistical tests for data snooping (e.g., White's Reality Check or Hansen's SPA) \cite{hansen2005test}. At present, few agent papers report these quantities, which makes their results difficult to distinguish from noise; regularization techniques (complexity penalties, ensemble methods) are useful heuristics but do not substitute for formal multiple-hypothesis testing.

\textbf{Open Question: Mathematical Bounds for Agent-Driven Search.}
While classical multiple-testing corrections (White's Reality Check, Hansen's SPA, and the Deflated Sharpe Ratio \cite{bailey2014deflating}) remain the statistical gold standard, a critical question emerges for LLM-based agents: can the agent's internal reflection/evolution loop provide novel mathematical bounds on overfitting risk? Current evidence suggests \textbf{not yet}. The reflection budget (\Cref{ssec:reflection}) limits search iterations but offers no statistical guarantee on the false discovery rate; the sealed test set protocol prevents adaptive overfitting but relies on experimental discipline rather than closed-form bounds; and factor originality constraints (e.g., AST-based similarity in AlphaAgent) shrink the effective search space without quantifying the residual multiple-testing penalty. \textit{This represents a significant research gap}: future work could explore whether LLM reasoning traces, factor genealogy logs, or Bayesian model averaging over discovered factors can yield principled, agent-native overfitting controls that complement or extend classical methods. Until such methods exist, evolutionary alpha systems should report both (1) search-budget and protocol constraints and (2) classical multiple-testing statistics (e.g., deflated Sharpe ratio \cite{bailey2014deflating}) to enable rigorous evaluation.

\begin{table*}[!tb]
	\centering
	\small
	\setlength{\tabcolsep}{3pt}
	\caption{Representative alpha-discovery mechanisms and design considerations. The considerations identify review dimensions discussed in this survey. Tags describe artifact availability (R0--R1) or background literature (BG); see \Cref{tab:reproducibility-summary}.}
	\label{tab:alpha-discovery-detailed}
	\begin{tabularx}{\linewidth}{@{}l p{2.0cm} p{1.8cm} X p{2.7cm} p{1.4cm}@{}}
		\toprule
		\textbf{Paradigm} & \textbf{Mechanism} & \textbf{Search object} & \textbf{Dimensions to examine} & \textbf{Works} & \textbf{Tags} \\
		\midrule
		Code-based   & LLM $\rightarrow$ Code & Semantic factors & Structured prompting; hypothesis-to-code mapping & CogAlpha \cite{cogalpha_2025} & \texttt{R0} \\
		Code-based   & Multi-Agent Debate & Interpretable factors & Debate-based critique and refinement & FactorMAD \cite{factormad_2025} & \texttt{BG} \\
		Retrieval    & Vector Search & Factor library & Point-in-time (PIT) library governance & RAG-Fintech \cite{rag_fintech_2025} & \texttt{R0} \\
		Retrieval    & Prompt-Refined RAG & Financial corpus & Point-in-time (PIT) corpus governance & PRISM \cite{prism_financial_retrieval_2025} & \texttt{BG} \\
		Evolutionary & MCTS / Deep RL & Combinatorial operations & Allowed operations; search budget; multiple testing & Alpha Jungle (R0), Alpha$^{2}$ (R1) \cite{navigating_alpha_jungle_2025,alpha2_2024} & \texttt{R0--R1} \\
		Evolutionary & LLM-Guided \mbox{Evolution} & Sparse portfolios & Search budget and sealed-test separation & EFS \cite{efs_2025} & \texttt{BG} \\
		\bottomrule
	\end{tabularx}
\end{table*}

\Cref{tab:alpha-discovery-detailed} organizes representative mechanisms and
their design considerations.

\textbf{Extension-window additions (2026-03-10 to 2026-09-30).} Additional
studies automate factor or strategy discovery through agentic
generation--backtest loops
\citep{weng2026alphalogics,yuan2026alphacrafter,yu2026alphamemo,li2026ai,liu2026xalpha,yi2026alphaschema,zhao2026goant,wang2026alphadiverse,li2026evermine,zhao2026alphapareto,kim2026favor,ye2026agonalpha,li2026vst,kvasiuk2026madevolve,sharma2026algoevolve,kim2026evolvetrade},
with further variants pushing toward trajectory-level self-iteration
\citep{zhu2026alphaseek}, constrained hypothesis-to-factor pipelines in
cryptocurrency markets \citep{huang2026hypotheses}, self-evolving coding agents
for quantitative research \citep{li2026autoscientist}, and stratified
strategy-alignment objectives \citep{huang2026strat}.

The reported evaluation settings span held-out or out-of-sample windows,
per-trade or turnover-based costs, and named index universes such as
CSI~300/500 or S\&P~500. Interpreting results across these settings requires
attention to the split, cost, and universe conventions used by each study.

A parallel mini-cluster interrogates the evaluation layer
itself: closed-loop benchmarks and referee protocols for mining agents
\citep{qu2026clqt,qu2026propose,wang2026crossalpha,deng2026search,cao2026systradebench,zhang2026openfingym},
together with leakage and validity analyses
\citep{ye2026alpha,zhang2026temporal,yao2026beyond}, now make evaluation design
an explicit research object rather than an implementation detail. Two further
studies deserve note: a direct assessment of LLMs on technical
market-analysis tasks \citep{ntale2026ai}, and evidence that nominally valid
LLM signals can fail across regime boundaries \citep{yang2026when}---a useful
reminder that alpha-mining output is regime-conditional even when protocol
fields are reported.

\paragraph{Synthesis and Empirical Considerations}
Code-based generation, retrieval-augmented discovery, and evolutionary search explore different parts of the factor hypothesis space. Their empirical value depends on out-of-sample survival: recursive search can overfit noisy backtests, so walk-forward testing, factor-correlation checks, and turnover-adjusted transaction costs remain essential. The next section examines how discovered signals are aggregated into portfolios in \Cref{sec:portfolio}.



\section{Portfolio Management Capability}
\label{sec:portfolio}

The preceding section examined how agents generate alpha signals. This section
addresses the question: \textquotedblleft How do agents manage multi-asset
portfolios?\textquotedblright

The evidence is strongest where papers specify the action object---weights,
orders, constraints, or rebalancing rules---and thinnest where portfolio
management appears only as a conceptual layer.

\paragraph{Functional Scope and Allocation Paradigms}
Portfolio management represents the functional layer where individual asset signals are synthesized into constrained, multi-asset capital allocations. While single-asset trading models emit directional buy/sell signals in isolation, portfolio-level agents must account for cross-asset covariance, leverage limits, sector exposure budgets, and rebalancing turnover. Across the surveyed literature, portfolio agents operate through three primary approaches:
\begin{itemize}
	\item \textbf{Strategic Asset Allocation} (\Cref{ssec:allocation}): Synthesizing macro market regimes and factor signals into baseline target asset distributions.
	\item \textbf{Tactical Position Sizing} (\Cref{ssec:sizing}): Dynamically modulating individual asset exposures based on volatility, conviction, and liquidity.
	\item \textbf{Dynamic Rebalancing} (\Cref{ssec:rebalancing}): Managing execution turnover and rebalancing intervals under transaction-cost and market-friction constraints.
\end{itemize}
Key empirical considerations include the specification of rebalancing calendars, explicit universe filtering (accounting for delistings and liquidity), and turnover-induced transaction cost modeling.

\begin{tcolorbox}[
		colback=gray!10,
		colframe=gray!50,
		colbacktitle=gray!30,
		coltitle=black,
		fonttitle=\bfseries\sffamily\small,
		title={Classical Baseline: Mean-Variance Optimization and Kelly Criterion},
		boxrule=0.5pt,
		arc=2pt,
		left=5pt,
		right=5pt,
		top=3pt,
		bottom=3pt
	]
	\small
	Classical portfolio theory casts allocation as mean-variance optimization
	(Markowitz 1952), which maximizes risk-adjusted returns given covariance
	estimates, or as growth-optimal betting (Kelly 1956), which maximizes the
	log-utility of wealth. The Black-Litterman model incorporates investor views
	into equilibrium returns. These frameworks yield closed-form solutions under
	Gaussian assumptions but struggle with non-stationary higher-order moments and
	tail risks---precisely where LLM-based approaches may add value through
	adaptive reasoning and regime-aware dynamic allocation.
\end{tcolorbox}

\paragraph{Comparability note} Protocol-comparable portfolio claims typically report MR-1 (data/universe), MR-2 (time split), MR-3 (I/O contracts + constraints), and MR-4 (execution \& costs); see \Cref{sec:challenges}.


\subsection{Asset Allocation}
\label{ssec:allocation}

Asset allocation sets portfolio weights across asset classes and broad themes
(sectors, regions, styles), shaping the long-horizon risk/return envelope.

\textbf{Representative Approaches}. Classical baselines include mean-variance
optimization \cite{markowitz1952}~[BG] and Black-Litterman \cite{black1992}~[BG]; both hinge
on estimates (returns, covariances, or implied priors) that can drift quickly.
Risk-based allocations (risk parity / HRP-style clustering) are often preferred
when estimates are noisy. In agentic systems, LLMs commonly act as \emph{view
generators} \cite{lee2025llmbl,kirtac2025sentiment} conditioned on news and macro
releases. \citet{lee2025llmbl} show how LLM-generated views can be fed
into the Black-Litterman framework, while \citet{kirtac2025sentiment}
use LLM-based sentiment analysis to drive portfolio optimization decisions.

A parallel line of work pursues end-to-end neural allocation at scale:
end-to-end networks have been proposed for variance minimization in large
portfolios \cite{bongiorno2025endtoend}, and deep architectures combining
3D-CNNs with BiLSTMs target medium-term covariance forecasting
\cite{reis2025covariance}. Graph attention-based heterogeneous multi-agent
reinforcement learning \cite{graph_attention_portfolio_2025} offers adaptive
portfolio optimization through multi-agent coordination, and pair-trading work
that couples graph attention networks for pair selection with deep
reinforcement learning for execution provides a non-LLM background example of
relation-aware portfolio construction \cite{xia2025hierarchicalpairtrading}~[BG].
HARLF \cite{harlf_2025} integrates hierarchical reinforcement learning with
lightweight LLM-driven sentiment analysis for financial portfolio optimization.
These neural approaches complement mean-variance optimization by capturing
non-linear dependencies that classical methods may miss
\cite{zouaoui2024mptlstm,uysal2024riskbudgeting}.

These pipelines, however, introduce severe leakage risks wherever they ingest
text and macro data. The time-series evaluation literature therefore requires
\textbf{Point-in-Time (PIT) Textual Alignment}: macro inputs must use the
unrevised values available at the exact release timestamp, and validation
pipelines must prevent future information from leaking through preprocessing or
feature construction \cite{dataleakage2022prediction,lstm2025leakage}~[BG]. For the
same reason, \emph{regime tagging} is best generated online from $t-k$ lookback
windows, since full-sample regime clustering can leak future volatility states
\cite{hamilton1989new,bai2003multiple}~[BG].

\textbf{Technical Mechanisms}. HRP-style allocation reduces sensitivity to
covariance noise by allocating at the cluster level before distributing within
clusters. Regime-aware allocation (statistical or LLM-driven) can switch risk
budgets when volatility/liquidity conditions change, but it is best treated as
a hypothesis to test rather than a guarantee. For agentic trading, the practical
core is constraint-aware optimization (exposure, leverage, turnover) with
explicit transaction-cost assumptions \cite{lopezdeprado2016hrp,ledoit2004honey,jagannathan2003constraints}~[BG].

\subsection{Position Sizing}
\label{ssec:sizing}

Position sizing determines how much capital to allocate to each trading signal,
balancing the upside of strong signals against the risk of outsized losses.
This tactical decision directly shapes portfolio volatility and drawdown.

\textbf{Representative Approaches}. The Kelly criterion \cite{kelly1956} [BG] links size
to estimated edge and uncertainty, but is fragile to estimation error; most
implementations therefore use fractional Kelly. Recent work revisits portfolio
optimization from complementary angles \cite{boyd2024markowitz70,salo2024fiftyyears}~[BG], highlighting the growing importance of
constraint-aware sizing in modern markets.
Alternatives include risk-budget schemes (risk parity-style) and decision-informed
neural methods \cite{dinn2025portfolio} that integrate learned forecasts and LLM
inputs into portfolio decisions.

Beyond these rules, recent formulations widen the sizing design space. ODE-based
dynamic optimization \cite{alkhudaydi2025ode} offers continuous-time solutions to
the mean-variance problem, while cardinality- and bound-constrained formulations
\cite{li2024cardinality,caner2024tracking} address practical trading constraints
that theoretical models often ignore. On top of these formulations, agentic
systems may inject qualitative risk flags extracted from text; such inputs,
however, require strict time alignment and auditability.

\textbf{Technical Mechanisms}. In practice, sizing layers use conservative shrunk
estimates and fractional Kelly-style scaling to keep worst-case losses bounded.
\begin{tcolorbox}[title=Protocol Implication]
	For comparability, we recommend reporting a \textbf{Position Sizing Checklist}: (1) \textbf{Risk Metric}: What is minimized? (Vol, CVaR, MaxDD), (2) \textbf{Constraints}: What are the boundaries? (Long-only, Leverage $< 1.5$, Sector $< 20\%$), and (3) \textbf{Scaling Logic}: How is conviction mapped to size? (Fractional Kelly, Volatility Targeting).
\end{tcolorbox}
When LLMs contribute qualitative risk signals, they can be treated as features with strict timestamping and ablation tests (text signal on/off) to verify incremental value \cite{kelly1956,maclean2011kelly,rockafellar2000optimization}~[BG].

\subsection{Rebalancing}
\label{ssec:rebalancing}

Rebalancing adjusts portfolio weights to maintain target allocations as market
prices move and new signals arrive. This operational decision trades the
benefit of preserving the intended risk profile against the cost of frequent
trading.

Across the mapped literature, agentic portfolio construction and allocation
already appear in several applied settings. Agentic AI systems have been
applied to dynamic crypto-asset
allocation, where adaptability matters most under high volatility~
\cite{castelli2025building}; as background examples, generative-AI pipelines
construct sector portfolios
from LLM selections combined with classical optimization~
\cite{voronina2025generative}~[BG], and PolyModel theory coupled with iTransformer
architectures targets hedge-fund portfolio construction~
\cite{zhao2024hedge}~[BG]. On the transparency side, explainable DRL techniques aim
to make financial policy decisions interpretable~
\cite{escudero2024explainabl}~[BG]. Strategy-guided exploration mechanisms can
broaden agent robustness for portfolio workflows, though they are best read
here as a generic robustness idea rather than direct evidence for rebalancing~
\cite{szot2026expanding}~[BG].

\textbf{Representative Approaches}. Canonical policies include threshold rebalancing (act only when weights drift beyond bands), time-based schedules, and cost-aware optimization that trades off tracking error vs. transaction/impact/tax costs.

Learning-based methods then refine how these policies are decided. Heuristic-guided reward learning has been proposed for portfolio optimization, showing how learned heuristics can improve rebalancing decisions \cite{ijcai2025heuristic}, and multi-objective gradient descent enables simultaneous optimization of competing portfolio objectives during rebalancing \cite{multobj2024gradient}.

Agentic workflows often add \emph{event awareness} on top of these schedules (e.g., going risk-off before FOMC), which creates a second leakage surface. Comparisons are most credible when evaluations distinguish between \textbf{Calendar Events} (scheduled ahead of time, e.g., Earnings Calendar) and \textbf{Shock Events} (unexpected news): rebalancing driven by shocks should rely on real-time proxies (e.g., VIX spike, spread widening) observable at $t$, rather than future-annotated event labels \cite{tradinggpt2023,unnikrishnan2024newsrl}. FinVision \cite{finvision2024}~[BG] is better read as a multimodal stock-prediction example than as direct evidence for this taxonomy.

News-driven rebalancing has emerged as a particularly active application: one
line of work combines financial news analysis with reinforcement learning for
portfolio management, enabling dynamic rebalancing in response to real-time
sentiment shifts \cite{unnikrishnan2024newsrl}. Liquidity-aware selection and
rebalancing strategies \cite{abensur2024liquidity,sharpe2025liquidity} further
incorporate machine learning-based liquidity classification to guide allocation
and execution timing.

\textbf{Technical Mechanisms}. No-trade regions (bands) reduce churn by ignoring
small drift until it matters for risk, while execution models (VWAP/TWAP/implementation
shortfall (IS); \Cref{ssec:execution}) and
explicit slippage assumptions determine whether a policy survives outside a
backtest. For agentic trading, the key protocol caveat is separating \emph{decision
	timing} (signal/optimizer) from \emph{fill timing} (execution) to avoid
overstating performance. Where tax-aware logic is included, the evaluation
should match the account type and jurisdiction that the tax model assumes
\cite{constantinides1986capital,davis1990portfolio,almgren2001optimal,perold1988shortfall}~[BG].

\begin{table*}[!tb]
	\centering
	\caption{Portfolio management method comparison: traditional methods, candidate LLM enhancements, and key challenges at each decision level.}
	\label{tab:portfolio-comparison}
	\begin{tabular}{lp{4cm}p{4.5cm}p{3cm}}
		\toprule
		Level            & Traditional Methods                                                               & LLM Enhancement                                                                   & Key Challenges                                                \\
		\midrule
		Asset Allocation & Mean-variance optimization, Black-Litterman, Risk parity, Hierarchical clustering & Regime detection from narratives, View generation, Market sentiment analysis      & Covariance estimation errors, Regime misclassification        \\
		\midrule
		Position Sizing  & Kelly criterion, Risk parity equalization, Convex optimization (VaR/CVaR)         & Qualitative risk assessment from news, Constraint handling with reasoning         & Estimation error sensitivity, Over-aggressive sizing          \\
		\midrule
		Rebalancing      & Threshold-based, Time-based, Cost-aware optimization                              & Strategic timing around events, Transaction cost estimation, Tax-aware harvesting & Transaction costs, Market impact, Tax optimization complexity \\
		\bottomrule
	\end{tabular}
\end{table*}

\textbf{Extension-window additions.} Building on agentic crypto-asset
allocation \cite{castelli2025building}, additional studies include screening pipelines that
feed portfolio construction \cite{caner2026designing}, LLM-driven
small-capitalization trading with integrated allocation
\cite{kargarzadeh2026large}, recursively self-optimizing multi-agent portfolio
systems \cite{yang2026recursive}, and a market-regime council that performs
dynamic credit assignment across specialized agents \cite{pei2026market}. These
systems make the portfolio layer an explicit, inspectable component, though
their evaluation conventions (rebalancing frequency, constraints, cost models)
remain heterogeneous.

\paragraph{Synthesis and Practical Considerations}
Portfolio management translates alpha into diversified holdings under allocation, sizing, and rebalancing constraints. LLM-generated views can be paired with classical optimization to connect qualitative context with quantitative risk bounds; credible evaluation depends on reporting the asset universe, rebalancing schedule, and turnover costs. \Cref{sec:risk} examines controls across the trading lifecycle.


\section{Risk Management Capability}
\label{sec:risk}

The preceding sections examined alpha generation and portfolio management. This
section addresses the question: \textquotedblleft How do agents control
financial risks?\textquotedblright

We treat risk management here as the set of constraints, monitors, and
post-trade controls that shape allowable actions. The discussion stays tied to
the evidence base because many risk mechanisms are specified at the
architecture level but rarely isolated in empirical evaluation.

\paragraph{Systemic Role and Risk Governance Architecture}
Risk management forms the essential protective foundation of an autonomous trading system, ensuring capital preservation, regulatory compliance, and resilience against market shocks. Because LLMs can exhibit hallucination, overconfidence, or vulnerability to adversarial input, real-world deployment requires independent, deterministically governed risk guardrails. Across the trading lifecycle, risk architecture is structured across three defensive stages:
\begin{itemize}
	\item \textbf{Pre-Trade Risk Controls} (\Cref{ssec:pretrade}): Hard gatekeeping filters verifying instrument whitelists, maximum order sizes, leverage ceilings, and liquidity availability prior to order submission.
	\item \textbf{In-Trade Monitoring and Dynamic Exposure} (\Cref{ssec:realtime}): Real-time surveillance tracking aggregate portfolio value-at-risk (VaR), sector concentration limits, trailing stop-losses, and emergency circuit breakers.
	\item \textbf{Post-Trade Attribution and Compliance} (\Cref{ssec:posttrade}): Systematic trade reconciliation, slippage analysis, performance attribution, and regulatory audit logging.
\end{itemize}
From an engineering standpoint, the key challenge lies in establishing non-bypassable, deterministic overrides that can intercept errant LLM actions before market transmission without introducing unacceptable execution delays.
These safety mechanisms apply across the pre-trade, in-trade, and post-trade stages.
The stages are detailed in \Cref{ssec:pretrade}, \Cref{ssec:realtime}, and
\Cref{ssec:posttrade}; cross-stage guardrails are discussed in
\Cref{ssec:safety-mechanisms}.

\begin{tcolorbox}[
		colback=gray!10,
		colframe=gray!50,
		colbacktitle=gray!30,
		coltitle=black,
		fonttitle=\bfseries\sffamily\small,
		title={Classical Baseline: VaR, CVaR, and Drawdown Control},
		boxrule=0.5pt,
		arc=2pt,
		left=5pt,
		right=5pt,
		top=3pt,
		bottom=3pt
	]
	\small
	Traditional risk management quantifies tail risk through Value-at-Risk (VaR)---the
	quantile of the loss distribution---and Conditional VaR (CVaR), the expected
	loss beyond VaR. Drawdown control limits peak-to-trough declines via dynamic
	programming or convex optimization. These metrics provide computationally
	tractable risk bounds but assume stationary distributions and known correlation
	structures. LLM agents can instead produce natural-language explanations of
	risk exposures and generate scenarios; to reach an auditability standard
	comparable to classical frameworks, however, those explanations must be
	grounded in verifiable metrics and immutable logs
	\cite{khatchadourian2026replayable}~[BG].
\end{tcolorbox}

\begin{figure*}[!tb]
	\centering
	\includegraphics[width=0.96\linewidth]{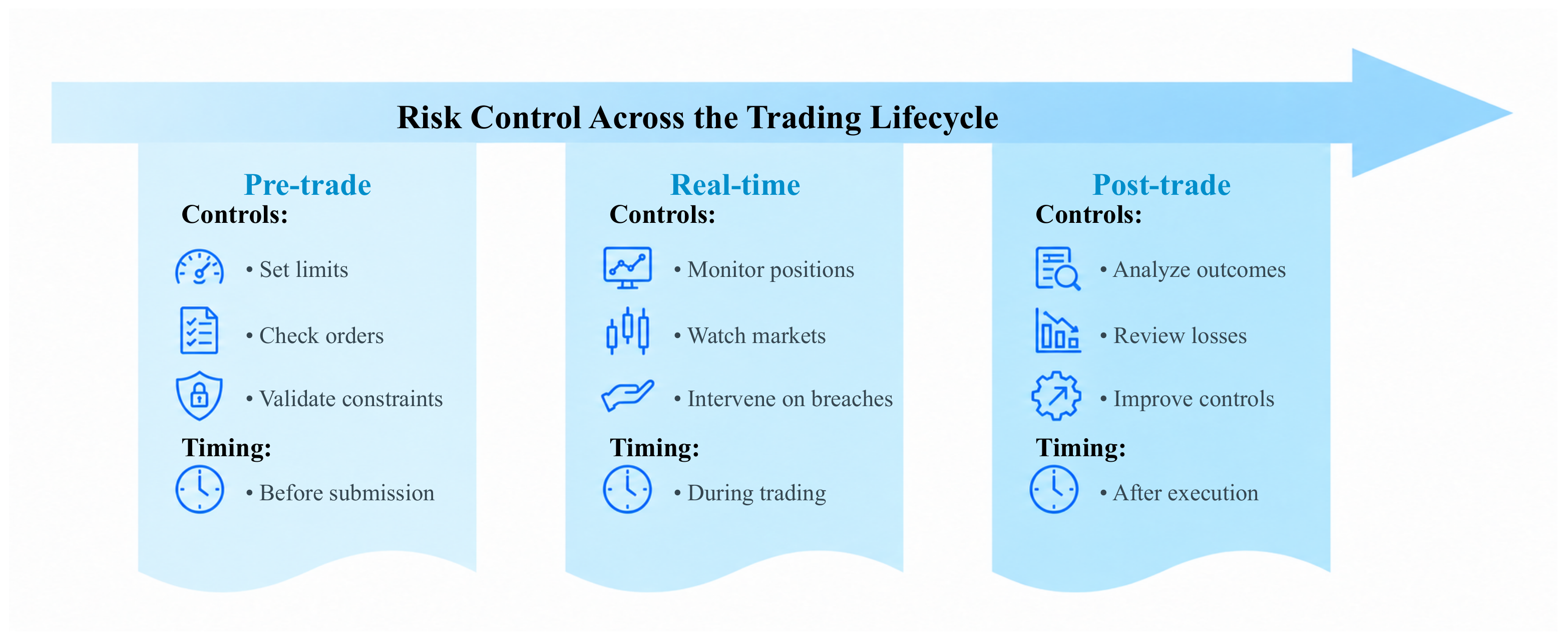}
	\caption{Risk control timeline across the trading lifecycle. Pre-trade risk
		control sets limits and checks before orders are submitted. Real-time risk
		control monitors positions and markets during trading, intervening when
		thresholds are breached. Post-trade risk analysis analyzes outcomes to
		improve future risk management.}
	\label{fig:risk-timeline}
\end{figure*}

\Cref{fig:risk-timeline} illustrates the lifecycle overview.

\paragraph{Comparability note} Protocol-comparable risk claims typically report MR-3 (risk action semantics + priorities), MR-4 (execution/cost assumptions that risk controls depend on), and MR-6 (artifacts \& logs); see \Cref{sec:challenges}.

\subsection{Pre-trade Risk Control}
\label{ssec:pretrade}

Pre-trade risk control prevents excessive risk by checking orders against risk
limits before they are submitted to the market. This preventive stage is the
first line of defense against catastrophic losses.

\textbf{Representative Approaches}. Value-at-Risk (VaR) limits cap the maximum
potential loss at a specified confidence level (e.g., 99\% one-day VaR of \$1M).
Conditional VaR (CVaR) extends this by estimating the expected loss in the
tail beyond the VaR threshold \cite{rockafellar2000optimization}~[BG].
Exposure constraints limit concentrations to
single assets, sectors, or counterparties (e.g., no more than 10\% of portfolio
in any single stock), and compliance checks ensure that trades respect
regulatory requirements and internal policies. On the agentic side, some
prototypes use LLMs to translate human-readable policies into executable
constraints \cite{cosler2023nl2spec}~[BG], while others employ structured,
model-based risk representations to estimate market risk transparently. This
translation step, however, injects non-determinism into a layer that is
expected to behave deterministically---a tension the protocol implication below
addresses directly.
\begin{tcolorbox}[title=Protocol Implication]
	For protocol comparability, we recommend reporting: (1) \textbf{VaR Config} (Confidence Level, Holding Period, Window), (2) \textbf{Scenario Source} (Standard vs Custom), and (3) \textbf{LLM Verification}: policy-to-code translations are best validated against deterministic test sets rather than raw LLM chat outputs for compliance.
\end{tcolorbox}

\textbf{Technical Mechanisms}. Portfolio risk models estimate the risk
contribution of each potential trade, considering both the standalone risk of
the position and its correlation with existing holdings. Let $\Sigma$ denote the
covariance matrix of asset returns with entries $\sigma_{ij}$, and let $w$ denote
portfolio weights. The portfolio variance is $V = w^\top \Sigma w$, and the
\textit{marginal contribution to variance} is
$\frac{\partial V}{\partial w_i} = 2(\Sigma w)_i$. A commonly used
\textit{risk contribution} decomposition is $RC_i = w_i(\Sigma w)_i$, while the
\textit{marginal contribution to volatility} is
$\frac{\partial \sigma_p}{\partial w_i} = (\Sigma w)_i / \sigma_p$ with
$\sigma_p=\sqrt{w^\top \Sigma w}$ \cite{roncalli2013risk}~[BG].
Scenario analysis simulates portfolio performance under stress scenarios
(market crash: -30\% equities, interest rate spike: +200bps, volatility surge:
$\sigma \times 3$) to identify hidden risks that might not manifest under normal
conditions \cite{bcbs2009stress}. Rule engines encode regulatory and internal policies as executable
rules (e.g., Reg T margin requirements, Reg NMS order protection rules); here
LLMs can parse natural-language policies into such rules automatically, which
is precisely where the non-determinism caveat above applies.

Adjacent work supports this layer at two different levels. Agent-based market
models generate realistic transaction costs and slippage for liquidity risk
assessment~
\cite{vytelingum2025agent}~[BG], while the STRIDE framework offers guidance on when a
risk-sensitive application calls for agentic AI, an AI assistant, or a direct
LLM call~
\cite{asthana2025stride}~[BG].

\subsection{Real-time Risk Control}
\label{ssec:realtime}

Real-time risk control monitors positions and market conditions during trading,
intervening when risk thresholds are breached or anomalous conditions emerge.
This reactive stage catches risks that pre-trade models miss due to changing
market conditions or model error.

\textbf{Representative Approaches}. Position monitoring tracks portfolio metrics
(greeks for options, duration for bonds, beta for equities) in real time, alerting
when they exceed thresholds. Drawdown stops halt trading when portfolio losses
reach specified levels (e.g., stop trading if down 5\% from peak), and volatility
scaling reduces position sizes as market volatility increases, maintaining
constant risk exposure. Text streams can extend this monitoring with early
warnings \cite{tetlock2007giving}~[BG]; slower-moving macro uncertainty indices
complement them \cite{baker2016measuring}~[BG], though these indices are not
real-time signals.
Because real-time control is also a first-order design concern for cross-market
robustness, the division of labor matters: LLMs are generally unsuitable for the
sub-millisecond \textbf{Critical Path}, so a robust architecture separates a
\textbf{Gating Layer} (deterministic algorithms such as circuit breakers for
sub-millisecond stops) from an asynchronous \textbf{Explanation Layer} (LLMs for
cause attribution). Evaluations should then disclose \textbf{Latency Budgets} and
\textbf{False Positive Rates} for all triggers.

\textbf{Technical Mechanisms}. Stream processing systems update risk metrics
from tick data, trades, and order-book changes. Anomaly detection can flag price
spikes, volume surges, and order-book imbalance as stress signals
\cite{easley2012flow,kirilenko2017flash}~[BG]. Circuit breakers halt trading when
drawdown or single-position loss limits are breached. Manual review before
restart helps prevent stress-driven decisions
\cite{lo2002psychophysiology,kahneman1979prospect}~[BG].

\begin{tcolorbox}[title=Protocol Implication]
	We suggest reporting any dynamic exposure rule in display form rather
	than embedding it in prose. A simple example is
		\[
			w_i(t)=
			\min\!\left(
			w_i^{\max},
			\frac{\mathrm{risk\_budget}}{\mathrm{risk\_per\_unit}_i}
			\right)
		\]
	which caps concentration as prices and volatility change.
\end{tcolorbox}

\subsection{Post-trade Risk Analysis}
\label{ssec:posttrade}

Post-trade risk analysis learns from trading outcomes to improve future risk
management. This retrospective stage identifies what risks were missed, how
effective controls were, and where improvements are needed.

\textbf{Representative Approaches}. Performance attribution decomposes
returns into component sources (alpha return, beta, timing, stock selection), identifying
which risks contributed to gains or losses; LLM-assisted attribution can narrate
how market factors propagated through the portfolio structure. Stress testing simulates portfolio
performance under historical crisis scenarios (2008 financial crisis and other major stress episodes) to assess resilience \cite{bcbs2009stress}. Regime analysis examines how portfolio performance
varied across market regimes (bull, bear, high volatility), identifying
conditions where risk management failed. LLM-enhanced systems can summarize
these risk outcomes, but guarding against post-hoc rationalization requires
every narrative claim to be grounded in immutable logs and replayable evidence
\cite{tan2025fred,khatchadourian2026replayable}~[BG]. For the same reason,
\textbf{Counterfactual Reasoning} (e.g., "what if" scenarios) is best labeled
\textit{Hypothetical Inference}---an assumption-dependent construction, not an
empirical fact.

\textbf{Technical Mechanisms}. Causal analysis identifies the root causes of
risk events, distinguishing between market risk (systemic factors affecting all
assets), idiosyncratic risk (asset-specific events), and model error (incorrect
assumptions or predictions). Techniques include Granger causality tests, Shapley
value decomposition for attribution, and structural break detection to identify
when risk models failed. Counterfactual reasoning estimates what would have
happened under different risk management decisions (as an illustrative example,
``if we had reduced position size by 50\%, drawdown would have been 3\% instead of 8\%''), enabling learning
from near-misses. Report generation produces human-readable risk
summaries for compliance and management review, with LLMs translating quantitative
metrics (``VaR exceeded by 20\% on March 15th'') into narrative explanations
(``Risk was elevated due to concentrated tech exposure during the AI sector
correction; position limits were triggered but too late to prevent significant
losses''), making risk analysis interpretable for non-technical stakeholders
\cite{granger1969investigating,shapley1953value,bai2003multiple,pearl2009causality}~[BG].

\begin{table*}[!tb]
	\centering
	\caption{Risk management method comparison: mechanisms and candidate LLM enhancements across the three lifecycle stages.}
	\label{tab:risk-comparison}
	\small
	\setlength{\tabcolsep}{4pt}
	\renewcommand{\arraystretch}{1.1}
	\begin{tabularx}{\linewidth}{@{}p{1.4cm} >{\raggedright\arraybackslash}X >{\raggedright\arraybackslash}X@{}}
		\toprule
		Stage      & Mechanisms                                               & LLM Enhancement                                                       \\
		\midrule
		Pre-trade  & VaR/CVaR limits; exposure constraints; compliance checks & Rule interpretation; policy translation; natural-language constraints \\
		Real-time  & Position monitoring; drawdown stops; volatility scaling  & Anomaly detection; news monitoring; early-warning signals             \\
		Post-trade & Attribution; stress testing; regime analysis             & Narrative explanation; report generation; learning insights           \\
		\bottomrule
	\end{tabularx}
\end{table*}

\Cref{tab:risk-comparison} summarizes risk management approaches across the
three stages. Effective risk management depends on all three stages working in
concert: pre-trade controls prevent excessive risk, real-time controls catch
emerging risks, and post-trade analysis enables continuous improvement.

\subsection{Safety and Constraint Mechanisms}
\label{ssec:safety-mechanisms}

Deployable trading agents require safety layers that sit outside the model's generative reasoning loop. These include pre-trade exposure checks, hard position limits, concentration caps, drawdown triggers, and kill-switch logic that can halt order flow when market conditions or model behavior leave the approved operating envelope. In the current literature, such mechanisms are often mentioned in passing but rarely specified at reproducible detail. We therefore treat safety architecture as part of the risk-management interface, not as an optional deployment afterthought.

\textbf{Extension-window additions.} The extension cohort begins to fill this
gap: parameterized stop-loss/take-profit layers for autonomous strategies
\cite{li2026optimal} make exit-rule design an explicit, tunable risk interface,
and dynamic objective selection under safeguard constraints with LLM oversight
\cite{sakurai2026dynamic} treats risk bounds as first-class control inputs.
A complementary background analysis systematizes how trading agents fail under
adversarial and regime-shifted conditions \cite{wang2026sok}~[BG]; it is not
primary evidence for this gap.



\paragraph{Synthesis and Governance Considerations}
Risk governance spans pre-trade limits, in-trade monitoring, and post-trade analysis, with deterministic safety mechanisms constraining actions across all three stages. FinRS \cite{liu2025risksensit} incorporates risk sensitivity into the trading objective; dedicated limits, gating, and monitoring provide complementary safeguards at the execution and governance layers. Part~III turns to adaptation and evolution, beginning with adaptive learning paradigms in \Cref{sec:learning}.


\section{Learning Paradigms}
\label{sec:learning}
\label{part:adaptation}

The preceding parts of this survey established the architectural foundations of
agentic traders and the capabilities these architectures enable. This section
addresses the question: \textquotedblleft How do agents learn from market
feedback?\textquotedblright

Learning enables agents to adapt their behavior based on experience and changing
market conditions. This section covers learning mechanisms; governance of
persistent state and update cycles is treated in
\Cref{sec:evolution}.

\paragraph{Adaptive Mechanisms and Learning Spectrum}
Financial markets are intrinsically non-stationary: statistical patterns decay, volatility regimes shift abruptly, and trading signals erode as market participants adapt. Consequently, static trading agents face rapid performance degradation unless equipped with mechanisms for continuous adaptation. Across the surveyed literature, agent learning spans three foundational paradigms along the spectrum of parameter persistence and adaptation speed:
\begin{itemize}
	\item \textbf{In-Context Learning (ICL)} (\Cref{ssec:icl}): Adaptation through prompts that conditions future decisions on retrieved market exemplars and reflective feedback without modifying model weights.
	\item \textbf{Supervised Fine-Tuning (SFT)} (\Cref{ssec:sft}): Domain adaptation through weight updates that trains base LLMs on specialized financial instructions, historical trade rationales, and domain corpora.
	\item \textbf{Reinforcement Learning (RL)} (\Cref{ssec:rl}): Policy optimization that trains agents to maximize risk-adjusted sequential rewards (e.g., cumulative return, Sharpe ratio) within interactive trading environments.
\end{itemize}
The primary methodological trade-offs involve balancing sample efficiency and adaptation agility against catastrophic forgetting and distribution-shift vulnerability.

\subsection{Reward and Feedback Design}
\label{ssec:reward-feedback}

Reward specification is a first-order design choice rather than an implementation detail. Trading environments impose delayed, noisy, and partially confounded feedback: realized P\&L arrives after execution, inventory risk accumulates across steps, and reward proxies can be gamed through simulator artifacts. We therefore distinguish immediate execution feedback (fill quality, slippage, rejected orders) from delayed portfolio feedback (P\&L, drawdown, turnover, and risk-adjusted return). Papers claiming adaptive improvement should report which of these signals are optimized, over what horizon, and with what safeguards against reward hacking.

\begin{figure*}[!tb]
	\centering
	\includegraphics[width=\linewidth]{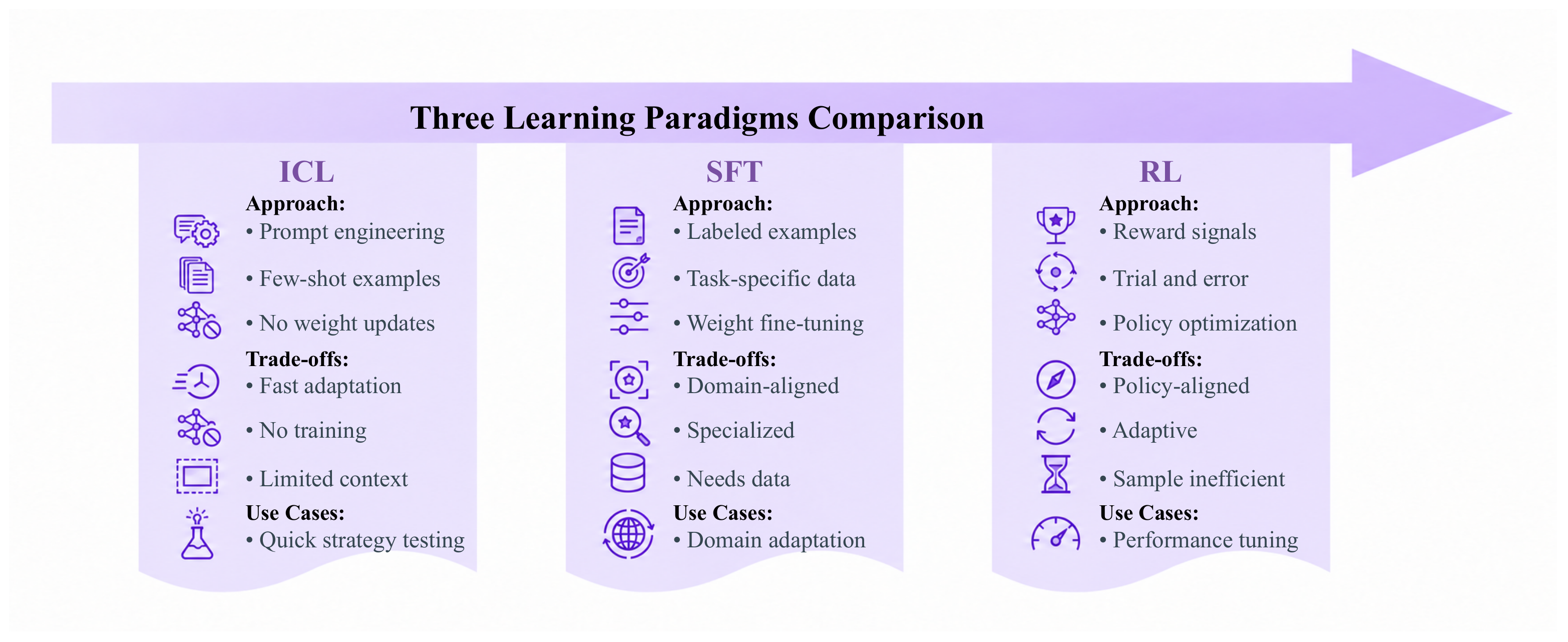}
	\caption{Conceptual comparison of three learning paradigms. The panels
		summarize what each loop updates, the main trade-offs emphasized in the
		design literature, and illustrative use cases.}
	\label{fig:learning-paradigms}
\end{figure*}

Figure~\ref{fig:learning-paradigms} provides a conceptual comparison of the three paradigms.

\subsection{In-context Learning}
\label{ssec:icl}

In-context learning (ICL) enables agents to adapt from examples provided in the
prompt without updating model parameters, following the GPT-3-style few-shot
paradigm \cite{brown2020gpt3}~[BG]. This rapid adaptation mechanism is particularly
valuable in dynamic markets where conditions change faster than models can be
retrained.

\textbf{Representative Approaches}. FinAgent \cite{wu2024finagent} uses ICL by
retrieving relevant trading episodes from memory and inserting them as examples
in the prompt, enabling the LLM to adapt its reasoning to similar situations
without parameter updates.

\textbf{Technical Mechanisms}. ICL selects and orders examples for the prompt,
using retrieval and context-management choices described in
\Cref{sec:memory} \cite{brown2020gpt3}~[BG], \cite{wu2024finagent}. RAG also
adds current financial documents from external stores. Because neither updates
model parameters, any adaptation is temporary and bounded by the context window.

\subsection{Supervised Fine-tuning}
\label{ssec:sft}

Supervised fine-tuning (SFT) updates model parameters on labeled demonstrations
of desired behavior, teaching agents to imitate expert trading decisions. This
approach transfers knowledge from expert traders or historical successful
strategies into the agent.

\textbf{Representative Approaches}. The examples below distinguish finance-domain tuning methods from
end-to-end trading systems. The extension
cohort begins to move this boundary: post-training of local quantitative
research agents \cite{wang2026alphadiverse} and cognitive fine-tuning for
stable financial reasoning \cite{pan2026learning} bring weight updates closer
to the trading loop, though fully end-to-end SFT evidence for trading agents
remains thin. The FinGPT instruction-tuning benchmark \cite{fingpt_2023}~[BG]
illustrates finance-domain instruction tuning on financial NLP tasks. Two supervised variants matter most for trading agents: \emph{behavior
cloning}, which maps market states to expert actions so that trader
demonstrations become usable policy targets \cite{ross2011dagger}~[BG]; and
\emph{multi-task adaptation}, which trains on sentiment, portfolio, and risk
tasks jointly and produces representations that may transfer across finance
subtasks \cite{caruana1997multitask}~[BG].

\textbf{Technical Mechanisms}. SFT relies on instruction and demonstration
datasets with prompt-response pairs (market context $\rightarrow$ expert
decision or analysis).
Training minimizes a supervised loss (e.g., cross-entropy for discrete actions,
regression for sizing/allocation), with standard regularization and careful
held-out evaluation to check robustness to unseen regimes. Parameter-efficient
methods (e.g., LoRA-style methods) reduce cost while preserving much of the
base model; finance-oriented benchmarking work has explored these trade-offs
\cite{finlora_benchmark_2025}. General parameter-efficient variants are background
mechanisms \cite{qlora_2023,qalora_2023}~[BG]. SFT can produce persistent
behavioral changes but is sensitive to label quality and distribution shift.

\subsection{Reinforcement Learning}
\label{ssec:rl}

Reinforcement learning (RL) is used in trading agents when the objective is
inherently sequential (portfolio rebalancing, execution, or iterative factor
construction), the feedback is delayed, and the policy needs to trade off return,
risk, and costs under constraints.

\textbf{Representative Approaches}. FinRL-DeepSeek \cite{benhenda2025finrl} combines
LLM-based priors with policy optimization, reporting risk-adjusted return improvements
over baselines in simulated environments\footnote{Specific
	performance metrics vary by configuration (e.g., CPPO-DeepSharpe vs. CPPO-DeepVol)
	and market regime; readers should consult the original paper for detailed
	protocol assumptions, cost models, and train/test split discipline.}. MacroHFT \cite{zong2024macrohft}
introduces memory-augmented context-aware RL for high-frequency trading, while
EarnHFT \cite{qin2024earnhft} explores hierarchical RL for HFT; both are
conventional deep-RL trading systems rather than LLM agents and are retained
as non-agentic baselines.
Deep RL for alpha discovery \cite{alpha2_2024} frames factor construction as a
program-generation task, optimizing an evaluation metric over the search space
while balancing alpha performance and diversity.

Adjacent work marks the boundary between learning rules and persistent
cross-time change: modular memory architectures retain experience across tasks
\cite{dorovatas2026modular}~[BG], explicit state dynamics improve temporal
coherence \cite{subaharan2026controlling}~[BG], and code-based alpha evolution
refines strategies from execution traces and natural-language feedback
\cite{cogalpha_2025}~[BG]. The latter update-governance questions are taken up
in \Cref{sec:evolution}.

\begin{table*}[!tb]
	\centering
	\small
	\setlength{\tabcolsep}{3pt}
	\caption{Conceptual design trade-offs of learning paradigms. These descriptions
        characterize adaptation mechanisms;
        training cost and data requirements depend on the task and implementation.}
	\label{tab:learning-comparison}
	\begin{tabularx}{\linewidth}{@{}l p{2.6cm} p{2.5cm} X p{3.7cm}@{}}
		\toprule
		Paradigm & Adaptation mechanism & Data source & Typical strengths & Design considerations \\
		\midrule
		ICL & Context updates & Prompts and examples & Immediate task conditioning & Context capacity; no persistent weight update \\
		SFT & Supervised weight updates & Labeled demonstrations & Persistent domain specialization & Label coverage; retraining; overfitting \\
		RL & Reward-driven policy updates & Interactive trajectories & Optimizing sequential decisions & Simulator fidelity; reward design; sample cost \\
		\bottomrule
	\end{tabularx}
\end{table*}

\textbf{What should be reported (and why it matters)}. Because RL results are
highly sensitive to experimental setup, trading papers should explicitly specify
(i) chronological data splits (train/validation/test or walk-forward) and how
hyperparameters are chosen to avoid leakage; (ii) the cost model (commissions,
spread/slippage, and market impact assumptions); and (iii) simulator fidelity
(fill/latency assumptions, action/state/reward definitions, constraints, and
whether the simulator matches the instrument/microstructure being claimed).
Key risks include simulator overfitting and reward hacking: policies can exploit
artifacts of the backtester/cost model and fail catastrophically out-of-sample,
especially under non-stationarity and regime shifts.

\paragraph{Synthesis and Algorithmic Considerations}
In-context learning conditions decisions on prompt examples and retrieved experience; supervised fine-tuning updates parameters from demonstrations; reinforcement learning optimizes sequential policies through reward feedback. FinAgent illustrates prompt-context adaptation \cite{wu2024finagent}, while FinRL-DeepSeek and Alpha$^2$ use RL \cite{benhenda2025finrl,alpha2_2024}. These approaches differ in persistence and feedback, and their evaluation depends on example and label coverage, temporal splits, simulator fidelity, and reward design. \Cref{sec:multiagent} next examines collective coordination.


\section{Multi-Agent Coordination}
\label{sec:multiagent}

\textit{Note}: Multi-agent coordination is a cross-cutting organizational
dimension: role structure is architectural, while coordination dynamics can
shape capabilities and adaptation. This section focuses on those interactions
and appears in Part III because its emphasis is on emergent coordination.

The preceding sections examined the architecture and capabilities of agentic
traders. This section addresses the question: \textquotedblleft How do multiple
agents collaborate in markets?\textquotedblright

Multi-agent systems enable specialization and decomposition, with different
agents focusing on different aspects of the trading problem (analysis,
execution, risk management). What remains unsettled is whether coordination
itself delivers superior trading performance once computational budget,
information access, and protocol design are held constant.

\paragraph{Coordination Patterns and Market Ecology}
Financial decision-making in institutional environments is inherently collaborative, distributed across macroeconomic analysts, sector specialists, quantitative modelers, risk officers, and execution traders. Multi-agent trading frameworks mirror this division of labor by deploying specialized agents that interact, deliberate, and coordinate. In the literature, multi-agent organizations are structured along three principal paradigms:
\begin{itemize}
	\item \textbf{Role-Based Collaboration} (\Cref{ssec:role}): Flat or peer-to-peer agent topologies where specialized personas (e.g., fundamental bull/bear debaters, sentiment analysts) exchange structured messages to reach trading consensus.
	\item \textbf{Hierarchical Networks} (\Cref{ssec:hierarch}): Multi-tiered command structures where supervisory portfolio manager agents aggregate tactical recommendations from subordinate asset specialists.
	\item \textbf{Synthetic Market Ecologies} (\Cref{ssec:ecology}): Agent-based market simulations where heterogeneous populations of autonomous buyers, sellers, and liquidity providers interact, allowing the study of emergent price dynamics, market impact, and reflexivity.
\end{itemize}
The central research challenges involve attributing performance gains to specific communication topologies (via rigorous role ablations), managing communication latency overheads, and preventing groupthink or herd dynamics.

\paragraph{Comparability note} Protocol-comparable coordination claims should report MR-7 (roles/permissions, message protocol, consensus, shared state, and evaluation); see \Cref{sec:challenges}.

\begin{figure*}[!tb]
	\centering
	\includegraphics[width=\linewidth]{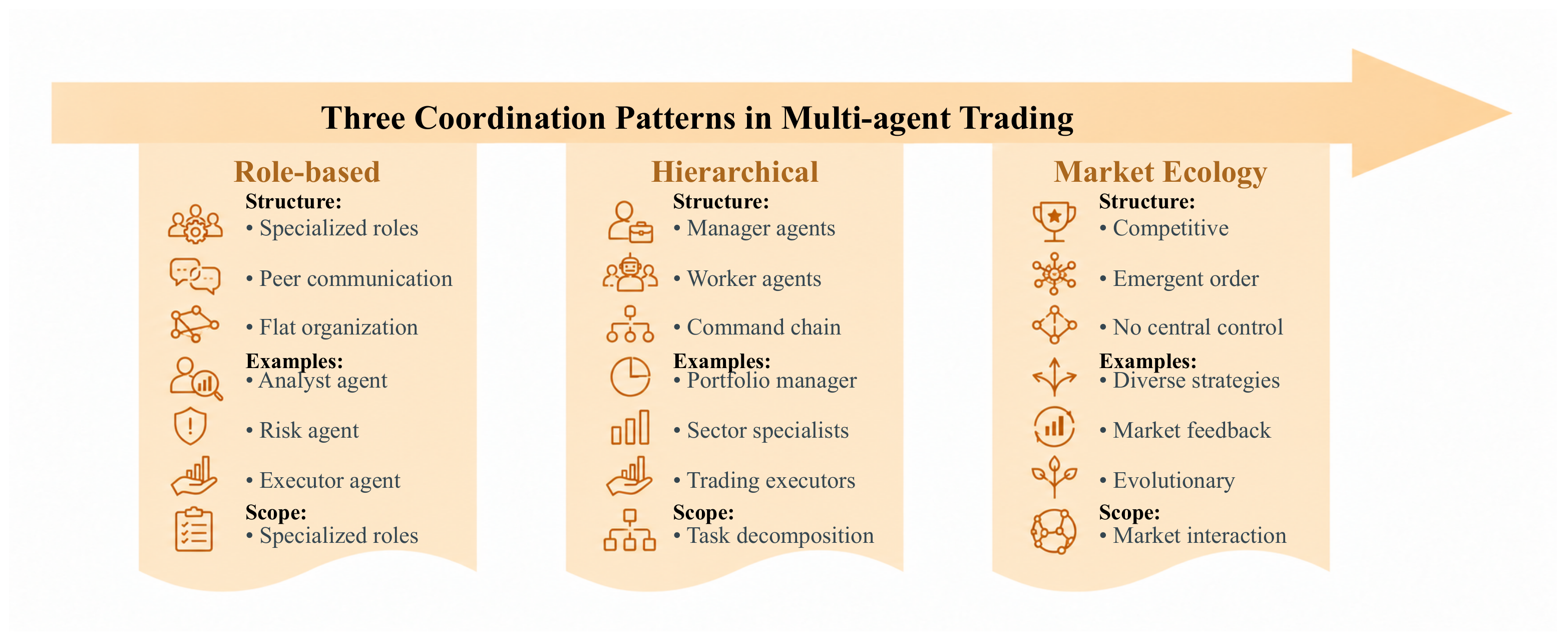}
	\caption{Three coordination patterns in multi-agent trading. Role-based
		collaboration assigns specialized roles (analyst, trader, risk manager) to
		different agents. Hierarchical organization layers agents from strategic
		(asset allocation) to tactical (position sizing) to execution (individual
		trades). Market ecology models agents interacting through markets with feedback
		and emergence. The structure/example/scope panels are stylized
		archetypes of these organizational designs.}
	\label{fig:coordination-patterns}
\end{figure*}

Figure~\ref{fig:coordination-patterns} illustrates these stylized archetypes.

\subsection{Role-based Collaboration}
\label{ssec:role}

Role-based collaboration assigns specialized roles to different agents, each
focusing on a specific aspect of the trading problem while coordinating to
achieve collective goals. This division of labor enables deep specialization
while maintaining coherent decision-making \cite{wooldridge2009multiagent}~[BG].

\textbf{Representative Approaches}. We distinguish three evidence
layers. \textit{Primary trading systems} in the current section-level subset
show role specialization most directly: TradingAgents \cite{xiao2024tradingagents}
assigns analyst/trader/risk/portfolio roles; FINCON \cite{yu2024fincon}
uses a manager--analyst architecture with Conceptual Verbal Reinforcement (CVRF) to update investment beliefs; TradingGroup
	\cite{tradinggroup_2025} combines role specialization with
reflection; and the REIT pipeline \cite{li2026reits_multiagent} makes an
analysis--prediction--decision--execution workflow explicit. These papers make
decomposition visible, yet complex role assignments can still obscure the source
of gains. Supporting \textit{attribution} therefore requires logging decision
contributions (which agent proposed or vetoed an action) and running
role/communication ablations that verify gains stem from coordination rather
than from increased computational budget.
General LLM multi-agent optimization work such as HiveMind supplies relevant
background on contribution-guided credit assignment and online prompt refinement
\cite{xia2026hivemind}~[BG].

\textit{Benchmarks and simulated arenas} form a second layer: TraderBench,
MarS, and related multi-agent market arenas
\cite{lopez-lira2025canlargela,li2024mars,yuan2026traderbench}~[BG] stress
specialization and interaction patterns under their respective simulation
protocols. Trading-performance comparisons across these layers require matched
execution semantics and computational budgets.

\textit{Background MAS/ABM literature} forms a third layer, including social or
behavioral market models \cite{bohorquez2024simulation,wheeler2023scalable}~[BG],
RL-plus-ABM interaction studies \cite{dicks2023many,vanOort2023adaptive}~[BG],
workspace delegation \cite{nie2026awcp}~[BG], and reciprocity-oriented foundations
\cite{diau2025finance,gao2022understanding}~[BG]. These works inform
mechanism design through models of delegation, reciprocity, and market interaction.

\textbf{Communication Protocols}. Role-based collaboration rests on communication
protocols that coordinate specialized agents effectively. Broadcasting allows an
agent to share
information with all team members simultaneously, which suits market-wide alerts
or portfolio status updates. Peer-to-peer communication enables direct exchange
between specific agents, such as the risk manager warning the trader about
excessive exposure. Shared memory provides a common knowledge base where agents
can read and write information, enabling asynchronous coordination without
requiring all agents to be available simultaneously. These protocols need to be
designed to balance information richness against communication overhead, as
excessive messaging can slow decision-making in fast-moving markets
\cite{wooldridge2009multiagent}~[BG].

\textbf{Consensus Mechanisms}. When specialized agents disagree, consensus mechanisms
aggregate diverse opinions into coherent decisions. Voting treats all agents equally,
with the majority or plurality determining the final action. Expert weights assign
different influence to agents based on their historical accuracy or domain expertise,
allowing the analyst to dominate stock selection while the risk manager influences
position sizing. Debate-based consensus enables agents to present arguments and
counterarguments, as illustrated by the bull--bear researchers in
TradingAgents \cite{xiao2024tradingagents}. FinCon instead refines investment
beliefs across episodes through CVRF \cite{yu2024fincon}. The choice of consensus mechanism
involves trade-offs between speed (voting is fast) and quality (debate can uncover
blind spots but takes longer). Dynamic mechanisms that adapt based on market
conditions or agent confidence are particularly valuable in volatile markets
\cite{hu2024consensus,wooldridge2009multiagent}~[BG].

\textbf{Task Allocation}. Effective role-based collaboration requires
task allocation that matches work to agent capabilities and current workload.
Capability-based allocation assigns tasks based on specialized expertise, such as
routing earnings analysis to the fundamental analyst while the technical analyst
handles technical signals. Workload-based allocation considers current load,
preventing bottlenecks when one agent is overwhelmed. Hybrid approaches combine
both factors, ensuring that urgent but non-specialized tasks (such as emergency
risk reduction) can be handled by any available agent. Task allocation may be
centralized (a supervisor agent assigns work, as in TradingAgents) or decentralized
(agents self-assign based on capability and availability). Dynamic reallocation
enables the system to adapt when agents fail or become overloaded, improving
robustness \cite{wooldridge2009multiagent}~[BG].

\textbf{Conflict Resolution}. Disagreements between specialized agents are inevitable,
particularly when agents have different perspectives (optimistic analyst vs pessimistic
risk manager) or when their roles create tensions (trader seeking profit vs risk manager
seeking safety). Negotiation allows agents to propose compromises, such as reducing
position size to satisfy both profit targets and risk limits. Arbitration delegates
dispute resolution to a higher authority (the supervisor in TradingAgents), ensuring
that critical decisions are not deadlocked. Authority hierarchies establish which
agent's view dominates in which domain (the risk manager has veto power over
position sizing, while the analyst controls stock selection). Learning from past
conflicts improves resolution over time, with agents developing heuristics for
when to escalate disagreements and when to accept compromises
\cite{wooldridge2009multiagent}~[BG].

\subsection{Hierarchical Organization}
\label{ssec:hierarch}

Hierarchical organization structures agents into layers, with higher-level
agents making strategic decisions and lower-level agents executing tactical
actions. This structure enables coherence across time scales, from long-term
asset allocation to short-term execution.

\textbf{Representative Approaches}. FinMem \cite{yu2023finmem} illustrates hierarchy within a single agent's
memory rather than a multi-agent command structure. It employs a
working memory for summarization, observation, and reflection, together with
long-term memory divided into shallow, intermediate, and deep processing layers.
These layers use different retention and decay schedules to support daily trading
decisions; they are not separate working, episodic, and semantic stores. H-MEM \cite{sun2025hmem}
extends this with semantic hierarchy, reporting efficiency improvements through
hierarchical memory organization. In multi-agent settings, strategic agents
set high-level targets (``reduce tech exposure by 20\%'') and tactical agents
determine specific trades to achieve these targets. General-purpose frameworks provide complementary organizational examples:
MetaGPT uses predefined roles and standard operating procedures for software
engineering \cite{xiao2023metagpt}~[BG], while DyLAN and HiveMind explore
adaptive team selection or prompt refinement
\cite{liu2023dynamic,xia2026hivemind}~[BG]. These mechanisms inform the design
of trading hierarchies without establishing trading performance themselves.

\textbf{Benchmarks and Protocols}. Coordination claims are difficult to compare
without common evaluation protocols and cost and execution semantics. Recent work
introduces live and replay-based benchmarks for trading agents that stress
sequential decision-making under real or event-driven market dynamics, including
AI-Trader \cite{fan2025aitrader} [BG], Agent Market Arena \cite{qian2025whenagentstrade}~[BG],
LiveTradeBench \cite{livetradebench_2025}~[BG], ContestTrade \cite{cao2025contesttrade}~[BG],
and PredictionMarketBench \cite{arora2026predictionmarketbench}~[BG].
These benchmarks help separate architecture effects (coordination design, memory,
workflow) from model-backbone effects, but remain sensitive to market selection,
episode construction, and transaction cost assumptions.
\begin{tcolorbox}[title=Protocol Implication]
	For multi-agent systems, we recommend explicit reporting of:
	(i) \textit{cross-play} evaluation (swap individual agents to test interoperability vs. team overfitting),
	(ii) explicit \textit{communication constraints} (channel bandwidth and tool-call budgets),
	(iii) \textit{crowding and impact} metrics (e.g., impact amplification under correlated policies), and
	(iv) \textit{role ablations} to attribute gains to coordination rather than increased information access.
\end{tcolorbox}

\textbf{Goal Decomposition}. Hierarchical organization requires breaking high-level
objectives into actionable sub-goals that lower-level agents can execute. Top-down
decomposition starts with strategic objectives (``target 15\% annual return and
10\% volatility'') and recursively breaks them into smaller targets (``allocate
60\% to equities, 30\% to bonds, 10\% to cash'' then ``select 50 technology stocks
with equal weights''). Constraint-based decomposition helps ensure that sub-goals
respect risk limits, liquidity constraints, and regulatory requirements. Dynamic
decomposition adjusts targets based on market conditions, such as reducing return
targets during bear markets or tightening risk limits when volatility spikes.
The effectiveness of goal decomposition depends on how well strategic intent can
be translated into tactical guidance without micromanaging lower-level agents; layers should specify clear I/O contracts (e.g., strategic targets vs. tactical orders) to prevent semantic drift \cite{wooldridge2009multiagent}~[BG].

\textbf{Feedback Mechanisms}. Hierarchies rely on upward feedback to close the loop
between strategy and execution. Bottom-up reporting enables tactical agents to
inform strategic agents about outcomes, constraints, and opportunities. For example,
execution agents report that liquidity is insufficient for the desired position size,
prompting strategic agents to revise the allocation. Exception escalation triggers
when lower-level agents encounter problems they cannot resolve, such as unexpected
market events or violations of risk limits. Performance feedback provides information
about which strategies are working, enabling strategic agents to reallocate resources
to successful approaches. Effective feedback requires appropriate aggregation: raw
data from execution must be summarized into metrics that inform strategic decisions
without overwhelming strategic agents with details \cite{wooldridge2009multiagent}~[BG].

\textbf{Exception Handling}. Lower-level agents inevitably encounter situations that
exceed their authority or capability, requiring escalation to higher-level decision-makers.
Predefined escalation triggers determine when to elevate issues, such as position
limits exceeded, unusual market conditions, or conflicting signals. Time-sensitive
exceptions (market crashes, circuit breakers) require immediate escalation with minimal
delays. Resolution strategies range from fully automated (circuit breakers automatically
reduce exposure) to human-in-the-loop (unusual patterns reported to human supervisors).
Learning from exceptions improves future handling, with systems developing better
triggers and resolution strategies over time. The design of exception handling involves
trade-offs between false positives (escalating too often, creating noise) and false
negatives (missing critical issues that require intervention) \cite{wooldridge2009multiagent}~[BG].

\textbf{Temporal Abstraction}. Different layers of the hierarchy operate at different
time scales, enabling separation between long-term planning and short-term reaction.
Strategic agents operate at daily, weekly, or monthly frequencies, making decisions
about asset allocation, factor tilts, and overall portfolio construction. Tactical
agents operate at intraday or minute-level frequencies, managing individual positions,
entry and exit timing, and execution quality. This separation prevents short-term
noise from disrupting long-term strategy: a temporary price spike should not trigger
a reallocation of the entire portfolio. However, temporal abstraction creates challenges
for coordination: strategic agents need to set bounds (risk limits, position limits) that
constrain tactical agents without being overly restrictive. Multi-scale learning enables
agents to learn at different time horizons, with strategic agents learning from
long-term patterns while tactical agents adapt to short-term dynamics
\cite{sutton1999options}~[BG].

\subsection{Market Ecology}
\label{ssec:ecology}

Market ecology models agents as participants in a market environment,
interacting through trading and prices rather than direct communication. In the
current literature, this pattern is best read as an exploratory setting for
studying collective behavior, not as mature evidence that ecological
coordination improves live trading outcomes.

\textbf{Representative Approaches}.
Market simulators \cite{hashimoto2023agent}
populate markets with heterogeneous agents to generate synthetic market data and
to stress test strategies under controlled conditions. Related work studies how
agent-environment feedback can produce boom-bust dynamics and how repeated
interaction may lead to undesirable emergent coordination, motivating regulatory
and monitoring considerations.
FinEvo \cite{zou2026finevo} proposes an ecological market-game framing to study the
evolution of interacting financial strategies, emphasizing that competitive performance
can depend on population composition and adaptive feedback rather than isolated backtests.
Validating ecological models requires rigorous calibration against real market microstructure to avoid \textit{simulator overfitting}, and protocols must further audit for \textit{timestamp leakage} and \textit{consensus smoothing} so that emergent coordination is not an artifact of look-ahead bias or noise suppression.

Pioneering work on generative agents \cite{park2023generativeagents}~[BG] established the viability of LLM-based simulacra, and subsequent financial-market simulations \cite{hashimoto2023agent} use such agents to probe how collective behavior shapes market dynamics in controlled settings.

\textbf{Market Mechanisms}. Market ecology relies on established market structures
to mediate agent interactions without requiring direct coordination. Continuous double
auctions enable buyers and sellers to trade at any time, with prices emerging from
the real-time balance of supply and demand. Call auctions aggregate orders over
specific time intervals (open, close, volatility interruptions), determining prices
through batch execution that reduces informational asymmetry. Dealer markets
designate specific market makers who provide liquidity by continuously quoting bid
and ask prices, with agents trading against these quotes rather than directly with
each other. Each mechanism creates different incentives and dynamics: continuous
double auctions favor fast reactions to information, call auctions reduce timing
risk, and dealer markets provide guaranteed liquidity but at a cost (the bid-ask
spread). The choice of market mechanism shapes the ecology by determining which
strategies are viable and how information propagates through prices
\cite{kyle1985continuous,glosten1985bid}~[BG].

\textbf{Price Formation}. In market ecology, no single agent sets prices; instead,
prices emerge from the aggregate behavior of all agents. Order aggregation combines
individual buy and sell orders into a collective order book, with the intersection
of supply and demand determining the market-clearing price. Supply-demand balance
shifts as agents react to news, portfolio changes, or risk management needs, causing
prices to adjust. Informational efficiency measures how quickly prices incorporate
new information, with efficient markets preventing agents from consistently profiting
from public information. However, when agents are heterogeneous (different beliefs,
time horizons, risk tolerances), prices may deviate from fundamental values for
extended periods, creating opportunities for some agents at the expense of others.
The dynamics of price formation depend on agent diversity: homogeneous agents tend
to produce rapid convergence to consensus prices, while heterogeneous agents create
more complex patterns with mispricings that persist until corrected by informed agents
\cite{kyle1985continuous,glosten1985bid}~[BG].

\textbf{Reflexivity and Feedback Loops}. Market ecology creates reflexive relationships
where agent actions affect prices, which in turn affect agent perceptions and actions.
This two-way feedback can produce complex dynamics not present in systems with
one-way causality. Positive feedback amplifies initial movements: price increases
attract trend-following agents who buy, pushing prices higher and attracting more
trend followers, potentially creating bubbles. Negative feedback stabilizes: price
increases trigger profit-taking by mean-reversion agents who sell, pushing prices
back toward fundamentals. The balance between positive and negative feedback determines
market stability. Reflexivity becomes particularly powerful when agents learn and
adapt: successful strategies attract imitators, changing market dynamics and potentially
rendering the original strategy unprofitable. This co-evolution of agents and markets
creates continuously changing patterns, making long-term prediction difficult even
as short-term patterns may be exploitable \cite{shiller1981,kirilenko2017flash}~[BG].

\textbf{Emergent Phenomena}. The interaction of heterogeneous agents through market
mechanisms can generate collective phenomena that are not apparent from studying
individual agents in isolation. Bubbles may occur when positive feedback dominates,
with rising prices attracting speculative buying disconnected from fundamentals.
Crashes may follow when bubbles burst or when negative feedback triggers panic selling,
with prices plummeting as agents rush to exit simultaneously. Flash
crashes---extreme short-term collapses in which prices plummet and recover within
minutes---are mechanistically plausible when high-frequency agents interact with market structure.
Tacit collusion is another plausible risk when agents learn to coordinate
behavior implicitly through repeated interaction without explicit communication,
raising regulatory concerns about market manipulation. Because these phenomena depend on the collective behavior of many agents and are difficult to predict,
agent-based simulation is more appropriately viewed as a hypothesis-generation tool for
understanding market dynamics and stress cases \cite{kirilenko2017flash,shiller1981}~[BG].

\paragraph{Strategy Crowding and Alpha Decay}
As LLM-based agents proliferate, similar training data and reasoning patterns
may lead to \textit{strategy homogenization}. When multiple agents:
\begin{itemize}
	\item Use similar foundation models (GPT-4, Claude)
	\item Fine-tune on similar financial corpora
	\item Employ comparable CoT or RAG patterns
\end{itemize}
they may generate correlated trading signals, creating mechanistically plausible risks such as:
\begin{enumerate}
	\item \textbf{Alpha decay}: Common signals are arbitraged away faster
	\item \textbf{Self-reinforcing trends}: Similar entry points amplify momentum
	\item \textbf{Correlated exits}: Simultaneous stop-loss triggering increases volatility
		\item \textbf{Flash crash risk}: Rapid unwinding of crowded positions
	\end{enumerate}

The evidence base currently lacks studies of multi-agent market impact at
scale---a critical gap that requires simulation-based research before
deployment.

\textbf{Regulatory Constraints}. Regulators shape market ecology by constraining
agent behavior to maintain market integrity and stability. Position limits prevent
any single agent from dominating the market or taking excessive risks. Circuit breakers
halt trading during extreme volatility, providing time for information to diffuse and
potentially preventing panic selling. Market maker obligations require designated
liquidity providers to quote prices continuously, ensuring that other agents can always
trade. Reporting rules may oblige agents to disclose large positions or trades,
reducing information asymmetry. Manipulation rules prohibit agents from engaging in
behavior that artificially affects prices (spoofing, layering, pump-and-dump schemes).
The design of these regulations must balance competing goals: too little regulation
enables manipulation and instability, while too much regulation reduces liquidity and
efficiency. Agent-based simulation enables regulators to test proposed rules before
implementation, identifying unintended consequences and optimizing regulatory parameters
\cite{mifid2,gdpr,wp29adm}.

\begin{figure*}[!tb]
	\centering
	\includegraphics[width=\linewidth]{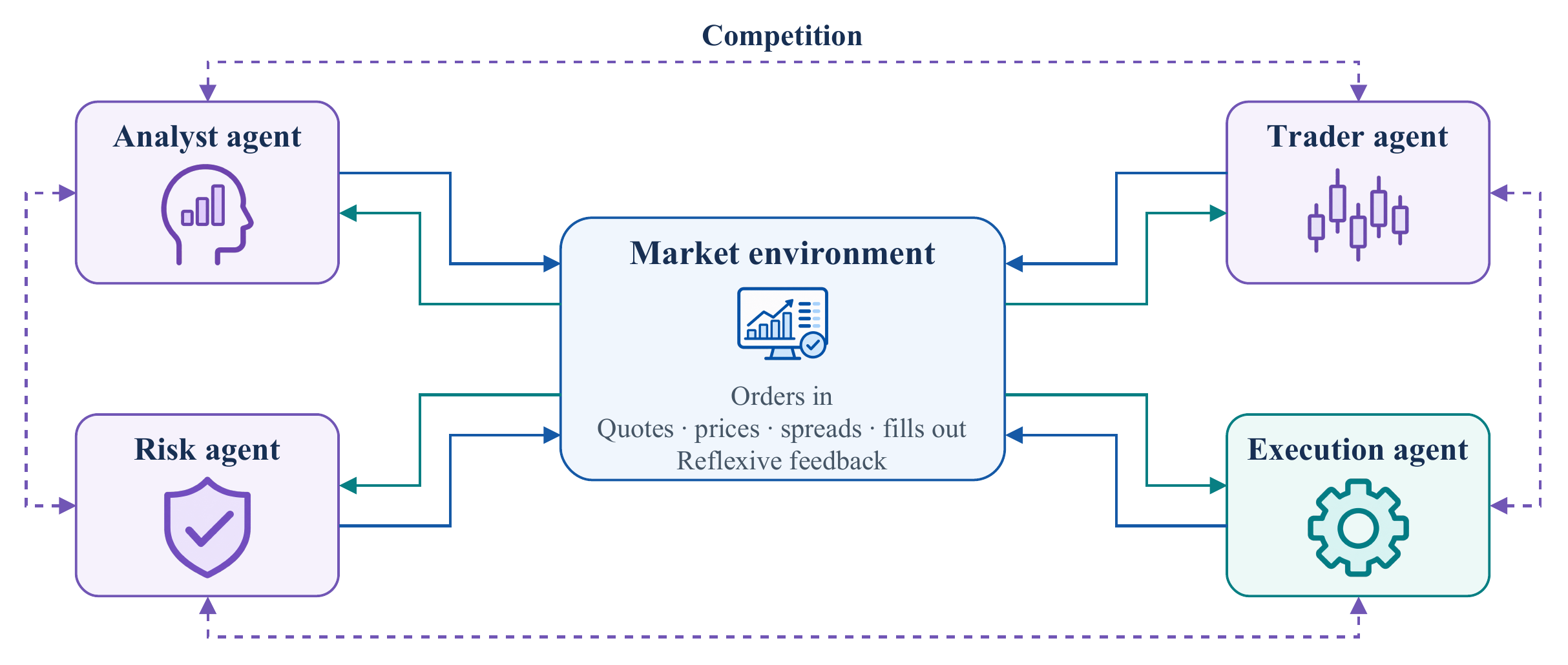}
	\caption{Market ecology interaction network. Agents interact both with one
		another and through the market environment, where prices, spreads, quotes,
		and fills feed back into subsequent behavior. The figure highlights
		reflexive feedback and competition channels that can support exploratory
		study of crowding, instability, and adaptation.}
	\label{fig:market-ecology}
\end{figure*}

\begin{table*}[!tb]
	\centering
	\caption{Multi-agent coordination patterns and design considerations. \textbf{Tags}: R=Reproducibility level (R0--R3), E=Execution semantics reported. Tags apply to the named trading systems in each row (see \Cref{sec:protocol} for full coding rules).}
	\label{tab:coordination-comparison}
	\begin{tabularx}{\linewidth}{@{}l p{2.6cm} p{2.0cm} p{2.5cm} X p{1.5cm}@{}}
		\toprule
		Pattern & Example & Comm. & Decision making & Coordination considerations & Tags \\
		\midrule
		Role-based & TradingAgents \cite{xiao2024tradingagents}, FINCON \cite{yu2024fincon}, TradingGroup \cite{tradinggroup_2025} & Direct, rich communication & Debate; belief updates & Communication and decision aggregation & \texttt{R0--R2,E} \\
		Hierarchical & Multi-agent investment pipeline \cite{li2026reits_multiagent} & Vertical, layered feedback & Top-down commands, bottom-up reporting & Layered reporting and decision aggregation & \texttt{R0,E} \\
		Market Ecology & Agent simulations; FinEvo \cite{zou2026finevo} & Indirect, via prices/trades & Decentralized, independent & Simulator calibration and market coupling & \texttt{R0,E} \\
		\bottomrule
	\end{tabularx}
\end{table*}

TradingAgents is coded R2, whereas FinCon and TradingGroup are R0; all three
report order-level execution semantics. The hierarchical investment pipeline
reports target weights, and FinEvo uses market-game execution. These tags
record reporting categories rather than equivalent execution realism.

\Cref{tab:coordination-comparison} compares their communication, authority, and design considerations.

\paragraph{Latency Constraints on Coordination}
The coordination mechanisms discussed above must operate within the latency budget established in \Cref{sec:action}. Coordination cost is dominated by communication, consensus formation, and decision aggregation, so voting is typically cheaper than debate while expert-weighted aggregation sits in between. In latency-sensitive regimes, systems may need to prefer the simplest mechanism that preserves acceptable decision quality. This latency-aware coordination design remains an open research direction for financial multi-agent systems.

The representative systems above illustrate several coordination mechanisms.
Extension-window examples include selective-consensus trading
\citep{li2026trusttrade}, co-evolving MARL agents \citep{jia2026evonash},
live deliberative analysis \citep{letteri2026agenticaita}, and a belief-to-trade
execution layer \citep{wang2026beyond}. Alpha-mining pipelines are discussed
in \Cref{sec:alpha}. Background market ecosystems examine self-organization
\citep{hashimoto2026financial,saxena2026machine}, specialized asset markets
\citep{casas2026conceptualization,passos2026artificial}, mechanism design
\citep{sheng2026formal,qian2026information,varghese2026information}, and
behavioral dynamics \citep{ouyang2026dissecting,park2026reflexivity}.
Language-grounded testbeds \citep{shi2026cstrader} and prediction-market arenas
\citep{barot2026polyswarm} provide further settings for studying interaction.
These background works inform coordination design and market behavior; the
ledger separates them from evaluated, action-producing trading systems.

\paragraph{Synthesis and Coordination Trade-offs}
Role-based collaboration, hierarchy, and market ecology differ in how agents
communicate and where coordination occurs. Communication cost, consensus
quality, and attribution shape these designs; role ablations and
compute-matched single-agent baselines help distinguish coordination effects.
The final section examines persistent adaptation and structural self-evolution
in \Cref{sec:evolution}.


\section{Adaptation and Self-Evolution}
\label{sec:evolution}

\begin{figure*}[!tb]
	\centering
	\includegraphics[width=\linewidth]{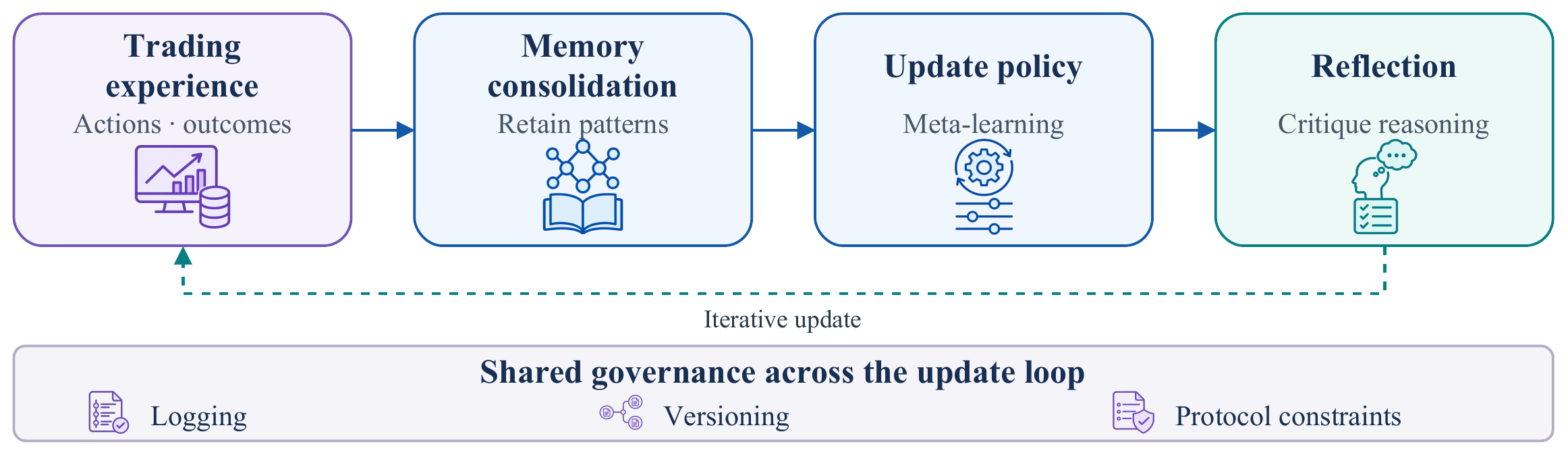}
	\caption{Conceptual governed update loop. Agents accumulate experience through
		trading, consolidate important patterns into memory, apply meta-learning-like
		update policies, and use reflection to critique reasoning under logging and
		rollback constraints.}
	\label{fig:evolution-cycle}
\end{figure*}

Adaptation is the umbrella term for updates to an agent's state or behavior.
Here, self-evolution denotes agent-managed changes retained across decision
episodes; structural or architectural self-evolution is the narrower case in
which strategy code, update procedures, modules, interfaces, workflows, or
system topology change. Memory consolidation changes what experience is
retained, while meta-learning changes how learning is performed; either can
support persistent adaptation without changing the architecture. Reflection
within a single decision episode does not by itself qualify as self-evolution.
Building on the stores and retrieval operations in Memory Architecture and the
ICL, SFT, and RL update modes in Learning Paradigms, this section examines
memory consolidation, meta-learning, and self-reflection in that order:
\textquotedblleft How are agent updates carried across decision episodes and evaluated?\textquotedblright
The conceptual update loop is shown in \Cref{fig:evolution-cycle}.

\textit{Note}: Many mechanisms described in this section (e.g., memory
consolidation, meta-learning, self-reflection) originate in general AI
research; in trading, they remain experimental and should be interpreted
through explicit \textit{update governance} rather than as industry-standard
practice. Studies marked [BG] are background literature without end-to-end
trading evaluation. Adaptation\slash self-evolution claims are only comparable
when the update loop is reproducible (MR-6; see \Cref{sec:challenges}): papers
should report what can change (prompts, memory summaries, retrieval indices,
weights), when updates trigger, and what historical window is used. They should
preserve immutable logs, frozen execution and cost semantics, and a rollback
path, and evaluate the post-update state on sealed data.

\subsection{Memory Consolidation}
\label{ssec:consolidation}

Building on the stores and retrieval operations described in
\Cref{sec:memory}, memory consolidation changes what those stores retain over
time: it compresses experience into reusable patterns while discarding details
that no longer aid retrieval.

\textbf{Representative Approaches}. Within the section-level primary subset,
memory-consolidation-like behavior appears in systems that retain episodic traces
and derived summaries for later retrieval, including FinMem and related
retrieval-centered trading agents \cite{yu2023finmem,wu2024finagent}. Background
work considers replay, summarization, abstraction, and forgetting as ways to
select or compress those traces. Broader systems such as TradingGPT remain in the
wider citation set and do not enter the section-level primary-study count
\cite{tradinggpt2023}.

\textbf{Technical Mechanisms}. Background methods score memories by recency,
outcome, or retrieval frequency and use replay, compression, or abstraction to
retain useful signals \cite{lin1992selfimproving,schaul2016prioritized,mcclelland1995complementary,hinton2006reducing}~[BG].
When market conditions change, temporal or contextual rules can resolve
conflicting memories \cite{mcclelland1995complementary,kirkpatrick2017overcoming}~[BG].
These operations may run periodically, respond to events such as regime shifts,
or combine both schedules \cite{kirkpatrick2017overcoming}~[BG]; their value
depends on whether the resulting state remains temporally valid and auditable.

\textit{Audit Trail Requirement}: Summarization can hallucinate details not present in the original log or lose the link to the trade decision. To remain auditable, consolidated memories must retain an immutable pointer (hash or timestamp) to the raw transaction logs. Adaptation protocols should periodically validate that summaries do not contain factual errors or future information compared to the ground-truth logs.

\subsection{Meta-Learning}
\label{ssec:metallearn}

Meta-learning enables agents to learn how to learn, improving the efficiency and
effectiveness of the learning process itself. Finance-oriented prototype systems considered here primarily explore multi-agent
evolutionary loops to generate and select strategies or factors under backtest
feedback \cite{quantevolve_2025,efs_2025,zou2026finevo}. The extension window adds a
distinct self-evolving cluster---experience-driven policy refinement
\cite{kim2026evolvetrade}, rubric-policy evolution \cite{chen2026sharp},
agent-to-agent alpha co-evolution \cite{li2026ai}, meta-evolution of trading
programs \cite{sharma2026algoevolve,kvasiuk2026madevolve}, and post-trained local
research agents \cite{wang2026alphadiverse}---alongside regime-adaptive continual
learning for portfolio management \cite{pan2026regime}. These results remain
prototype-level and must be interpreted under each study's evaluation and
cost/execution assumptions. Evolutionary simulation also provides an existence
proof for trend-following behavior \cite{chen2026be}, situating adaptation in a
market-ecology context. Across these lines, the central trading challenge is to
define tasks without temporal leakage and seal feedback against future outcomes.

\textbf{Task Splitting Requirement}. In trading, a ``task'' cannot simply be a random sample of assets or days. To prevent leakage, tasks must be split by time (e.g., 2010--2015 as source, 2016 as target) or by distinct, non-overlapping asset classes. Random shuffling of episodes across time violates the causal structure of markets, creating false ``adaptation'' that is merely memorization of future regimes.

\textbf{Representative Approaches}. Background MAML learns an initialization
that supports rapid adaptation, while learning-to-learn methods can update the
learning-rate schedule, exploration strategy, or architecture
\cite{finn2017maml}~[BG]. Finance-oriented prototypes apply iterative critique
to factor or policy search: AlphaAgent regularizes factor generation, and
TradingGroup combines reflection with data synthesis
\cite{alphaagent_2025,tradinggroup_2025}. General multi-agent optimization
provides a related refinement pattern, but remains background rather than
finance-specific evidence \cite{yuksel2024multi}~[BG].

\textbf{Technical Mechanisms}. Meta-learning evaluates adaptation across a task
distribution, using support data for within-task updates and query data to
measure post-update performance. MAML formalizes this as inner-loop adaptation
and outer-loop optimization; few-shot and domain-generalization methods extend
the same idea to low-data or unseen related tasks
\cite{finn2017maml,vinyals2016matching,snell2017prototypical,muandet2013domain}~[BG].
Memory or replay can support these updates, but concerns how learning uses stored
experience rather than how that experience is represented
\cite{finn2017maml}~[BG]. Transfer learning applies learned representations or
behavioral strategies to new tasks, while domain adaptation fine-tunes them with
target-specific data \cite{pan2010survey}~[BG].

\subsection{Self-Reflection}
\label{ssec:reflection}

Self-reflection enables agents to critique their own reasoning and to catch
mistakes before they propagate into actions, thereby improving decision
quality.

\textbf{Representative Approaches}.\par
Reflexion \cite{shinn2023reflexion}~[BG] adds an explicit
critique--revise loop: after proposing a decision, the agent diagnoses weaknesses in
its own reasoning and regenerates an improved plan. In trading settings, such
reflection can check for common failure modes
(feature/label leakage, stale data,
ignored costs, regime mismatch). Self-consistency
and retrospective error analysis complement Reflexion by exposing unstable rationales
and recurring mistake patterns.
In a finance-specific setting, CryptoTrade uses a reflective mechanism to refine
zero-shot cryptocurrency trading decisions by analyzing outcomes of prior trades
\cite{cryptotrade_zero_shot_2024}. Within the section-level primary subset, the
main trading-facing anchor is TradingGroup, which combines role specialization
with self-reflection and data synthesis \cite{tradinggroup_2025}; other
examples illustrate critique and refinement mechanisms.

\begin{table*}[!tb]
	\centering
	\small
	\setlength{\tabcolsep}{3pt}
	\caption{Conceptual update mechanisms and uses of experience. All entries are background mechanisms (\texttt{BG}); the descriptions identify update operations and the role of retained experience.}
	\label{tab:evolution-comparison}
	\begin{tabularx}{\linewidth}{@{}l p{2.3cm} p{3.3cm} X X p{1.4cm}@{}}
		\toprule
		Mechanism       & Description                 & Key Mechanism                                         & Update work       & Experience use              & Tags        \\
		\midrule
		Consolidation   & Summarize, abstract, forget & Importance scoring, compression & Memory summarization, abstraction, and forgetting & Selects retained traces for later retrieval & \texttt{BG} \\
		Meta-Learning   & Learn to learn              & MAML \cite{finn2017maml}, memory-based opt & Inner-loop adaptation and outer-loop optimization & Support/query data across tasks & \texttt{BG} \\
		Self-Reflection & Critique and refine         & Reflexion \cite{shinn2023reflexion}, self-consistency & Critique and revision of a proposed plan & Critique informs a regenerated plan & \texttt{BG} \\
		\bottomrule
	\end{tabularx}
\end{table*}

\begin{tcolorbox}[title={Technical Mechanisms: Protocol Implication},
  title after break={Protocol Implication (continued)}]
	Effective self-reflection is a \,\emph{protocol}, not a free-form ``thinking harder'' step; we suggest five stages. (i)
	\textit{Critique generation}: produce actionable
	checks tied to the trade proposal (data timestamps, assumptions, transaction costs,
	liquidity/slippage, and risk constraints). (ii) \textit{Refinement}: run a small, bounded
	number of critique--revise iterations with early stopping; log the critique and the
	resulting change for auditability. (iii) \textit{Consistency/uncertainty}: sample multiple
	reasoning paths; treat disagreement as a deferral/abstain signal to reduce overconfident
	errors, but validate thresholds empirically (consensus is not correctness). (iv)
	\textit{Post-trade review}: categorize failures (prediction vs execution vs risk) and
	store compact lessons in memory keyed by instrument/regime so they can be retrieved at
	decision time. (v) \textit{Pre-trade verification}: enforce hard constraints (risk limits,
	compliance rules, and operational feasibility) as deterministic checks.
\end{tcolorbox}

\textit{Risk of Search Bias}: Unbounded reflection acts as a search process. To prevent ``testing into success'' (retrying until a profitable trade appears by chance), protocols must enforce a \textbf{Reflection Budget} (max iterations) and refrain from using Out-of-Sample feedback to terminate the loop. Reflective traces are not a substitute for testing, monitoring, or out-of-sample evaluation \cite{wang2022selfconsistency,madaan2023selfrefine,geifman2017selective}~[BG].

\paragraph{Synthesis and Governance Dimensions}
The subsections and \Cref{tab:evolution-comparison} distinguish updates to
retained experience, learning procedures, and reasoning outputs; persistent
changes to strategy code, update procedures, or system organization are
structural self-evolution.
Whether any update carries across later trading decisions is an empirical
question under the governance conditions above. These distinctions frame the
cross-cutting challenges and future directions in \Cref{sec:challenges}.



\section{Technical Challenges, Methodological Guidelines, and Open Problems}
\label{sec:challenges}

The preceding sections have systematically synthesized the architectural foundations,
functional capabilities, and adaptation pathways of agentic trading systems. This section
examines the principal technical and practical bottlenecks that arise when deploying
LLM agents in financial markets, consolidates methodological best-practice guidelines
(MR-1--MR-7) to foster reliable empirical evaluation, and outlines key open research problems.

As observed across the empirical literature ($n=\EvidencePrimaryN$), researchers adopt
varying conventions regarding data splits, transaction frictions, asset universe definition,
and execution timing. To promote cross-study comparability, mitigate data snooping, and
facilitate real-world translation, we distill these recurring empirical considerations into
a structured set of methodological recommendations:

\begin{tcolorbox}[
	colback=tradingagent_light!20,
	colframe=tradingagent_blue!65,
	colbacktitle=tradingagent_blue!15,
	coltitle=black,
	title={\bfseries Methodological Guidelines and Recommended Evaluation Practices (MR-1--MR-7)},
	title after break={\bfseries Methodological Guidelines (continued)},
	boxrule=0.6pt,
	arc=2pt,
	parbox=false,
	before skip=6pt,
	after skip=10pt,
	left=6pt,
	right=6pt
]
	To support rigorous benchmarking and reproducibility, future research on agentic trading is encouraged to document the following empirical dimensions:
	\begin{itemize}
		\evidenceitem{\textbf{MR-1 (Data \& Universe)}: Specify asset universe construction criteria, survivorship bias treatment, and synchronized data availability timestamps across all perceived modalities.}
		\evidenceitem{\textbf{MR-2 (Temporal Splitting)}: Enforce strict chronologically ordered train/validation/test partitions, walk-forward or rolling windows, and explicit embargo periods to prevent look-ahead bias; for iterative alpha search, maintain a frozen holdout test set.}
		\evidenceitem{\textbf{MR-3 (Action Semantics)}: Explicitly define output format (discrete orders vs. continuous portfolio target weights), decision timestamps, order mapping rules, and portfolio constraints (leverage, gross/net exposure).}
		\evidenceitem{\textbf{MR-4 (Execution \& Costs)}: Explicitly state execution timing, fill price assumptions (e.g., next-open or volume-weighted prices), rebalancing intervals, and realistic friction models (brokerage fees, bid-ask spread, market impact).}
		\evidenceitem{\textbf{MR-5 (Leakage Controls)}: Verify causal temporal integrity across feature normalization, text sentiment alignment, and target label construction.}
		\evidenceitem{\textbf{MR-6 (Artifacts \& Verification)}: Provide open-source code, reproducible configuration scripts, random seeds, and structured decision logs; for adaptive systems, record update frequencies and rollback criteria.}
		\evidenceitem{\textbf{MR-7 (Multi-Agent Evaluation)}: In collaborative or competitive systems, document agent role topologies, communication protocols, and ablation experiments isolating the marginal contribution of inter-agent interactions.}
	\end{itemize}
\end{tcolorbox}

\begin{table*}[t]
	\caption{Applicability of the MR-1--MR-7 evaluation guidelines by study type. \textbf{E}: Essential to interpreting the relevant result; \textbf{R}: Recommended for verification; \textbf{O}: Task-dependent; \textbf{--}: Generally outside the stated task. These designations summarize this survey's methodological recommendations.}
	\label{tab:mr-matrix}
	\small
	\centering
	\begin{tabular}{l c c c}
		\toprule
		\textbf{MR Item} & \textbf{End-to-End} & \textbf{Execution} & \textbf{Component} \\
		& \textit{(Full Trading)} & \textit{(Ord/Wt Algo)} & \textit{(Perception/Risk)} \\
		\midrule
		MR-1 (Data \& Universe) & \textbf{E} & O & \textbf{E} \\
		MR-2 (Time Split \& Leakage) & \textbf{E} & \textbf{E} & \textbf{E} \\
		MR-3 (Action Semantics) & \textbf{E} & \textbf{E} & -- \\
		MR-4 (Exec \& Costs) & \textbf{E} & \textbf{E} & -- \\
		MR-5 (Leakage Audit) & \textbf{E} & \textbf{R} & \textbf{R} \\
		MR-6 (Artifacts) & \textbf{R} & \textbf{R} & \textbf{R} \\
		MR-7 (Multi-Agent Eval) & O & -- & -- \\
		\bottomrule
	\end{tabular}
	\begin{flushleft}
		\footnotesize{\textbf{Reading the matrix:} Applicability follows the study's task and reported result. Action semantics and cost assumptions are essential to interpreting trading or execution outcomes, while a standalone perception study may be evaluated without generating orders. Universe selection can be exogenous to an execution algorithm. Artifacts and extraction checks support verification, and multi-agent evaluation applies when interaction is part of the design. The matrix guides interpretation; corpus-level reporting counts follow the separate coding rules in \Cref{sec:protocol}.}
	\end{flushleft}
\end{table*}

\Cref{tab:mr-matrix} relates the guidelines to each study's task. For
\textbf{end-to-end trading}, universe construction and cost models help interpret
profitability; an \textbf{execution algorithm} receiving parent orders may take
the universe as given. A standalone \textbf{component study}, such as sentiment
analysis, needs time-consistent evaluation but may operate without a transaction
cost model. The matrix distinguishes relevant evaluation dimensions from fields
outside the study's scope.


\subsection{Technical Challenges}

The subsections below synthesize technical challenges, methodological guidance,
and representative research directions.

\subsubsection{Hallucination in Finance}

LLM hallucination, which generates plausible but false information, poses
\textit{material} risk in finance because trading decisions may be executed,
and losses realized, before the underlying claim can be verified. Only one
primary study explicitly foregrounds this issue, so the discussion below
combines that direct evidence with cross-section risk synthesis.

\textbf{Propagation Mechanisms.}
In agentic workflows, hallucinated claims propagate through tool calls into
downstream actions. A fabricated earnings report can trigger:
\begin{enumerate}
	\item Erroneous position entry (immediate loss when corrected)
	\item Stop-loss cascade (if other agents detect the ``signal'')
	\item Confidence-based position scaling (larger losses due to false certainty)
\end{enumerate}

\textbf{Mitigation Limitations.}
Binding outputs to verifiable sources (FinMCP, HydraRAG) shifts rather than
eliminates risk, because source selection itself can be erroneous.
Cross-validation helps only where multiple independent sources exist, which is
often not the case for time-sensitive trading signals; moreover, current
fact-checking mechanisms operate at second-scale latency, which is incompatible
with millisecond trading decisions. We therefore cite these systems as
illustrative tooling, not as empirically validated mitigations.

\textbf{Regulatory Implications.}
Firms deploying LLM agents may face heightened liability if hallucinated outputs
cause client losses; documentation of source binding, validation checks, and
human review may be needed to withstand regulatory scrutiny.

\subsubsection{Look-ahead Bias}

We treat look-ahead bias here as methodological guidance tied directly to
MR-2 and MR-5. Recent work includes dedicated analyses of temporal leakage in LLM backtesting
\cite{zhang2026temporal}, look-ahead-freedom as a verifiable correctness
property \cite{fonseca2026look}, deployment-evidence limits of reported alpha
\cite{ye2026alpha}, and execution-assumption audits
\cite{yao2026beyond}, alongside leakage-masking benchmarks
\cite{zhu2026knowing,qu2026clqt,cai2026openpm}. Methodological work goes further
by engineering the problem away: anonymization-first evaluation asks whether
``blindfolded'' LLMs can still trade when asset identities and dates are
withheld \cite{jeon2026can}, directly operationalizing look-ahead-freedom as a
test condition.

Look-ahead bias occurs when models inadvertently access future information during
training or evaluation, producing unrealistic performance estimates that fail to
generalize to real trading conditions. The risk is particularly insidious in
time-series settings, where leakage can enter through improper cross-validation
that shuffles temporal data, technical indicators computed with future
information (e.g., centered moving averages), or textual sources that contain
hindsight about market movements. Detection requires careful examination of
model inputs and training procedures, including causal inference tests that
check whether model performance degrades when strict temporal constraints are
enforced: a model that appears strong under naive random splits can degrade
sharply under proper time-consistent evaluation, revealing substantial
look-ahead contamination \cite{lopezprado2018dangers}. Prevention requires
strict temporal splitting (train on past, test on future), walk-forward
validation that simulates real-time deployment, and causal modeling so that
information flows only from past to future. Without these controls, even the
most sophisticated learning and adaptation procedures fail under evaluation
leakage.

\subsubsection{Latency vs Accuracy Tradeoff}

Latency is a design constraint that links reasoning depth, model inference,
communication, and execution timing.
System 2 reasoning (reflective, multi-step) yields better decisions but is too
slow for high-frequency trading, where decisions must be made in microseconds,
while System 1 reasoning (reactive, fast) lacks the depth for complex decisions
involving multi-hop reasoning or extensive analysis. In latency-sensitive
regimes, the throughput advantage of faster decision policies can therefore
outweigh modest accuracy gains from slower deliberation, especially when
opportunities are fleeting and execution queues are competitive. Hybrid
approaches address this tension with cascaded designs: fast System 1 handles
routine decisions using \Cref{sec:perception} perceptual features and escalates
difficult cases to slower System 2 reasoning that leverages \Cref{sec:reasoning}
deliberative mechanisms. Model distillation offers a complementary route,
compressing large reasoning models into smaller, faster models that retain much
of their accuracy. Future work should develop tradeoff mechanisms that allocate
computational budget dynamically according to decision importance, market
volatility, and time pressure, potentially learned via reinforcement learning.

\subsubsection{Explainability vs Performance}

The explainability--performance tension is a cross-section governance
concern. Representative approaches include debate-based explainable
trading \cite{huang2026multi}, post-hoc explanation of learned portfolio
policies \cite{escudero2024explainabl}, and audits of asset-specific preferences
embedded in financial LLMs \cite{wu2026auditing}. The discussion below
accordingly provides background and author guidance linked to deployment
governance.

More complex models (deep neural networks, ensemble methods, large language
models) often achieve better performance but are less interpretable, creating a
fundamental tension be\-tween explainability and predictive power. This tension
has practical as well as academic weight: depending on jurisdiction and system
role, regulatory and legal frameworks (e.g., the EU's MiFID II) may impose
governance, record-keeping, and explainability expectations for algorithmic
trading and decision support \cite{mifid2}, and GDPR provisions with subsequent
guidance can imply obligations to provide ``meaningful information'' about
automated processing in certain contexts, though scope and applicability vary
\cite{gdpr,wp29adm}. When an agent's alpha generation module (\Cref{sec:alpha})
triggers a large position, risk managers and regulators need to know
why---whether the driver was a particular news event, a technical-indicator
pattern, or a reasoning chain. Neuro-symbolic approaches pair neural components
for pattern recognition with symbolic components for transparent reasoning,
aiming to preserve both performance and interpretability. Attention
visualization highlights which inputs most influenced the model (e.g., specific
news articles or price movements), while counterfactual explanations specify
what would have had to change to alter a decision (e.g., ``the trade would not
have occurred if the FOMC minutes had been less hawkish''). Regulatory and
governance expectations around explainability are likely to motivate further
advances in this area, particularly for the risk-management and compliance
applications of \Cref{sec:risk}.

Many academic systems demonstrate explanations over offline datasets or
controlled prototypes; production deployments typically require additional
engineering for auditability (decision logs, data lineage, and controlled
model/prompt changes) to make explanations actionable for governance.

\subsubsection{Computational Cost and Economic Feasibility}

The economic viability of LLM-based trading agents depends on the relationship
be\-tween inference costs and generated returns. A simple annualized estimate is
\[
	\text{Annual API Cost} = D \times N \times C_{req},
\]
where $D$ denotes trading days per year, $N$ the number of decisions per day, and $C_{req}$ the effective per-request cost implied by model pricing, prompt/completion token budgets, and any tool-specific charges. Because vendor pricing and model offerings change rapidly, empirical cost estimates should specify the model version, pricing snapshot date, token assumptions, tool usage, and the break-even gross P\&L required for net profitability after inference and execution costs. Recent work directly examines whether agentic trading systems can cover their inference costs once those costs are charged against returns \cite{duan2026can}.

\subsubsection{Evaluation Protocols and Deployment Safeguards}

This subsection translates the survey-wide minimum reporting framework
(MR-1--MR-7) into concrete deployment safeguards, drawing on the recurring
reporting gaps identified earlier. Evidence-backed guidelines and common anti-patterns for
agentic trading research and deployment include:
\begin{itemize}
	\item \textbf{Guard against temporal leakage}: use strictly time-ordered
	      splits and walk-forward evaluation, and ensure feature engineering and
	      scaling are fit only on past data \cite{lstm2025leakage}.
	\item \textbf{Avoid unrealistic cost models}: include fees, bid--ask spread,
	      slippage, and market impact, and report sensitivity to cost assumptions
	      \cite{perold1988shortfall,olby2025right}.
	\item \textbf{Check universe construction}: document inclusion and exclusion
	      rules and timestamp availability, and audit for survivorship/selection bias
	      when building backtest datasets \cite{brown1992survivorship}.
	\item \textbf{Mitigate backtest overfitting}: separate model selection from
	      final evaluation, use robust out-of-sample procedures, and be explicit about
	      multiple-testing risk \cite{backtest_overfitting_2024}.
	\item \textbf{Add deployment safeguards}: maintain immutable action logs that
	      link each action to its inputs, evidence, and model versions, and use
	      verifiable retrieval, citation checks, and cross-source consistency
	      tests \cite{khatchadourian2026replayable,tan2025hydrarag}.
\end{itemize}

The extension cohort supplies early empirical anchors for this guidance:
backtest-to-live gaps measured directly in crypto markets \cite{yu2026can}, a
live hybrid trading agent evaluated under a shared task protocol
\cite{rashid2026fin}, prospective process-aware evaluation of adaptive agents
\cite{luo2026foresight}, and a unified platform for tracking agentic trading
performance over time \cite{li2026nextfund}.


\subsection{Human-in-the-Loop Oversight}

Although this survey focuses on agentic systems, many realistic deployments retain human oversight at escalation points such as large notional changes, regime breaks, or tool failures. We therefore view human-in-the-loop design as a practical control boundary: stronger autonomy should imply stronger logging, clearer intervention rules, and explicit specification of when human approval is mandatory.


\subsection{Negative Results and Failed Designs}

The literature is likely affected by survivorship bias at the survey level as well as at the backtest level. Few papers report failed prompts, abandoned reward schemes, degraded live performance, or architectures that increased turnover without improving net returns. We therefore interpret the absence of negative results as a reporting limitation rather than as evidence that the design space is uniformly promising.


\subsection{Future Directions}

\subsubsection{Neuro-Symbolic Integration}

Current LLM-based agents rely almost entirely on neural computation, which is
flexible but opaque and unreliable where exact calculation or logical
guarantees are required. As discussed in \Cref{sec:reasoning}, purely neural
approaches struggle with tasks such as portfolio optimization under constraints
or exact arithmetic. Neuro-symbolic integration addresses this by combining
neural networks (for pattern recognition and learning from data) with symbolic
reasoning (for logic, constraint satisfaction, and exact computation): hybrid
architectures can use neural networks to extract structured information from
text (prices, dates, entities) and symbolic reasoning to perform financial
calculations, enforce budget constraints, and maintain logical consistency.
Specific approaches include (1) \textbf{neural theorem provers} that use neural
networks to guide proof search for financial constraints; (2)
\textbf{differentiable logic} that integrates logical constraints into neural
network loss functions; (3) \textbf{program synthesis}, in which LLM-generated
code is formally verified; and (4) \textbf{neuro-symbolic ILP} (Inductive Logic
Programming), which learns logical rules from neural representations. The
research objective is to combine neural flexibility with symbolic precision
while keeping the resulting systems auditable enough for financial use.

\subsubsection{Longer-Term Directions: Coordination, Governance, and Infrastructure}

Three longer-term directions extend these expert-system design considerations.
First, larger multi-agent market simulations could serve as controlled
``wind tunnels'' for studying crowding, flash episodes, and systemic-risk
propagation. The current evidence base is concentrated in small role-based
teams, motivating evaluation of larger agent populations. Second, increasing
agent autonomy motivates governance research on auditable records, intervention
rules, and sandboxed evaluation that clarify who approved, monitored, or
overrode each action. Third, low-latency infrastructure and model compression
support deployment; evaluating their benefits requires end-to-end decision
latency, execution timing, and cost/benefit analysis.


\Cref{tab:claim-evidence-map} connects the survey's design themes with
representative studies and the evaluation dimensions needed to interpret them.

\begin{table*}[!t]
	\centering
	\small
	\setlength{\tabcolsep}{4pt}
	\caption{Synthesis of recurring design themes and their evaluation contexts. Representative studies illustrate the mechanisms discussed in the technical chapters; the final column identifies dimensions for interpreting their implementations and reported results.}
	\label{tab:claim-evidence-map}
	\begin{tabularx}{\linewidth}{@{}p{3.1cm}X X@{}}
		\toprule
		\textbf{Design theme} & \textbf{Representative studies} & \textbf{Interpretive dimensions} \\
		\midrule
		Closed-loop pipelines and evaluation settings & FinAgent~\cite{wu2024finagent}, TradingAgents~\cite{xiao2024tradingagents}, FinCon~\cite{yu2024fincon}, FinRL-DeepSeek~\cite{benhenda2025finrl}, Agent Market Arena~\cite{qian2025whenagentstrade}, FinMem~\cite{yu2023finmem}; evaluation examples include OpenPM~\cite{cai2026openpm}, CLQT~\cite{qu2026clqt}, and LiveOption~\cite{luo2026liveoption} & Connect observations, retained state, decisions, and market feedback; distinguish backtest, simulation, live, and benchmark settings. \\
		Action and execution interfaces & FinAgent~\cite{wu2024finagent}, TradingAgents~\cite{xiao2024tradingagents}, AI-Trader~\cite{fan2025aitrader}, FinRL-DeepSeek~\cite{benhenda2025finrl} & Distinguish target weights from orders, and interpret outcomes using decision timing, fill rules, and cost assumptions. \\
		Memory and adaptation mechanisms & FinAgent~\cite{wu2024finagent}, FinMem~\cite{yu2023finmem}, TradingGroup~\cite{tradinggroup_2025}, Alpha\textsuperscript{2}~\cite{alpha2_2024} & Separate storage and retrieval from prompt, parameter, and strategy updates; identify which changes persist across decisions. \\
		Multi-agent organization & TradingAgents~\cite{xiao2024tradingagents}, TradingGroup~\cite{tradinggroup_2025}, FinCon~\cite{yu2024fincon}, REIT multi-agent pipeline~\cite{li2026reits_multiagent} & Specify roles, communication, and decision authority; role ablations and matched resource budgets help assess coordination effects. \\
		Evaluation practice and reproducibility & Primary-subset protocol tables in Section~\ref{sec:protocol} and R0--R3 artifact descriptions & Interpret field coverage within the assembled cohorts, accounting for retrieval and coding differences; distinguish artifact availability from trading performance. \\
		\bottomrule
	\end{tabularx}
\end{table*}

Across the empirical literature, rigorous evaluation methods---such as sealed out-of-sample testing, realistic market impact modeling, and latency-aware execution---are increasingly recognized as foundational pillars for advancing the field. Combining architectural innovation with disciplined empirical validation will be paramount as autonomous agent systems transition from experimental prototypes toward resilient, production-grade market deployment.


\section{Conclusion}
\label{sec:conclusion}

\subsection{Synthesis of the Agentic Trading Landscape}

The integration of Large Language Models into quantitative finance represents a
fundamental paradigm shift from static, black-box forecasting models toward
autonomous, cognitive trading agents. By unifying multimodal perception, structured
memory retrieval, multi-step cognitive reasoning, and friction-aware order execution,
agentic systems introduce an adaptive decision loop capable of operating in
complex, non-stationary market environments.

This survey has provided a comprehensive, structured synthesis of this rapidly
expanding domain. Through the \textbf{Architecture--Capability--Adaptation (A-C-A)}
conceptual framework, we have systematically decoupled:
\begin{itemize}
	\item \textbf{Architecture (Part I)}: The foundational cognitive and operational
	      building blocks---perception across text, time-series, and charts; working
	      memory, associative vector storage, and structured knowledge bases; reactive, reflective, and strategic
	      reasoning paradigms; and order-routing action interfaces.
	\item \textbf{Capability (Part II)}: The primary quantitative functions realized by
	      autonomous agents---symbolic and code-based alpha discovery, multi-asset
	      portfolio optimization, and multi-tier pre-trade, in-trade, and post-trade
	      risk governance.
	\item \textbf{Adaptation (Part III)}: Learning and cross-decision update
	      mechanisms---in-context learning, persistent memory and reflection that
	      can update decisions without changing the architecture, supervised
	      fine-tuning, reinforcement learning, multi-agent coordination and
	      adaptive interactions, and structural self-evolution.
\end{itemize}

Drawing on \textbf{\EvidenceIncludedN}~systematically mapped studies from an
initial multi-source ledger and an arXiv extension through 2026-09-30, our
synthesis connects the core cohort of $n=\EvidencePrimaryN$~closed-loop trading
systems with background work on financial models, tools, and benchmarks.

\subsection{Market Microstructure and Practical Implications}

The proliferation of autonomous trading agents carries profound implications for
both market dynamics and quantitative practice:
\begin{itemize}
	\item \textbf{Market Dynamics and Ecology}: As autonomous agents interact at scale,
	      their collective behaviors will increasingly shape market liquidity, price
	      discovery, and volatility. Investigating emergent phenomena---such as strategic
	      crowding, herding around shared LLM foundations, and algorithmic feedback
	      loops---presents a fertile ground for synthetic market simulation.
	\item \textbf{Quantitative Engineering}: In practical asset management, agentic
	      architectures provide a principled mechanism to bridge qualitative domain
	      insights (earnings calls, regulatory filings) with rigorous quantitative
	      optimizers, transforming quantitative workflow from manual factor engineering
	      to autonomous research pipelines.
\end{itemize}

\subsection{Empirical Evaluation and Methodological Practices}

Empirical evaluation is an important dimension of the architectural landscape.
Across the pooled 80-study empirical subset, reported protocol-field coverage includes
time-consistent split discipline (\ProtocolSplitReportedPct\%), explicit
transaction-cost modeling (\ProtocolCostReportedPct\%), universe survivorship
controls (\ProtocolUniverseReportedPct\%), and execution semantics
(\ProtocolExecutionReportedPct\%). The cohort breakdown in
\Cref{tab:cohort-protocol-coverage} gives the corresponding counts under the
different retrieval and coding procedures. These pooled values summarize the
assembled literature; they do not establish a temporal trend in reporting practices.

The methodological guidelines (MR-1--MR-7) organize practices that support
interpretable comparisons and further validation. Walk-forward evaluation,
realistic execution frictions, and transparent artifacts help assess generalization,
while transfer to live markets also requires prospective evaluation under the
relevant market and operational conditions.

\subsection{Concluding Outlook}

Looking forward, the frontier of agentic trading lies at the intersection of
deep neural perception, symbolic reasoning, and institutional-grade governance.
Promising avenues include architectures that combine neural hypothesis generation with symbolic
reasoning and constraint satisfaction, large-scale
agent-based market sandboxes for systemic risk testing, and verifiable execution
contracts with full cryptographic auditability.

By providing a clear taxonomy, decomposing the end-to-end decision pipeline, and
highlighting core methodological principles, this survey aims to serve as an
enduring conceptual roadmap for researchers, quantitative practitioners, and system
architects shaping the future of autonomous financial intelligence.

\subsection{Supplementary Materials and Reproducible Resources}

The survey's conceptual, empirical, and methodological contributions are tied to
materials accompanying the manuscript. Parts I--III present the A-C-A synthesis,
with illustrative system mappings in \ref{app:aca-reclassification};
the empirical protocol summary is reported in Supplementary S2; and the
methodological recommendations are stated as MR-1--MR-7 in Section~\ref{sec:challenges}.
The supporting review materials are organized as follows:
\begin{itemize}
	\item \textbf{Supplementary S1}: Search and screening records, including the current
	      S0 v2 candidate registry with per-record decisions.
	\item \textbf{Supplementary S2}: The current n=80 protocol registry and coverage
	      summary, alongside a preserved initial-window v1 evidence-mapping snapshot.
	\item \textbf{Supplementary S3}: Coding schema and extraction-audit materials,
	      including a targeted 15-record sample from the legacy initial-window 41-record
	      primary-role frame and a record of current amendments to historical codes.
\end{itemize}
Current per-study codes are recorded in S0 v2, with amendments documented
separately from historical coding snapshots. These materials make the selection,
protocol coding, and version changes traceable; repository availability is
described in the Data availability statement.

\section*{CRediT authorship contribution statement}
Yihan Xia: Methodology, Investigation, Data curation, Formal analysis,
Visualization, Writing -- original draft, Writing -- review \& editing.
Panpan You: Investigation, Data curation, Formal analysis, Visualization,
Writing -- original draft, Writing -- review \& editing.
Taotao Wang: Conceptualization, Methodology, Supervision, Project
administration, Resources, Validation, Writing -- review \& editing.
Fang Liu: Conceptualization, Validation, Writing -- review \& editing.
Han Qi: Conceptualization, Validation, Resources, Writing -- review \& editing.
Xiaoxiao Wu: Conceptualization, Validation, Writing -- review \& editing.
Shengli Zhang: Conceptualization, Validation, Writing -- review \& editing.

\section*{Declaration of competing interest}
The authors declare that they have no known competing financial interests or
personal relationships that could have appeared to influence the work reported
in this paper.

\section*{Acknowledgements}
The authors thank colleagues and reviewers whose feedback helped improve the
scope definition, evidence coding, and presentation of this survey.

\section*{Data availability}
The registry, protocol tables, and coding materials accompany the manuscript
as Supplementary Materials S1--S3. S1 contains the combined S0 v2 registry and
initial-window search/screening records. S2 provides the current empirical
protocol table and coverage summary alongside the historical mapping snapshot.
S3 contains the coding schema, historical annotations, current amendments, and
targeted extraction audit. An
anonymized repository containing the same materials and the LaTeX source will be
released upon acceptance, subject to journal policy and anonymization
requirements.

\nocite{deng2025autoquant,kang2026quanteval,lopez-lira2025canlargela}
\bibliographystyle{elsarticle-harv}
\begingroup
\interlinepenalty=10000
\bibliography{references}
\endgroup

\clearpage
\appendix

\section{Systematic Search Summary}
\label{app:search}

This appendix documents the information sources and query patterns that define
the survey's search boundary and support reproduction of the search.

\begin{table}[!htbp]
\footnotesize
	\centering
	\caption{Information sources and search summary (submitted as Supplementary Material S1).}
	\label{tab:search-sources}
	\begin{tabularx}{\linewidth}{@{}>{\RaggedRight\arraybackslash}p{1.7cm}>{\RaggedRight\arraybackslash}X@{}}
		\toprule
		Source              & Notes                                                                           \\
		\midrule
		ACM Digital Library & Keyword search over agentic trading, LLM agents, and financial decision-making. \\
		IEEE Xplore         & Keyword search focused on trading systems, execution, and risk control.         \\
		arXiv               & Preprints in cs.AI/cs.LG/q-fin with agent/trading keywords.                     \\
		SSRN                & Working papers in quantitative finance and computational finance.               \\
		Google Scholar      & Broad supplementary source for cross-venue discovery and citation chaining.       \\
		\bottomrule
	\end{tabularx}
\end{table}

\begin{table}[!htbp]
\footnotesize
	\centering
	\caption{Example query patterns used in the search.}
	\label{tab:search-queries}
	\begin{tabularx}{\linewidth}{@{}>{\RaggedRight\arraybackslash}p{1.7cm}>{\RaggedRight\arraybackslash}X@{}}
		\toprule
		Target              & Query pattern (example)                                                                                                                                          \\
		\midrule
		Agentic trading     & (\texttt{agent} OR \texttt{agentic} OR \texttt{LLM agent}) AND (\texttt{trading} OR \texttt{portfolio} OR \texttt{execution} OR \texttt{risk})                   \\
		LLM finance agents  & (\texttt{large language} \texttt{model} OR \texttt{LLM}) AND (\texttt{finance} OR \texttt{financial}) AND (\texttt{agent} OR \texttt{tool} OR \texttt{multi-agent})       \\
		Evaluation / validity & (\texttt{trading} OR \texttt{backtest}) AND (\texttt{look-ahead} OR \texttt{leakage} OR \texttt{transaction} \texttt{cost} OR \texttt{slippage} OR \texttt{survivorship}) \\
		\bottomrule
	\end{tabularx}
\end{table}

\textbf{Coverage note.} This survey discusses literature through the formal
registry cutoff of \textbf{2026-09-30}. The initial ledger was screened through
\textbf{2026-03-09}; at revision, we queried arXiv in an extension window
covering \textbf{2026-03-10 to 2026-09-30} with the same query families
in Table~\ref{tab:search-queries} supplemented by trading-agent,
multi-agent-market, and alpha-mining phrase patterns, retrieving 1{,}539
unique records. Of these,
\WindowExtensionRecordsN{} records entered eligibility assessment under the
same inclusion boundary; the combined v2 registry records the per-record
decisions. Supplementary S1 retains the initial-window registry, retrieval
counts, and screening logs. Extension query families are described here and in
\Cref{sec:protocol}; the extension retrieval dumps are separate working materials.
Table~\ref{tab:search-queries} lists representative example patterns; the
full six query families are enumerated in Sec.~\ref{sec:protocol}. Evidence
annotations, screening counts, and protocol-coded denominators all derive
from the combined registry snapshot.


\section{Corpus Snapshot of the Evidence Ledger}
\label{app:evidence-annotation-summary}

This appendix consolidates corpus-level descriptive findings that follow from
the review protocol but are not themselves part of the protocol definition.
Unless stated otherwise, denominator-based summaries here use the primary
empirical subset ($n=\PrimarySubsetN$), while the full included ledger contains
\EvidenceIncludedN{} studies partitioned into \EvidencePrimaryN{} primary empirical
studies and \EvidenceBackgroundN{} background studies.

The primary subset is concentrated in very recent work: within the current
ledger, \textbf{\PrimaryRecentN/\EvidencePrimaryN} primary studies were
published between 2024 and 2026. As coded by venue type in the registry,
\textbf{\PrimaryPeerReviewedN/\EvidencePrimaryN} primary studies appear in
peer-reviewed venues (journal or conference),
\textbf{\PrimaryNonPeerReviewedN/\EvidencePrimaryN} are preprints or otherwise
venue-unrecorded, and \textbf{\PrimaryPeerReviewUnknownN/\EvidencePrimaryN}
remain uncoded. The registry records the first public version year; a later journal version
may therefore have a different publication year. These counts describe the
publication landscape of the assembled corpus.

The primary subset reports different levels of protocol detail. In the pooled
80-study subset, \textbf{\ProtocolSplitReportedN/\EvidencePrimaryN} studies report
an extractable split protocol, \textbf{\ProtocolCostReportedN/\EvidencePrimaryN}
specify a transaction-cost model, \textbf{\ProtocolUniverseReportedN/\EvidencePrimaryN}
document universe/survivorship handling, \textbf{\ProtocolExecutionReportedN/\EvidencePrimaryN}
report execution timing or semantics, and
\textbf{\ProtocolArtifactsReportedN/\EvidencePrimaryN} meet the R1--R3 artifact-availability proxy, derived from the reproducibility
code rather than a separate direct artifact-extraction field.
\Cref{tab:cohort-protocol-coverage} separates these pooled rates by cohort
and records the retrieval and coding differences needed for interpretation.
Evaluation types are distributed across backtest, simulation, live, and benchmark evaluations; narrative-only and
unclear types are outside this empirical denominator.


\begin{table*}[htbp]
	\centering
	\caption{Reproducibility levels within the primary empirical subset ($n=\EvidencePrimaryN$).}
	\label{tab:reproducibility-summary}
	\begin{tabular}{@{}llp{6.5cm}rr@{}}
		\toprule
		\textbf{Level}                                                          & \textbf{Definition}           & \textbf{Criteria}                                                                                                                                                                                            & \textbf{Count} & \textbf{\%}     \\
		\midrule
		R0                                                                      & No runnable artifacts         & No publicly accessible code repository, or repository is broken/404, or it lacks essential files and cannot run                                                                                              & \ReproRZeroN   & \ReproRZeroPct  \\
		R1                                                                      & Code available (not runnable) & Public repository exists and is accessible, but key components are missing (dependencies, data access, eval scripts)                                                                                         & \ReproROneN    & \ReproROnePct   \\
		R2                                                                      & Runnable with gaps            & Complete code is available and data is accessible or clearly documented, and evaluation scripts are provided, but pinned environment, complete documentation, or an end-to-end pipeline may still be missing & \ReproRTwoN    & \ReproRTwoPct   \\
		R3                                                                      & Full reproduction package     & Complete code with pinned environment, data snapshots or clear access instructions, complete documentation, and an end-to-end evaluation pipeline                                                            & \ReproRThreeN  & \ReproRThreePct \\
		\midrule
		\multicolumn{3}{l}{\textbf{R2 or above (runnable with gaps or better)}} & \textbf{\ReproRTwoPlusN}      & \textbf{\ReproRTwoPlusPct}                                                                                                                                                                                                                      \\
		\bottomrule
	\end{tabular}
\end{table*}

\noindent\textbf{Reading note.} In the figures below, \texttt{Unknown} denotes
information that is not reported clearly enough in the source for reliable
extraction under our conservative coding policy; the panels are corpus-level
reporting distributions within the current ledger.

\begin{figure*}[!tb]
	\centering
	\begin{subfigure}{0.47\textwidth}
		\centering
		\includegraphics[width=\linewidth]{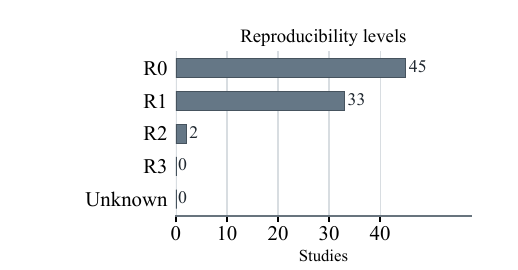}
		\caption{Reproducibility level.}
		\label{fig:repro-level-dist}
	\end{subfigure}
	\hfill
	\begin{subfigure}{0.47\textwidth}
		\centering
		\includegraphics[width=\linewidth]{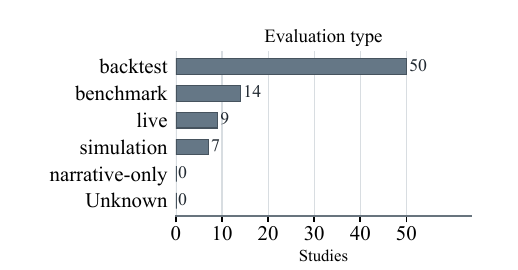}
		\caption{Evaluation type.}
		\label{fig:eval-type-dist}
	\end{subfigure}

	\vspace{0.8em}
	\begin{subfigure}{0.47\textwidth}
		\centering
		\includegraphics[width=\linewidth]{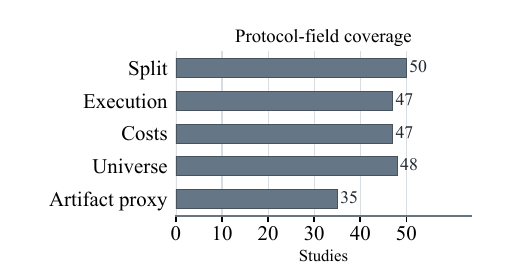}
		\caption{Protocol-field coverage.}
		\label{fig:protocol-field-coverage}
	\end{subfigure}
	\hfill
	\begin{subfigure}{0.47\textwidth}
		\centering
		\includegraphics[width=\linewidth]{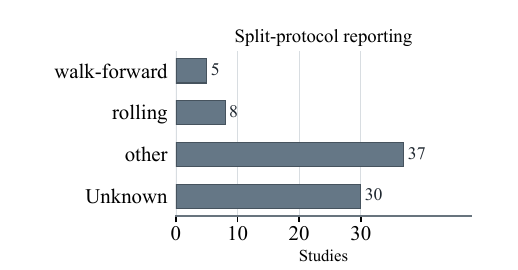}
		\caption{Split-protocol reporting.}
		\label{fig:split-protocol-dist}
	\end{subfigure}
	\caption{Corpus-level reporting distributions within the primary empirical subset ($n=\PrimarySubsetN$).}
\end{figure*}

\subsection{Extension-Cohort Landscape}
\label{app:extension-landscape}

\Cref{tab:extension-landscape} groups the extension-window background/context
studies by the role they serve in the map. Market-level multi-agent simulation
ecosystems, perception/retrieval infrastructure, and additional
validity-oriented studies are discussed in the body text
(\Cref{sec:perception,sec:memory,sec:multiagent,sec:challenges}); the table
collects representative studies from the remaining background cohort so that the extension window's
coverage can be inspected in one place.

\begin{table*}[!t]
	\centering
	\small
	\setlength{\tabcolsep}{4pt}
	\renewcommand{\arraystretch}{1.15}
	\caption{Extension-cohort landscape: background/context studies included in
		the ledger (2026-03-10 to 2026-09-30), grouped by the role they play for
		readers. These studies inform design and evaluation practice but are not
		part of the primary empirical subset ($n=\EvidencePrimaryN$).}
	\label{tab:extension-landscape}
	\begin{tabularx}{\linewidth}{@{}p{4.2cm} X@{}}
		\toprule
		\textbf{Reader-facing role} & \textbf{Representative studies} \\
		\midrule
		Benchmarks \& evaluation harnesses &
		BacktestBench \cite{wang2026backtestbench},
		Backtrader-Bench \cite{zhao2026backtrader},
		FinResearchBench II \cite{luan2026finresearchbench},
		BankerToolBench \cite{lau2026bankertoolbench},
		Foresight Arena \cite{nechepurenko2026foresight},
		ValueBlindBench \cite{chang2026valueblindbench},
		ShiJianBench \cite{gong2026shijianbench},
		FrontierFinance \cite{zhang2026frontierfinance},
		QuantEval \cite{kang2026quanteval},
		reliable multi-agent evaluation \cite{nguyen2026toward} \\
		\midrule
		Research tooling \& agent infrastructure &
		AlphaLab \cite{hogan2026alphalab},
		PolyGnosis 2.0 \cite{wang2026polygnosis},
		interaction-native knowledge harnesses \cite{borjigin2026absorbing},
		RIDGE \cite{cheng2026ridge},
		PandaAI \cite{li2026pandaai},
		agentic forecasting infrastructure \cite{chai2026agentic} \\
		\midrule
		Surveys \& cross-cutting analyses &
		agentic AI in finance \cite{aldridge2026agentic},
		LLM stock forecasting review \cite{zhang2026review},
		AI in equity/crypto markets \cite{zhu2026artificial} \\
		\midrule
		Adjacent financial-agent applications &
		AI-economist frameworks \cite{kato2026ai},
		multi-agent venture-capital due diligence \cite{alexandrou2026multi},
		verifiable forecast actions \cite{chen2026reasoning} \\
		\bottomrule
	\end{tabularx}
\end{table*}


\section{Representative A-C-A Reclassification Notes}
\label{app:aca-reclassification}

This appendix provides reclassification notes for 19 representative
works to illustrate how the Architecture-\hspace{0pt}Capability-\hspace{0pt}Adaptation (A-C-A)
lens distinguishes system design choices. For each work, we document: (i)
classification under prior frameworks (where available), (ii) A-C-A
reclassification with coding rationale, and (iii) the specific analytical
insight surfaced by the lens.

\subsection{Reclassification Procedure}

\textbf{Paper Selection.} We selected 19 works representing diverse architectural approaches, capabilities, and adaptation mechanisms. Selection criteria: (a) coverage of all three A-C-A dimensions, (b) citation count or recency (indicating influence), (c) availability of technical details sufficient for classification.

\textbf{Coding Procedure.} For each paper, one author prepared an initial A-C-A
classification and coding rationale, and the authors then reconciled edge cases
through discussion. The resulting notes document how the lens is applied to
the selected works. Inter-rater agreement and external validation of these
A-C-A classifications were not assessed.

\textbf{Prior Framework Identification.} For each work, we identified its classification in existing surveys where available: Wang et al. 2024 (LLM-Agent Survey), Ding et al. 2024 (LLM-Finance Survey), or task-centric benchmarks (FinBen, InvestorBench). Where a work was not explicitly labeled in one of those sources, the shorthand in the Prior column should be read as our closest reconstruction of how that work would map to the earlier scheme.

\subsection{Reclassification Dataset}

Memory descriptors in the table retain terms used in individual systems.
In the survey's operational taxonomy, experience retrieval is discussed under
associative vector storage when vector indexing is specified, while explicit
financial knowledge stores are discussed under structured knowledge bases;
these descriptors do not by themselves establish a particular implementation.

\Cref{tab:aca-reclassification-full} presents the complete
reclassification notes, organized by architectural family.

\begin{table*}[!tb]
	\centering
	\footnotesize
	\setlength{\tabcolsep}{2pt}
	\renewcommand{\arraystretch}{1.05}
	\caption{Representative A-C-A reclassification notes for 19 works, documenting architectural components, financial capabilities, adaptation mechanisms, and the rationale for each mapping.}
	\label{tab:aca-reclassification-full}
	\begin{adjustbox}{max width=\textwidth}
	\begin{tabularx}{\linewidth}{@{}>{\RaggedRight\arraybackslash}p{2.1cm} p{1.4cm} X p{0.45cm} p{0.45cm} p{0.45cm} p{0.85cm} >{\RaggedRight\arraybackslash}p{3.8cm}@{}}
		\toprule
		\textbf{Work}                                        & \textbf{Prior}                     & \textbf{A-C-A Reclassification}                                                                             & \textbf{P} & \textbf{M} & \textbf{R} & \textbf{Adapt} & \textbf{Key Insight from A-C-A}                        \\
		\midrule
		FinBERT \cite{araci2019finbert}                      & W: ``NLP Model'', D: ``Sentiment'' & Arch: Text Perception only; Cap: BG (no Action); Adapt: BG                                                  & \checkmark & --         & --         & BG             & Excluded as BG: demonstrates Action        Output criterion excludes pure prediction                         \\
		\addlinespace[1pt]
		FinGPT \cite{fingpt_2023}                            & W: ``LLM Tool'', D: ``Generation'' & Arch: Text Perception + Static Memory; Cap: BG (no closed-loop); Adapt: BG                                  & \checkmark & \checkmark & --         & BG             & Excluded as BG: Generation without         execution integration                                              \\
		\addlinespace[1pt]
		FinLlama \cite{iacovides2024finllama}                & D: ``Sentiment''                   & Arch: Text Perception; Cap: BG; Adapt: BG                                                                   & \checkmark & --         & --         & BG             & Same architecture as FinBERT but           instruction-tuned; capability depends on integration, not just model \\
		\addlinespace[1pt]
		\addlinespace[1pt]
		AlphaAgent \cite{alphaagent_2025}                    & D: ``Multi-agent''                 & Arch: Reflective/Strategic hybrid + Episodic Memory; Cap: $\alpha$; Adapt: Self-reflection                  & \checkmark & \checkmark & \checkmark & SE             & Multi-agent debate = validation mechanism  for Alpha, not coordination capability                      \\
		\addlinespace[1pt]
		FinAgent \cite{wu2024finagent}                       & W: ``Multi-modal'', D: ``Agent''   & Arch: Multi-modal P + Episodic M + Reflective R + Microstructure A; Cap: $\alpha$ (cross-modal); Adapt: ICL & \checkmark & \checkmark & \checkmark & ICL            & Core: cross-modal perception-to-action     pipeline, not ``multi-modality''                               \\
		\addlinespace[1pt]
		Alpha$^2$ \cite{alpha2_2024}                         & B: ``Alpha Mining''                & Arch: Working Memory + Strategic Reasoning (MCTS); Cap: $\alpha$ (evolutionary); Adapt: RL                  & --         & \checkmark & \checkmark & RL             & MCTS is Strategic Reasoning; evolutionary  = search-based adaptation                                   \\
		\addlinespace[1pt]
		RAG-Fintech \cite{rag_fintech_2025}                  & D: ``Retrieval''                   & Arch: Text P + Semantic M (KB) + Reflective R; Cap: $\alpha$ (retrieval-based); Adapt: ICL                  & \checkmark & \checkmark & \checkmark & ICL            & Retrieval = Memory architecture, not just  augmentation technique                                      \\
		\addlinespace[1pt]
		RAPTOR \cite{raptor_2025}                            & D: ``Portfolio''                   & Arch: Working M + Strategic R + Execution A; Cap: P (hierarchical); Adapt: ICL                              & --         & \checkmark & \checkmark & ICL            & Portfolio = Strategic + Execution          integration, not separate capability                                \\
		\addlinespace[1pt]
		AgentGuard \cite{chen2025agentguard}                   & W: ``Safety''                      & Arch: Reactive R + Action A with gating; Cap: R (real-time); Adapt: Rule-based                              & --         & --         & \checkmark & --             & Risk as gating mechanism in Action architecture, not post-hoc                                         \\
		\addlinespace[1pt]
		FinRS \cite{liu2025risksensit}                       & D: ``Risk''                        & Arch: Text P + Episodic M + Reflective R; Cap: R (risk-sensitive); Adapt: Reward reflection                             & \checkmark & \checkmark & \checkmark & ICL            & Risk capability derives from               Perception-Memory-Reasoning chain                                        \\
		\addlinespace[1pt]
		TradingAgents \cite{xiao2024tradingagents}           & W: ``Multi-agent''                 & Arch: Hierarchical role-based Coordination; Cap: $\alpha$+P+R distributed; Adapt: Hierarchical feedback     & \checkmark & \checkmark & \checkmark & ICL + SE         & Analyst, research, trader, risk, and manager teams form a decision hierarchy                                                   \\
		\addlinespace[1pt]
		TradingGroup \cite{tradinggroup_2025}                & D: ``Multi-agent''                 & Arch: Role-based + Shared Memory; Cap: $\alpha$; Adapt: Self-reflection                                     & \checkmark & \checkmark & \checkmark & SE             & Self-reflection coordinates roles,         not external mechanism                                           \\
		\addlinespace[1pt]
		FINCON \cite{yu2024fincon}                           & D: ``Consensus''                   & Arch: Role-based + Communication Protocol; Cap: P; Adapt: CVRF belief updates                                      & \checkmark & --         & \checkmark & ICL            & CVRF updates investment beliefs across episodes                                     \\
		\addlinespace[1pt]
		FinMem \cite{yu2023finmem}                           & W: ``Memory''                      & Arch: Working M + Layered long-term M + Strategic R; Cap: P; Adapt: Memory consolidation                       & --         & \checkmark & \checkmark & SE             & Memory hierarchy enables temporal          abstraction for Portfolio                                           \\
		\addlinespace[1pt]
		MetaGPT \cite{xiao2023metagpt}                       & W: ``Multi-agent''                 & Arch: Hierarchical (Manager + Engineers); Cap: BG (software engineering); Adapt: SOP-guided prompting                       & \checkmark & \checkmark & \checkmark & --             & General workflow design; no evaluated trading capability                                             \\
		\addlinespace[1pt]
		FinRL-DeepSeek \cite{benhenda2025finrl}              & B: ``RL Trading''                  & Arch: Working M + Reactive/Strategic hybrid R + Execution A; Cap: $\alpha$+P; Adapt: RL                     & --         & \checkmark & \checkmark & RL             & RL spans multiple capabilities with        shared architecture                                               \\
		\addlinespace[1pt]
		Reflexion \cite{shinn2023reflexion}                  & W: ``Reflection''                  & Arch: Working + Episodic M + Reflective R; Cap: BG (general); Adapt: Self-reflection                                   & --         & \checkmark & \checkmark & SE             & Self-reflection as adaptation mechanism    applicable to trading                                         \\
		\addlinespace[1pt]
		Hashimoto \cite{hashimoto2023agent}                  & B: ``Simulation''                  & Arch: Market Ecology (no central coordination); Cap: $\alpha$+P+R emergent; Adapt: Evolutionary             & \checkmark & \checkmark & \checkmark & RL + SE          & Ecology = emergent capabilities from       interaction, not design                                       \\
		\addlinespace[1pt]
		FinEvo \cite{zou2026finevo}                          & D: ``Evolution''                   & Arch: Market Ecology + Hierarchical; Cap: $\alpha$; Adapt: Evolutionary + SE                                & \checkmark & \checkmark & \checkmark & SE             & Combines ecology (interaction) with        hierarchy (structure)                                             \\
		\bottomrule
	\end{tabularx}
	\end{adjustbox}
	\par\vspace{3pt}
	\begin{minipage}{\linewidth}\footnotesize
		Prior-column shorthand: W = Wang et al.\ 2024 (LLM-Agent Survey); D = Ding et al.\ 2024 (LLM-Finance Survey); B = task-centric benchmark (FinBen, InvestorBench). Where a work was not explicitly labeled in those sources, the shorthand is our closest reconstruction of how it would map to the earlier scheme. P/M/R mark the Perception--Memory--Reasoning architectural components.
	\end{minipage}
\end{table*}

\subsection{Illustrative Patterns from Reclassification}

\textbf{Pattern 1: Architecture-Capability Non-Alignment.}
The same Reactive Architecture (FinAgent, AgentGuard) supports different capabilities ($\alpha$ vs R) depending on perceptual input and memory integration. This cross-cutting pattern is obscured by capability-centric categorizations.

\textbf{Pattern 2: Adaptation Level Distinction.}
The adaptation dimension distinguishes prompt- and memory-mediated updates
(FinAgent, RAG-Fintech, FinRS, and FinCon) from weight-based RL
(Alpha$^2$, FinRL-DeepSeek) and structural search (AlphaAgent, FinEvo).
MetaGPT provides a background example of SOP-guided prompting, rather than
financial SFT. Each update mechanism has distinct evaluation requirements
(Sec.~\ref{sec:challenges}).

\textbf{Pattern 3: Multi-Agent Architectural Hierarchy.}
Prior frameworks often use broad ``multi-agent'' labels. The A-C-A lens
distinguishes role-based, hierarchical, and market-ecology coordination patterns,
making the coordination mechanism visible rather than treating agent count as
the primary distinction.

\subsection{Coding Status}

The coding notes document qualitative author reconciliation and the rationale
for boundary cases, providing a traceable account of applying the A-C-A lens
to the selected works.

\subsection{Limitations of the Reclassification Notes}

\begin{enumerate}
	\item \textbf{Selection Bias:} 19 works selected from a broader review corpus spanning included studies, background references, and a small number of boundary cases used to stress-test the A-C-A lens; they were not randomly sampled.
	\item \textbf{Temporal Coverage:} 15 of these 19 illustrative works first appeared in 2024--2026 (79\%),
using the first public version year rather than a later publication year.
This denominator is separate from the 80-study empirical subset and reflects field recency. Earlier works may reveal different architectural patterns.
	\item \textbf{Prior Framework Availability:} Not all works classified in existing surveys. Where unavailable, we inferred likely categorization based on paper content and survey scope.
\end{enumerate}

Across these examples, the A-C-A lens distinguishes design choices and
surfaces architectural patterns that may be obscured by broader
system- or task-level labels.


\section{Excluded Studies and Rationale}
\label{app:excluded-studies}

\begin{table*}[htbp]
	\centering
	\small
	\setlength{\tabcolsep}{2pt}
	\caption{Representative final exclusions and background boundary cases.}
	\label{tab:excluded-studies}
	\begin{tabularx}{\linewidth}{@{}p{3.2cm}p{1.3cm}X@{}}
		\toprule
		\textbf{Study (Representative)}     & \textbf{Category} & \textbf{Decision rationale}                                                    \\
		\midrule
		\multicolumn{3}{l}{\textit{Final excluded records in the current canonical registry}}                                                                                \\
		\midrule
		Socio-Agentic LLMs for Socio-Technical Systems & E3 & Generic socio-technical agent paper outside finance/trading/portfolio/risk management. \\
		Learning Game-Playing Agents with Generative Code Optimization & E3 & General game-playing agent study without financial-market instantiation. \\
		NeSyCoCo & E3 & Neuro-symbolic compositional generalization study outside the trading domain. \\
		Agentic AI for Commercial Insurance Underwriting & E3 & Insurance underwriting rather than trading, portfolio construction, execution, or market-risk management. \\
		\midrule
		\multicolumn{3}{l}{\textit{Included background boundary cases}}                                                                       \\
		\midrule
		FinBERT (Araci 2019)                & E1/BG                & Finance-relevant sentiment model retained as background; no tradable action loop, so it is excluded from the primary empirical subset. \\
		FinGPT (Wang 2023a)                 & E1/BG                & Finance language-model/tooling work retained as background; no explicit closed-loop trading execution integration.        \\
		InvestorBench  & E1/BG                       & Finance benchmark retained as background; the registry codes no tradable action output, so it does not enter the primary empirical subset. \\
		\bottomrule
	\end{tabularx}
\end{table*}

\begin{figure*}[!tb]
	\centering
	\begin{tikzpicture}[
            font=\small,
            box/.style={rectangle, rounded corners, draw, text width=2.0cm, minimum height=1.0cm, align=center},
            decision/.style={rectangle, rounded corners, draw, text width=2.5cm, minimum height=1.2cm, align=center, fill=tradingagent_main!15},
            arrow/.style={->, >=stealth}
        ]
        \node[box, fill=gray!10] (start) at (0,0) {Candidate registry\\($n=\CandidateRecordsN$)};
        \node[box, fill=gray!15] (dedup) at (2.7,0) {Deduplicated\\($n=\RecordsAfterDedupN$)\\0 removed};
        \node[decision] (domain) at (5.7,0) {Finance/trading\\domain?};
        \node[box, fill=tradingagent_blue!12] (included) at (9,0) {Included ledger\\($n=\EvidenceIncludedN$)};
        \node[box, fill=tradingagent_red!10] (excluded) at (5.7,-2.2) {Outside scope\\($n=\CandidateExcludedN$)};
        \node[decision, text width=3.5cm] (empirical) at (11.8,-2.2) {Action Output + Closed-Loop\\+ Eligible Evaluation Type?};
        \node[box, fill=tradingagent_blue!25] (primary) at (15.9,0) {Primary empirical\\($n=\EvidencePrimaryN$)};
        \node[box, fill=tradingagent_blue!8] (background) at (15.9,-2.2) {Background\\($n=\EvidenceBackgroundN$)};
        \draw[arrow] (start) -- (dedup);
        \draw[arrow] (dedup) -- (domain);
        \draw[arrow] (domain) -- node[above] {Yes} (included);
        \draw[arrow] (domain) -- node[left] {No} (excluded);
        \draw[arrow] (included) |- (empirical);
        \draw[arrow] (empirical) |- node[pos=.25,right] {Yes} (primary);
        \draw[arrow] (empirical) -- node[midway,above,fill=white,inner sep=1pt] {No} (background);
    \end{tikzpicture}
\caption{Inclusion, exclusion, and evidence-tier decision flow. Final exclusions are outside the survey domain; finance-relevant records that fail Action Output, Closed-Loop Evaluation, or the eligible empirical evaluation-type rule remain in the included background tier.}
	\label{fig:selection-flow}
\end{figure*}

To keep the evidence-mapping process auditable, this appendix distinguishes
two decisions that are easily conflated. First, \emph{final exclusions} are
candidate-registry records outside the
finance, trading, portfolio, or risk-management scope of this survey. Second,
\emph{background-tier boundary decisions} are finance-relevant records that
remain in the included evidence ledger but do not enter the primary empirical
subset because they fail at least one of three criteria: Action Output,
Closed-Loop Evaluation, or an eligible empirical evaluation type. We document
both decision types---using PRISMA-ScR as a reporting aid for scoping-style
transparency rather than as a claim of a full systematic-review workflow---to
facilitate audit and assessment of potential selection bias.

\subsection{Exclusion Categories and Boundary Decisions}

Candidate records were counted as final exclusions only when they fell outside
the domain boundary of this survey. Finance-relevant records that were
prediction models, benchmarks, tools, or conceptual background papers were
retained as background or context where useful, but were excluded from the
primary empirical denominator. The operational criteria are:

\begin{enumerate}
	\item[\textbf{E1}] \textbf{No Action Output}: Pure prediction or analysis systems that do not generate tradable actions. Finance-relevant examples remain included as background.

\item[\textbf{E2}] \textbf{No Eligible Empirical Evaluation}: Finance-relevant studies without a qualifying backtest, simulation, live-trading, or benchmark evaluation. Narrative-only or unclear evaluation reports remain background/context even when action output and a closed-loop claim are coded present.

	\item[\textbf{E3}] \textbf{Non-Financial Domain}: Systems applied to non-financial markets or generic agent settings without financial trading, portfolio, execution, or risk-management instantiation. These form the final excluded set in the current canonical registry.

	\item[\textbf{E4}] \textbf{No LLM Component}: Traditional quantitative trading systems without a language-model component. These may be cited only as background context when needed for finance methodology.

	\item[\textbf{E5}] \textbf{Duplicate}: Multiple versions of the same work; the most complete version is retained.

	\item[\textbf{E6}] \textbf{Inaccessible}: Full text unavailable after retrieval attempts.
\end{enumerate}

\subsection{Examples of Final Exclusions and Background Cases}

\Cref{tab:excluded-studies} separates final exclusions from background
boundary cases. The aggregate exclusion count in the current canonical registry
is \CandidateExcludedN, all of which are outside the finance/trading scope; the
background cases remain in the \EvidenceIncludedN-record included ledger but do
not enter the \EvidencePrimaryN-record primary empirical subset.
Under the current boundary, the 102 included records tagged \texttt{primary} in
S0 v2 include 22 narrative-only records; these are assigned to the background
denominator, yielding 80 empirical primary and 105 background records.

\subsection{Inclusion/Exclusion Decision Flow}

\Cref{fig:selection-flow} illustrates the screening logic encoded in the canonical ledger and summarized in the main text, making the inclusion/exclusion workflow traceable.

\subsection{Impact of Exclusion Criteria on Evidence Set}

After deduplication, all \CandidateExcludedN{} final exclusions fall outside the
survey domain. Finance-relevant records failing the three-part empirical
criterion remain background/context. Reassigning 22 \texttt{narrative-only}
studies among 102 \texttt{primary}-tagged records yields 80 empirical primary
and 105 background studies. \Cref{tab:selection-flow-summary} reports these
macro-backed categories, rather than earlier screening-log totals.

\flushcolsend

\end{document}